\documentclass[11pt]{article}
\usepackage{fullpage}
\usepackage{microtype}
\usepackage{graphicx}
\usepackage{subcaption}
\usepackage{amsmath,amsbsy,amsfonts,amssymb,amsthm,bm}
\usepackage{algorithm,algorithmic,mathtools}
\usepackage{color,cases,multirow}
\usepackage[colorlinks=True, citecolor=blue]{hyperref}

\usepackage{authblk}

\newcommand\blfootnote[1]{%
  \begingroup
  \renewcommand\thefootnote{}\footnote{#1}%
  \addtocounter{footnote}{-1}%
  \endgroup
}

\newtheorem{theorem}{Theorem}

\theoremstyle{definition}

\DeclareMathOperator{\argmin}{argmin}

\newcommand{\RR}{\mathbb{R}}

\newcommand{\EE}{\mathbb{E}}

\renewcommand{\SS}{\mathbb{S}}

\newcommand{\cH}{\mathcal{H}}

\newcommand{\cR}{\mathcal{R}}

\newcommand{\bx}{\mathbf{x}}

\newcommand{\ba}{\bm{a}}
\newcommand{\bb}{\bm{b}}
\newcommand{\be}{\bm{e}}
\newcommand{\bB}{\bm{B}}

\newcommand{\bi}{\begin{enumerate}}
\newcommand{\ei}{\end{enumerate}}

\title{The Quenching-Activation Behavior of the Gradient Descent Dynamics for Two-layer Neural Network Models}

\author[2]{Chao Ma \thanks{\texttt{cham@princeton.edu}}}
\author[2]{Lei Wu \thanks{\texttt{leiwu@princeton.edu}}}
\author[1,2]{Weinan E \thanks{\texttt{weinan@math.princeton.edu}} \footnote{Also at Beijing Institute of Big Data Research.}
}

\affil[1]{Department of Mathematics, Princeton University}
\affil[2]{Program in Applied and Computational Mathematics, Princeton University}

\date{}

\begin{document}

\maketitle
\blfootnote{The first two authors contributed equally.} 

\begin{abstract}
A  numerical and phenomenological study of the gradient descent 
(GD) algorithm for training two-layer neural network models   
is carried out for different parameter regimes when the target function can be accurately 
approximated by a relatively small number of neurons. 
It is found that  for Xavier-like initialization, there are two distinctive 
phases in the dynamic behavior of GD in the under-parametrized regime:
An early phase in which the GD dynamics follows closely that of the corresponding 
random feature model and the neurons are effectively quenched, followed by
a late phase in which the neurons are divided into two groups: a group of  a few ``activated'' neurons that dominate the dynamics
and a group of background (or ``quenched'') neurons that support the continued activation and deactivation process.
This neural network-like behavior is continued into the mildly over-parametrized regime, where it undergoes a transition to a random feature-like behavior. 
The quenching-activation process seems to provide a clear mechanism for  ``implicit regularization''.
This is qualitatively different from the dynamics associated with the ``mean-field'' scaling where all neurons participate
equally and there does not appear to be qualitative changes when the network parameters are changed.
\end{abstract}

\section{Introduction}

In the past few years, much effort has been devoted to the understanding of the theoretical foundation
behind the spectacular success of neural network (NN)-based machine learning (ML)  and the many mysteries that surround it.
The main theoretical questions concern the convergence rate for the generalization error and
the optimization process.
It is impossible to review all the progresses that have been made so far. 
But we can summarize the major achievements together with the
  major remaining mysteries as follows.

1.  It has been realized that the main issue is the curse of dimensionality (CoD): As the dimensionality increases,
the number of parameters, the size of the training dataset or the number of iterations in the 
optimization algorithms needed may increase exponentially as a function
of the error tolerance. Classical approximation schemes that use
fixed basis functions such as splines or wavelets all suffer from  CoD. Training  neural network models to fit
target functions that are not in the right function spaces also suffer from CoD \cite{wojtowytsch2020can}.
 
2.  For two-layer neural network (2LNN) and the deep  residual neural network models, it has been proved that solutions with
``good'' generalization properties do exist. 
Specifically it has been shown that for the appropriate classes of target functions, 
the generalization error associated with the
global minimizers of  some properly regularized
2LNN  and deep residual neural network models obey Monte Carlo-like estimates: $O(1/m)$ for the approximation error and
$O(1/\sqrt{n})$ for the generalization gap where $m$ and $n$ are the number of free parameters in the model and the size
of the training dataset respectively \cite{e2018priori,ma2019priori}. 
The fact that these estimates do not suffer from CoD is one of the fundamental reasons behind the success of neural network
models in high dimensions.

 3. An important open questions is:
 Do standard optimization algorithms used in practice find good solutions?  
 
 It has been observed in practice that small test error can often be achieved with appropriate choices of the hyper-parameters,
 without the need for introducing explicit regularization \cite{neyshabur2014search, zhang2016understanding}.  Since NN-based models often work in the over-parametrized regime
 where the empirical risk can be reduced to zero for a large set of parameter choices \cite{zhang2016understanding}, some of which give rise to large test errors,  this means that there are some ``implicit regularization'' mechanisms at work for the optimization algorithm
 with these particular choices of hyper-parameters. 
At the same time, the need for extensive tuning  suggests that the performance of these models depends sensitively
on the choice of the hyper-parameters.  
It is important to understand the origin of this sensitive dependence.
 
4.  On the issue of the training process, a rather complete picture has been established for highly
over-parametrized NN models. 
Unfortunately the overall result is somewhat disappointing:  While one can prove
that GD converges to a global minimizer of the empirical risk \cite{du2018gradient,du2018deepgradient}, the generalization properties of this global minimizer is no
better than that of an associated random feature model  (RFM) \cite{jacot2018neural,ma2019comparative,arora2019exact}.   In fact, it has been shown that the entire GD paths for the
NN model and the associated RFM stay uniformly close for all time \cite{ma2019comparative}.
Consequently the NN model  performs no better than the  RFM and there is no ``implicit regularization'' in this regime.
This is a disappointing result from a theoretical viewpoint, since it sheds no light on the origin of the improved performance
of NN models over the RFMs or kernel methods. 

A natural  question is then:  
Can there be implicit regularization when the network is less highly over-parametrized?
What would be the mechanism for implicit regularization?

5.  Another important progress for the understanding of the training dynamics is the emergence of the ``mean-field'' models
\cite{chizat2018global,song2018mean,rotskoff2018parameters,sirignano2018mean}. It has been observed that if an extra $1/m$  factor is added to the expression for the NN functions (recall that $m$ denotes the number of neurons), then
the GD dynamics of the  NN model can be written equivalently as a continuous integral-differential equation (IDE) for the
evolution of the probability distribution over the parameter space.  
A nice feature of this IDE is that it is nothing but the Wasserstein gradient flow of the risk function.
Therefore one is hopeful that
the analytical tools developed in the optimal transport theory might be brought to bear.
Unfortunately, at the moment theoretical results are still quite sparse for this IDE (see however \cite{chizat2018global, chizat2020implicit, wojtowytsch2020convergence}).
Nevertheless there is no doubt that this IDE is a very interesting mathematical problem and could lead to much needed insight on 
the GD dynamics of NN models.


In practice, more often than not, people do not add this extra $1/m$ factor.
So a natural question is: What is the difference between the GD dynamics under the conventional and mean-field scaling?

In this series of papers, we set out to investigate these issues numerically and phenomenologically.
Our objective is to get some insight from this kind of experimental studies, which we hope will  be helpful for 
subsequent theoretical work.
In this paper we will focus on the original GD dynamics for 2LNN models.
Subsequent papers will consider multi-layer NN models as well as  accelerated training algorithms.

We will focus on the regression problem although most of the earlier successes of neural network-based machine 
learning were on classification problems . This is because that at the moment, our theoretical understanding for the classification
problem is much less advanced than for the regression problem, and this set of studies is very much guided by theory and for the 
purpose of theory.
We will also be working entirely with target functions that are made up.

\paragraph*{Notation} 
Throughout this paper, we denote by $\pi_0$ the uniform distribution over $\SS^{d-1}:=\{\bx\in\RR^d : \|\bx\|_2=1\}$. We use $X\lesssim Y$ to indicate that there exists an absolute constant $C$ such that $X\leq C Y$.  $X\gtrsim Y$ is similarly defined.


\section{Preliminaries}
Let us start with the basic formulation of supervised learning:
Given a dataset $ S=\{(\bx_i, y_i=f^*(\bx_i)), i=1, \cdots, n\}$, our task is to recover the target function $f^*$ as accurately as possible.
Throughout this paper, we assume that the data is drawn from $\pi_0$.

Denote by  $\cH_m$ the hypothesis space, where $m$ is roughly the number of free parameters 
for functions in the hypothesis space. Denote by $\hat{f}$ the model learned from the training set $S$.
Let
$$f_m = \argmin_{f \in \cH_m} \mathcal{R}(f),
$$
where $\cR(f)=\EE_{\bx}[(f(\bx)-f^*(\bx))^2]$ is the population risk. 
We can decompose the error $f^* -\hat{f}$ into:
$$ f^* - \hat{f} = f^* - f_m + f_m - \hat{f} $$
$f^* - f_m$ is the {\it approximation error}, due entirely to the choice of the
hypothesis space $\cH_m$.
$f_m - \hat{f}$ is the {\it estimation error}, the additional error due to the fact
that we only have a finite dataset. 
Typically the error behaves like 
$$ f^* - \hat{f} \sim m^{-\alpha(d)} +  n^{-\beta(d)} $$
If $\alpha(d)$ or $\beta(d)$ decreases (to 0)  as $d \rightarrow \infty$, we say that the model suffers from CoD. 
Since it is difficult to study the actual limit as $d \rightarrow \infty$, we will use as a working definition  that if 
$\alpha(d)$ or $\beta(d)$ shows clear dependence on $d$, then we say that there is a CoD.


\subsection{Two-layer neural networks}

Under conventional scaling, a two-layer neural network model is given by:
\begin{equation}
    f_m(\bx;\ba,\bB) = \sum_{j=1}^m a_j \sigma(\bb_j^T\bx) = \ba^T\sigma(\bB \bx),
\end{equation}
where $\ba\in \RR^m, \bB=(\bb_1, \bb_2,\dots, \bb_m)^T \in\RR^{m\times d}$, { $\sigma(t)=\max(0,t)$ is the ReLU activation function}. 
Later we will also consider the mean-field scaling where the expression above is replaced by:
\begin{equation}
    f_m(\bx;\ba,\bB) = \frac 1m \sum_{j=1}^m a_j \sigma(\bb_j^T\bx) = \frac 1m \ba^T\sigma(\bB \bx),
\end{equation}
but we will focus on the conventional scaling unless indicated otherwise.

The empirical risk and the population risks are given by 
\begin{align}
    \hat{\cR}_n(\ba,\bB) &= \frac{1}{n}\sum_{i=1}^n (f_m(\bx_i;\ba,\bB)-f^*(\bx_i))^2\\
    \cR(\ba,\bB) &= \EE_{\bx}[(f_m(\bx;\ba,\bB)-f^*(\bx))^2]
\end{align}
respectively. 

As a comparison, the random feature model is given by $f_m(\bx;\ba,\bB_0)$, where only the coefficient $\ba$ 
can be varied,  $\bB_0$ is randomly sampled from $\pi_0$ and is fixed during training. 


\subsection{GD dynamics with Xavier-type initialization}

The GD flow is given by:
\begin{equation}\label{eqn: GD-xv}
\begin{aligned}
    \dot{a}_j(t) &= -\frac{1}{n}\sum_{i=1}^n (f_m(\bx_i;\ba(t),\bB(t))-f^*(\bx_i))\sigma(\bb_j(t)^T \bx_i)\\
    \dot{\bb}_j(t) &= - \frac{1}{n}\sum_{i=1}^n (f_m(\bx_i;\ba(t),\bB(t))-f^*(\bx_i))a_j(t)\sigma'(\bb_j(t)^T\bx_i)\bx_i
\end{aligned}
\end{equation}
In practice, one has to discretize this flow using some finite learning rate.  We will consider the situation when 
the learning rate is fixed and sufficiently small such that the results reported below can be considered as the results
for the GD flow. 

We consider the  Xavier-type initialization 
\begin{equation}
    a_j(0) \sim \mathcal{N}(0,\beta^2), \qquad \bb_j(0)\sim \mathcal{N}(0,I/d).
\end{equation}
For the original Xavier initialization, $\beta=1/\sqrt{m}$. 
However,  we have found consistently for 2LNNs that the behavior of the GD dynamics is qualitatively 
 very close when $\beta=0$.
 We  refer to Appendix \ref{sec: beta-infuence} for some numerical results along this line. 
 This can also been seen from Theorem 3.3 in \cite{ma2019comparative} 
 which provides a rigorous justification of this observation in the  highly over-parametrized regime.  
 For simplicity we will focus on the case when  $\beta=0$. 

One of the main theoretical advances in the last few years is a thorough understanding of the optimization
and generalization properties in the {\it highly over-parametrized } regime. Specifically, 
it was proved in \cite{du2018gradient}  that GD converges exponentially fast to a
global minimizer in this regime.  
Subsequently it was shown in \cite{ma2019comparative}  that these GD solutions are uniformly close to
that of  the associated RFM with $\bB_0=\bB(0)$ as the features.
The key quantity used in the analysis is the Gram matrix $K:=(K_{i,j})\in\RR^{n\times n}$ with 
\[
    K_{i,j} = \frac{1}{n}\EE_{\bb\sim\pi_0}[ \sigma(\bx_i^T\bb)\sigma(\bx_j^T\bb)].
\]
The following theorem summarizes the main results of \cite{du2018gradient,ma2019comparative}.
\begin{theorem}\label{thm: hop}
Let $\lambda_n=\lambda_{\min}(K)$ and assume $\beta=0$. Denote  by $f_m(\bx;\tilde{\ba}(t), \bB_0)) $ the solutions of the
GD dynamics for the random feature model. For any $\delta\in (0,1)$, assume that $m\gtrsim n^2\lambda_n^{-4}\delta^{-1}\ln(n^2\delta^{-1})$. Then with probability at least $1-6 \delta$ we have 
\begin{align}
\hat{\cR}_n(\ba(t),\bB(t))&\leq e^{-m\lambda_n t}\hat{\cR}_n(\ba(0), \bB(0))\\
\sup_{\bx\in \SS^{d-1}}    |f_m(\bx;\ba(t), \bB(t)) - f_m(\bx;\tilde{\ba}(t),\bB_0)| &\lesssim \frac{(1+\sqrt{\ln(\delta^{-1})})^2\lambda_n^{-1}}{\sqrt{m}}.
\end{align}
\end{theorem}

The proof rests upon the following observations:
        \bi
     \item Time scale separation
     \begin{align}
            \dot{a}_j(t) &\sim O(\|\bb_j\|) =O(1)\\
            \dot{\bb}_j(t)&\sim O(|a_j|) = O\left(\frac1 {\lambda_n m}\right).
     \end{align}
In fact, in the highly over-parametrized regime, the dynamics of $\bb$ is effectively frozen.  Therefore GD for the 2LNN
degenerates to the GD for the corresponding RFM.

\item The Gram matrix $K$ for the corresponding RFM is non-degenerate \cite{xie2016diverse}, i.e. its eigenvalues are bounded away from 0.
This is responsible for the exponential convergence behavior.
\ei
From these results, we learn the following:   (1) In the highly over-parameterized regime, there are no  ``implicit regularization" effects  since the solutions found by the GD dynamics for the 2LNN models perform no better than that of the RFM solutions.  
(2) Optimization and generalization are very different issues. 

The picture described above contradicts with our experience in practice, where we do observe that NN models outperform RFM in terms of test accuracy, even in the over-parametrized regime.
This begs for the question:
Can there be implicit regularization in  the``mildly" over-parametrized regime, and if there is, what would be the 
mechanism for  such implicit regularization?
More generally, what is the qualitative behavior of the GD dynamics in different regimes including the under-parametrized regime?
A side issue is:
What happens to the time scale separation?
These and other related questions motivated this series of investigations.

There are four important large parameters: $m, n, d, t$.  They are  the number of neurons, the number of training samples, the input dimension and the training time, respectively. 
There are two obvious extreme situations that are of interest.  One is when $m \gg n$.  This was described above in Theorem \ref{thm: hop} 
and is relatively well understood.  The other is when $n \gg m$.  This will be investigated next when $n = \infty$.
The regime  when $m\sim n$ or $ m \sim n/(d+1)$ are also of interest since these are regimes where the ``resonance'' (or the closely related
``double descent'')  phenomena might occur, as we  learned from  previous work on RFM \cite{advani2017high, belkin2019reconciling,ma2019gd}. 

Most of our efforts will be devoted to target functions that can be accurately approximated by a relatively small
number of neurons. We will also discuss one example for which this is not the case.
We will come back to this choice of target functions at the end of the paper.

\section{GD dynamics for the case with infinite samples}

As a starting point, we first investigate the GD dynamics for the population risk, i.e. $n = \infty$.
We will see later the phenomena revealed here is indicative of the neural network-like (NN-like) behavior for the GD dynamics.~\footnote{ More  animations of the numerical results in this section can be found in the link \url{https://github.com/TheoreticalML/GD.quenching_activation}.}

Consider the target function $f^*(\bx)=\EE_{\bb\sim\pi^*}[a^*(\bb)\sigma(\bb^T\bx)]$ with $\pi^*$ being a probability distribution over $\SS^{d-1}$.
The population risk can be written as: 
\begin{align}
\nonumber    \cR(\ba,\bB) &= \EE_{\bx}[(\sum_{j=1}^m a_j \sigma(\bb_j\cdot \bx) - \EE_{\bb\sim\pi^*} [a^*(\bb)\sigma(\bb^*\cdot \bx)])^2]\\
\nonumber    &= \sum_{j_1,j_2=1}^m a_{j_1} a_{j_2} k(\bb_{j_1},\bb_{j_2}) - 2 \sum_{j=1}^m  a_j\EE_{\bb\sim\pi^*}[ a^*(\bb) k(\bb_j,\bb)] \\
    &\quad\quad + \EE_{\bb\sim\pi^*}\EE_{\bb'\sim\pi^*}[ a^*(\bb)a^*(\bb') k(\bb, \bb')],
\end{align}
where 
\begin{equation}\label{eqn: ker}
   k(\bb,\bb'):=\EE_{\bx}[\sigma(\bb\cdot\bx)\sigma(\bb'\cdot\bx)] =\|\bb\|\|\bb'\| \left(\sin\theta + (\pi-\theta) \cos \theta \right),
\end{equation}
with $\theta = \arccos(\langle\hat{\bb},\hat{\bb}' \rangle)$.


\subsection{Single neuron target function: Neuron activation and deactivation}
\label{sec: single-neuron-inifite-data}

\begin{figure}
    \centering
    \begin{subfigure}{0.4\textwidth}
    \includegraphics[width=\textwidth]{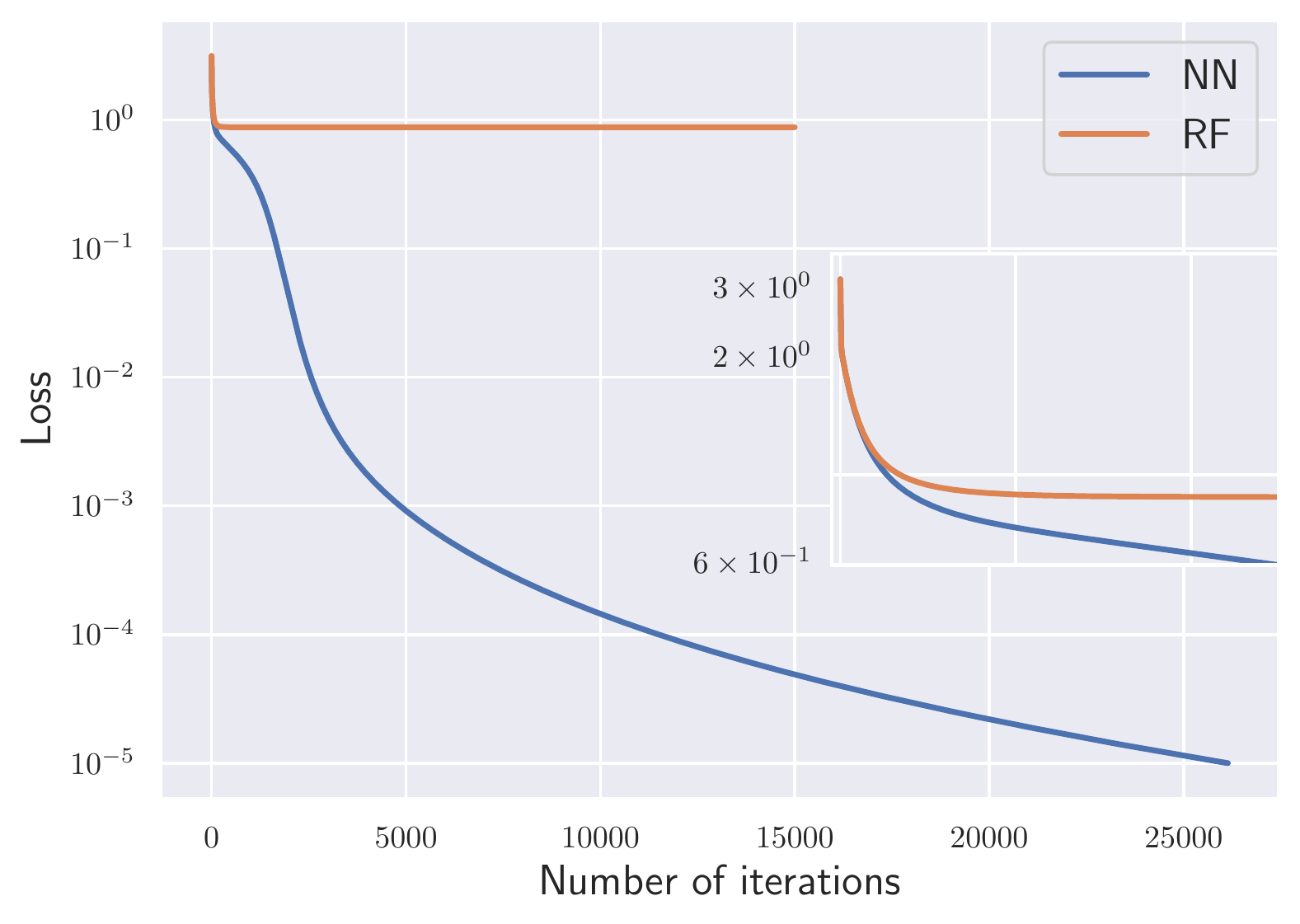}
    \caption{}
    \label{}
    \end{subfigure}
    \begin{subfigure}{0.4\textwidth}
    \includegraphics[width=\textwidth]{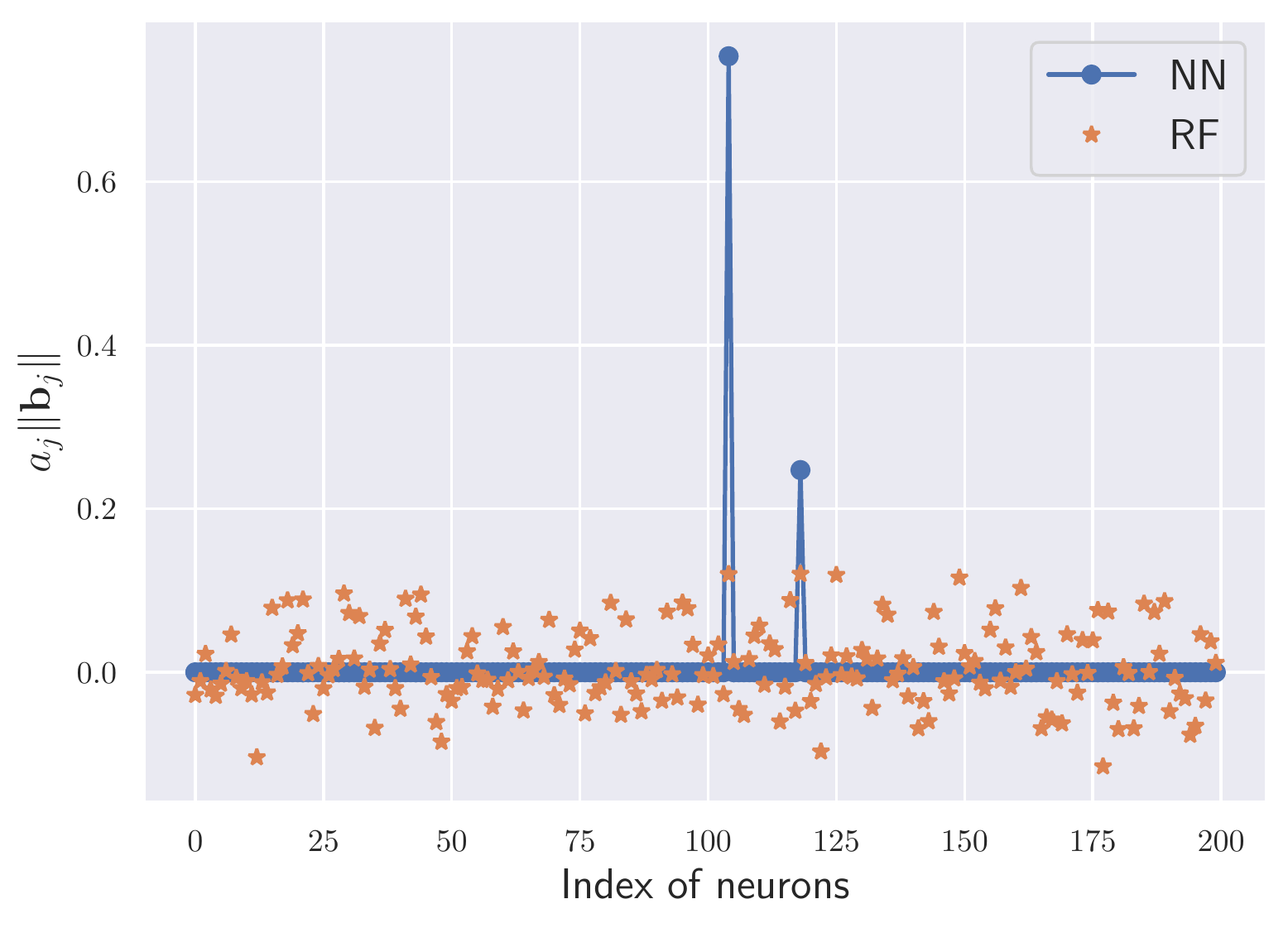}
    \caption{} 
    \label{}
    \end{subfigure}
    \begin{subfigure}{0.4\textwidth}
    \includegraphics[width=\textwidth]{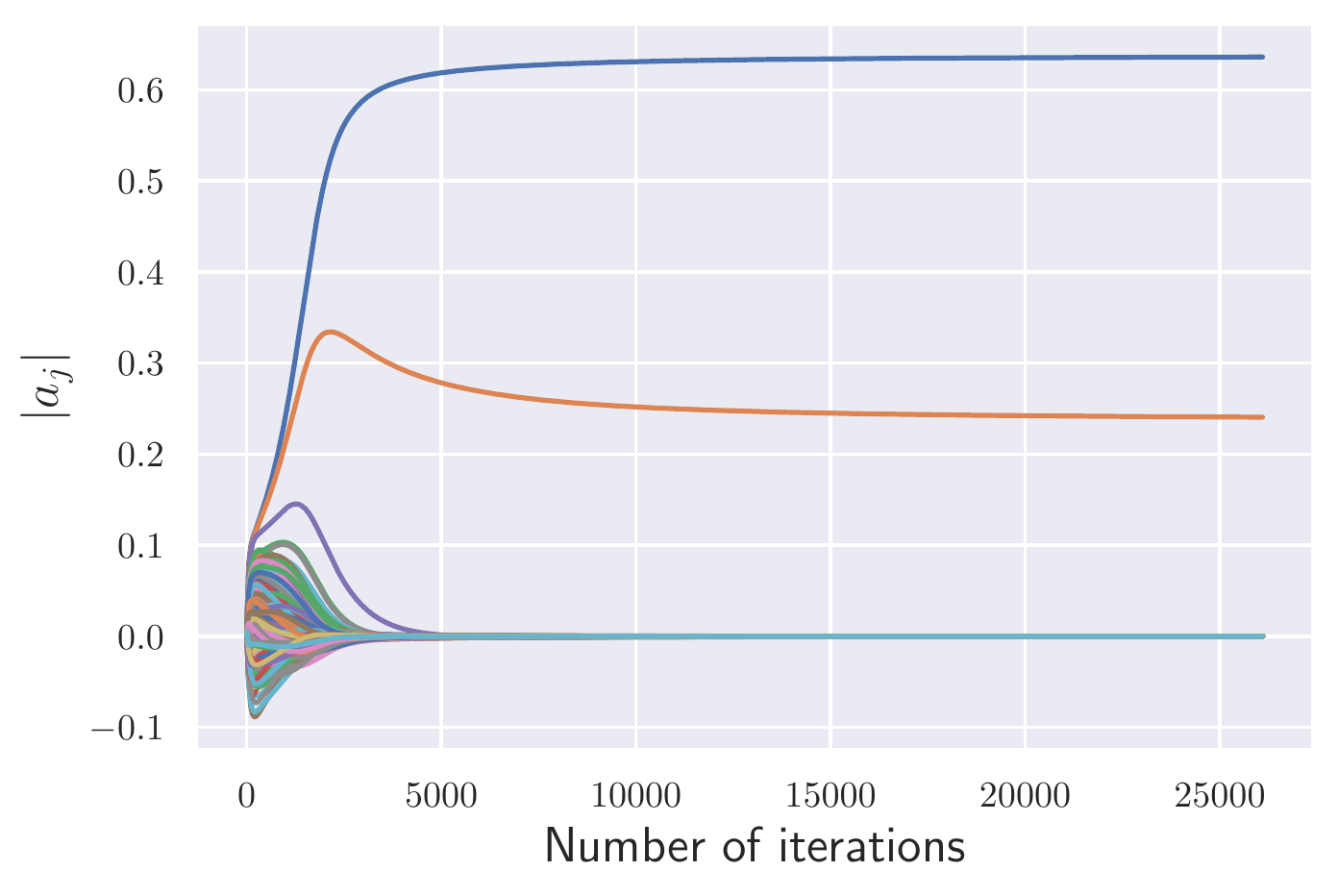}
    \caption{}
    \label{fig: one-1-c}
    \end{subfigure}
    \begin{subfigure}{0.4\textwidth}
    \includegraphics[width=\textwidth]{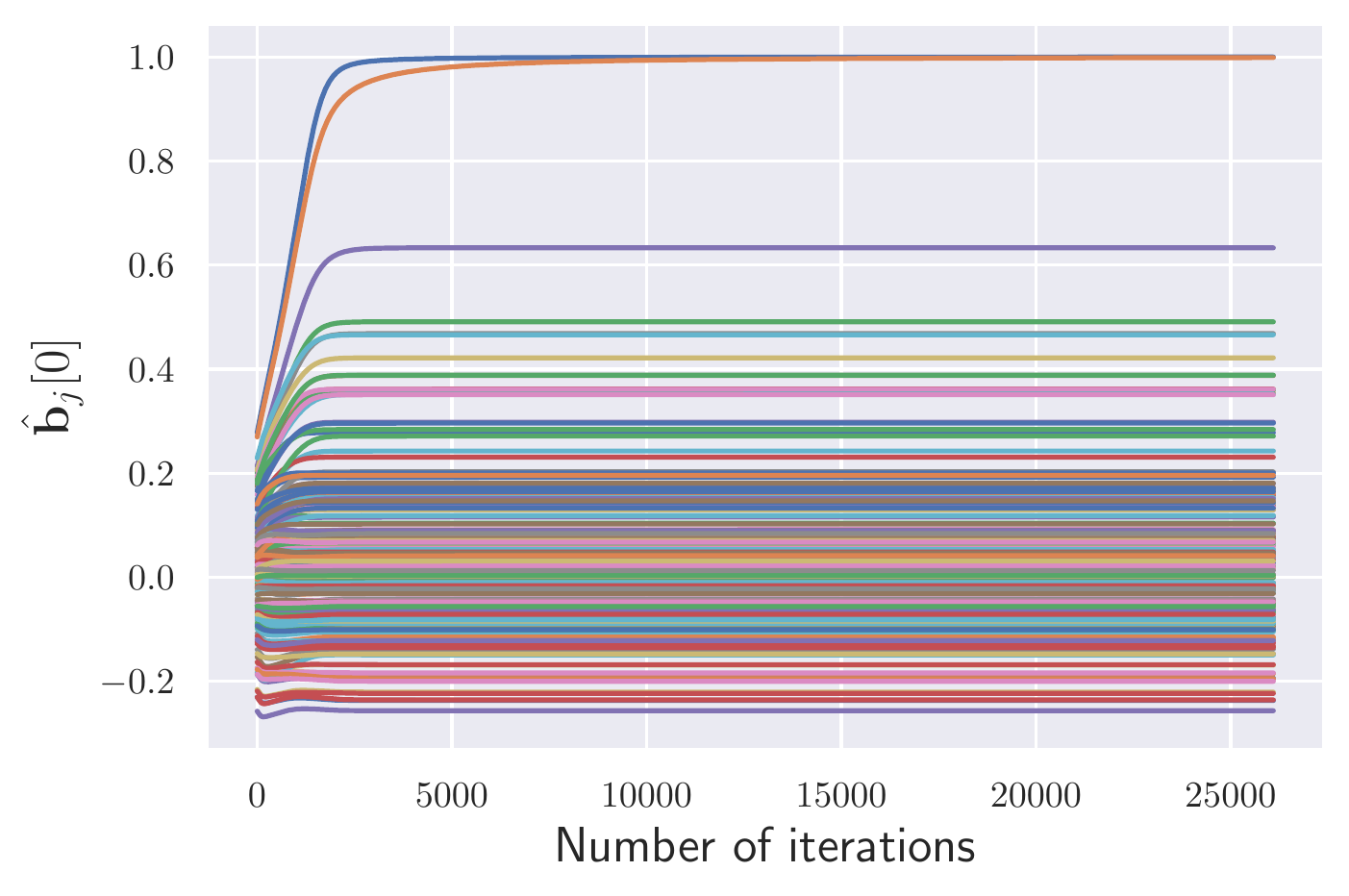}
    \caption{}
    \label{fig: one-1-d}
    \end{subfigure}
    \caption{\small The dynamic behavior of learning single-neuron target function. We also show the results of random feature model as a comparison. In this experiment, $m=200, d=100$ and the learning rate $\eta=10^{-3}$.
    (a) The dynamic behavior of the population risk. (b) The $a$ coefficient of each neuron for the converged solution. 
    (c) The dynamics of the $a$ coefficient of each neuron. (d) The dynamics of $\{\hat{\bb}_j[0]\}$, the projection of $\hat{\bb}_j$ to 
    the line spanned by $\bb^*=\be_1$.
    }
    \label{fig: one-1}
\end{figure}

First we look at the case when the target function is a single neuron: 
$$f_1^*(\bx) = \sigma(\bb^*\cdot \bx),$$ where $\bb^*=\be_1$.
The results are presented in Figure \ref{fig: one-1}.  One can observe several interesting features about the GD dynamics. 

\begin{itemize}
    \item Initially, the GD dynamics for the 2LNN is close to that of  the corresponding RFM.
    \item The GD  dynamics for the 2LNN and RFM depart from each other  around the time when the loss function for the RFM saturates. 
    \item The converged solution for the 2LNN  is very sparse.  In fact  only two neurons contribute significantly to the model in this experiment. 
\end{itemize}

To see what happens in detail, Figure \ref{fig: one-1-c} and \ref{fig: one-1-d} show the dynamics of outer layer coefficients $a$ and the first coordinate of $\hat{\bb}$, which is equal to $\langle \hat{\bb}, \bb^*\rangle$.
One can see that except for two neurons, the outer layer coefficients $a$  decay to 0. 

We also see that there are two phases in the GD dynamics.  
In the first phase,  the dynamics follows closely the GD dynamics for the RFM.  
In the second phase,  the outer layer coefficients $a$ are small except for two neurons.
 In the transition from the first to the second phase, except for two neurons, all the other neurons are
 ``quenched'' in the sense that their outer layer coefficients $a$ keep decreasing, consequently the dynamics of their inner layer
 coefficients $\bb$ become very slow. 
 The same behavior was observed with other  realizations of the initial data, except that the number
 of ``activated'' neurons can be different. But in any case, we always observe
 few ``activated'' neurons with large $a$ coefficients that  dominate the dynamics in the second phase.  
 Some of these activated neurons can also become deactivated.

\subsection{Circle neuron target function: Multi-step phenomenon}

Next we consider a more sophisticated target function:
{
\begin{equation}
f_2^*(\bx) = \EE_{\bb\sim \pi_2}[\sigma(\bb^T\bx)],
\end{equation}
where $\pi_2$ is the uniform distribution over the unit circle 
$
\Gamma=\{\bb: b_1^2+b_2^2=1 \text{ and } b_i=0\,\,\forall i> 2\}.
$
}

\begin{figure}[!h]
    \centering
    \begin{subfigure}[b]{0.32\textwidth}
        \includegraphics[width=\textwidth]{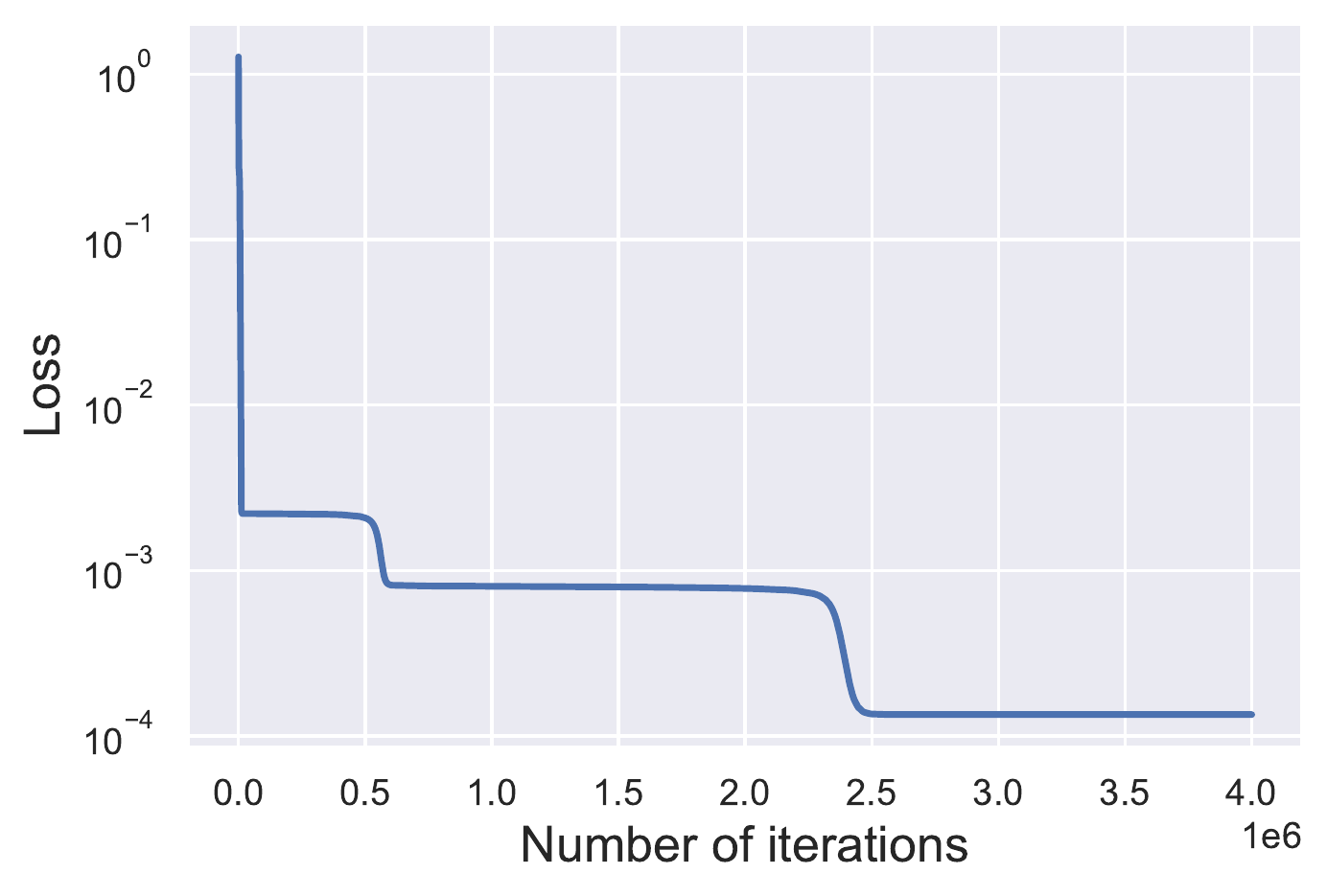}
        \caption{}
        \label{fig: circle-1a}
    \end{subfigure}
    \begin{subfigure}[b]{0.32\textwidth}
        \includegraphics[width=\textwidth]{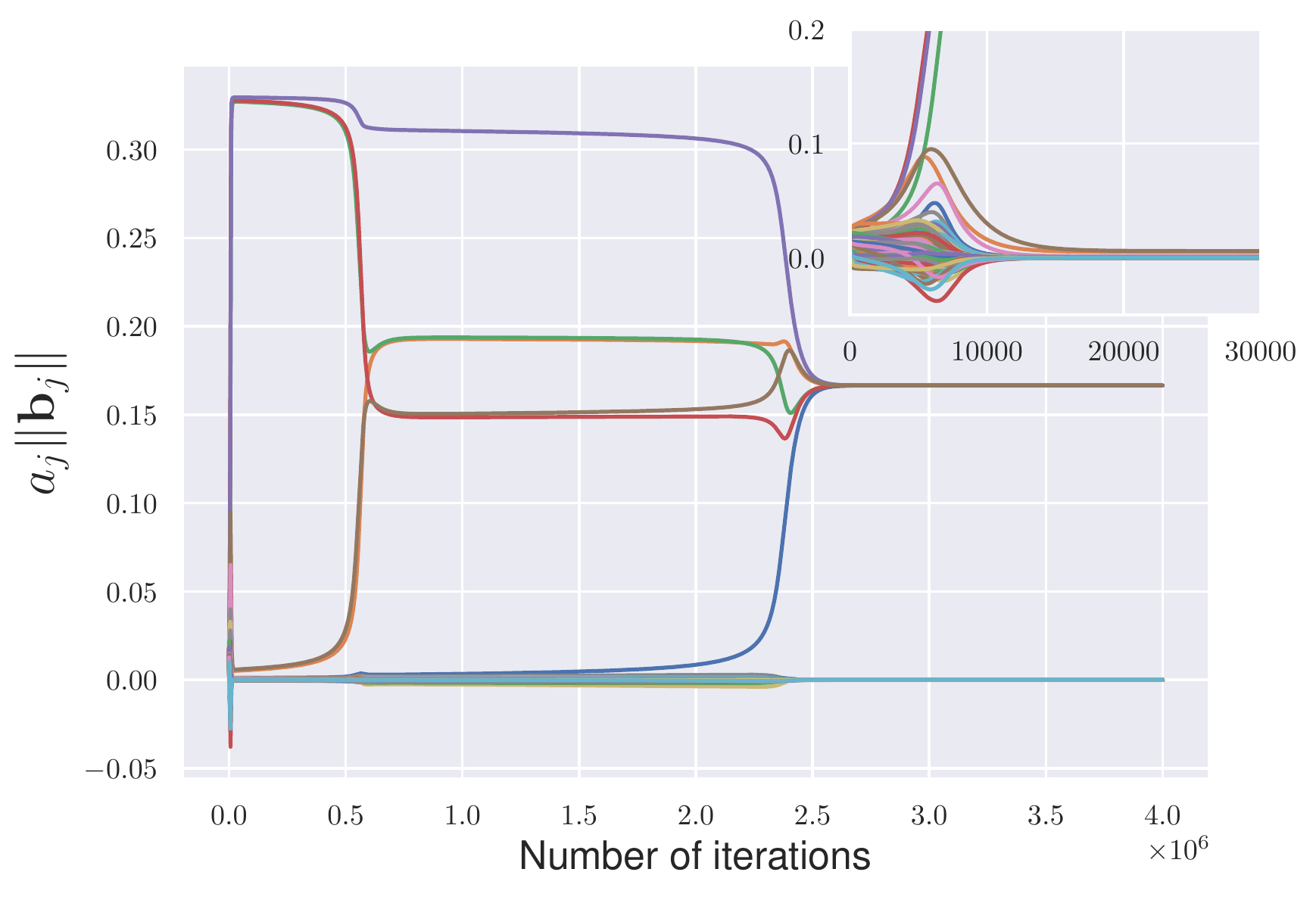}
        \caption{}
        \label{fig: circle-1b}
    \end{subfigure}
    \begin{subfigure}[b]{0.32\textwidth}
        \includegraphics[width=\textwidth]{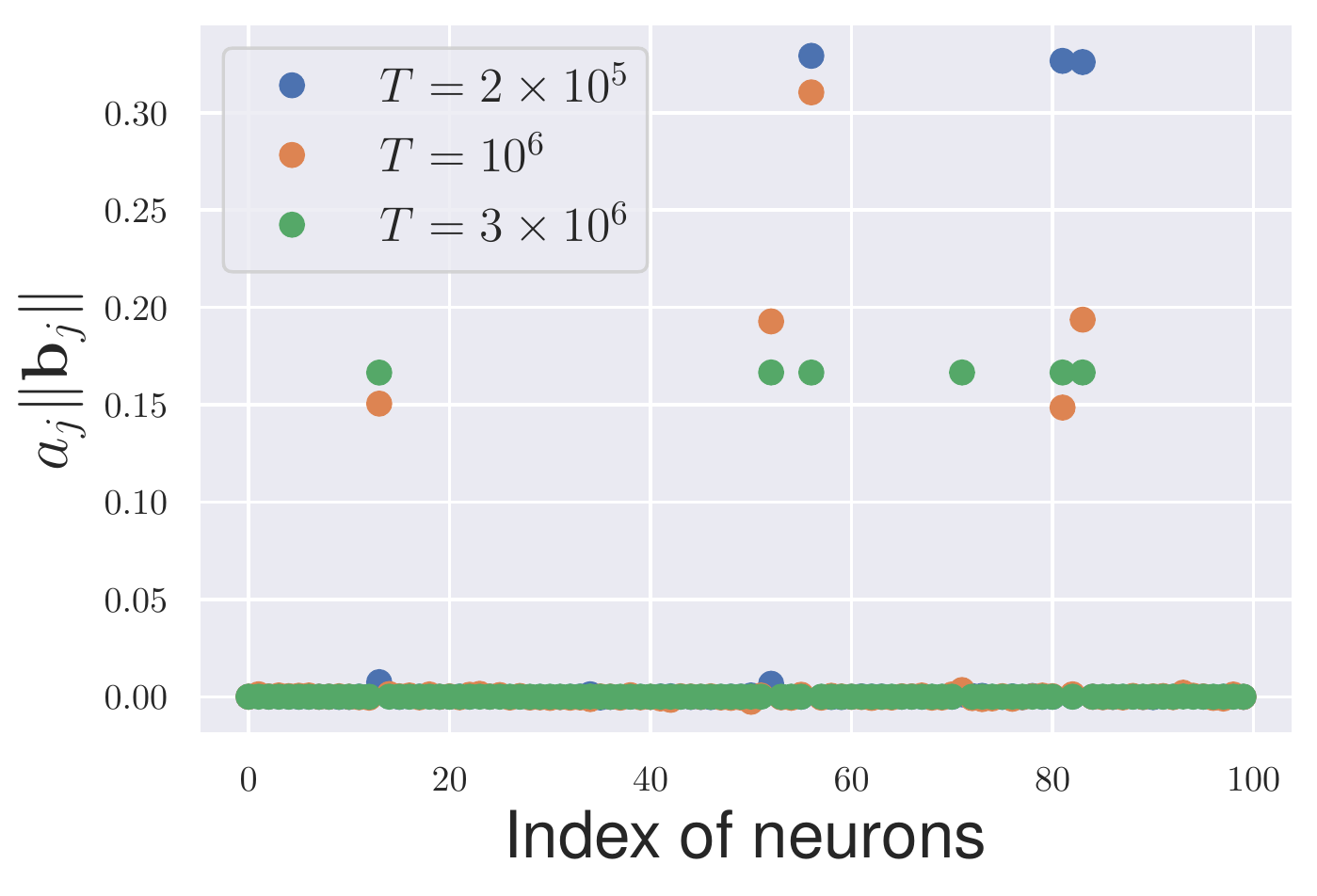}
        \caption{}
        \label{fig: circle-1c}
    \end{subfigure}
    \caption{\small The dynamic behavior of GD solutions for the circle neuron target function. Here $m=100, d=100$  learning rate $\eta=0.005$. (a) The dynamic behavior of the population risk. (b) The dynamics of the $a$ coefficient of each neuron,  the inset is the zoom-in of the first $30000$ iterations. (c) The $a$ coefficients of the solutions selected to represent the three steps. }
    \label{fig: circle}
\end{figure}

A typical dynamic behavior of population risk is shown in Figure \ref{fig: circle-1a}. 
We see that  there are still two phases. But in the second phase, the population risk decreases in a ``step-like'' fashion. To see what happens, we plot the dynamics of the $a$ coefficient of each neuron in Figure \ref{fig: circle-1b}. We see that  after the first phase, most of the neurons start to die out slowly  (see the inset of Figure \ref{fig: circle-1b}). 
As the GD dynamics proceeds,  a few new neurons are activated from the ``background neurons''. This activation process 
can be very slow, and the loss function is almost  constant before activation actually happens.
 The activation process is relatively fast and causes a fast decay of the loss function.
  Figure \ref{fig: circle-1c} shows three representative solutions for the three steps.  
  We see that from the first and second step,  two more neurons are activated. 
  From the second to the third step,  one more neuron pops out. 

\subsubsection{Detail analyses of the two phases}
\begin{figure}[!h]
    \centering
    \begin{subfigure}{0.4\textwidth}
    \includegraphics[width=\textwidth]{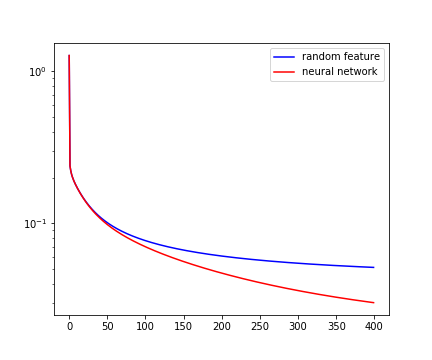}
    \caption{}
    \label{}
    \end{subfigure}
    \begin{subfigure}{0.4\textwidth}
    \includegraphics[width=\textwidth]{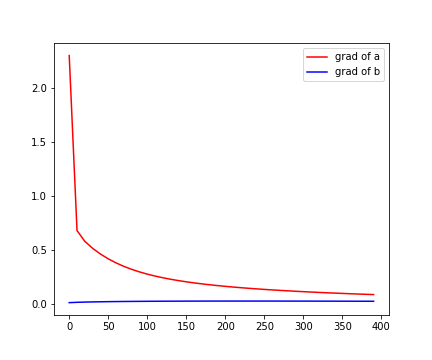}
    \caption{}
    \label{}
    \end{subfigure}
    \caption{(a) The population risk of the neural network and random feature models are close at early times.
    (b) The magnitude of the speed of  $a$ is much larger than that of $\bb$ at early times. }
    \label{fig: phase1-1}
\end{figure}

Next we analyze the dynamics in the two phases in some more detail.
Figure \ref{fig: phase1-1} compares the dynamics of the population risk and the speed of $a$ and $\bb$ during the first phase. During this phase, $\bb$ has hardly changed. So the dynamics is dominated by the dynamics of $a$.  This is consistent with the fact that during this phase, the GD dynamics follows that of the RFM.

\begin{figure}[!h]
    \centering
    \begin{subfigure}{0.4\textwidth}
    \includegraphics[width=\textwidth]{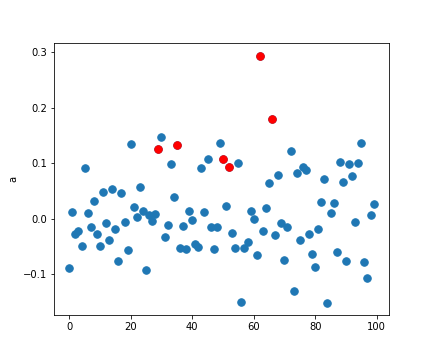}
 \caption{}
 \label{fig:phase1-2}
 \end{subfigure}
 \centering
 \begin{subfigure}{0.4\textwidth}
     \includegraphics[width=\textwidth]{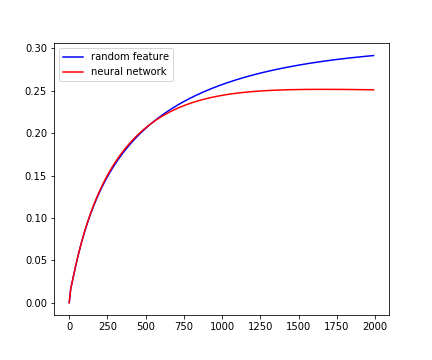}
    \caption{}
    \label{fig:phase2-1}
 \end{subfigure}
  \caption{\small (a) The coefficients of the optimal solution of the random feature model. The activated neurons of the neural network are marked red.
   These neurons roughly correspond to the ones with relatively large coefficients in the solution of the RFM. 
  (b) The largest coefficients of the NN and RF models. The two models depart from each other when some outer layer coefficients become large.}
 \label{fig: phase-xx}
\end{figure}

If the first phase continues indefinitely, then eventually the solutions will converge to that of the linear system $\nabla_{\ba} \cR(\ba,\bB_0)=0$, which can be explicitly written as  
\begin{equation}\label{eqn: rf-solution}
    \EE_{\bx}[\sigma(\bB_0\bx)\sigma(\bB_0\bx)^T] \ba = \EE_{\bx}[f^*(\bx)\sigma(\bB_0\bx)].
\end{equation}
For the present problem, a typical solution to \eqref{eqn: rf-solution}  is plotted in Figure \ref{fig:phase1-2}.
One can see that this solution has some large components. 
These large (outer layer) coefficients induce large changes for the corresponding inner layer parameters.
Once this happens, the mechanism for the GD dynamics to stay close to that of the RFM is no longer valid and
one should expect that the two dynamics depart from each other.  This is indeed what happens, as can be seen
from Figures \ref{fig:phase1-2} and \ref{fig:phase2-1}.

In the second phase, the neurons can be divided into two group:  the active neurons $\{(\tilde{a}_i,\ \tilde{\bb}_i)\}_{i\in A}$ and the background neurons $\{(\hat{a}_j,\ \hat{\bb}_j)\}_{j\in B}$, where  $A$ and $B$ are the set of indices for the active and background neurons, respectively.
For the active neurons,  their outer and inner layer coefficients change at the same time scale.
For the background neurons, the fact that their outer layer coefficients are small  implies that the dynamics of their outer layer
coefficients is much faster than their inner layer coefficients. Therefore we expect that their outer layer coefficients are approximately
at equilibrium with their inner layer coefficients. Let $f_A(\bx;t)=\sum_{j\in A} a_j(t) \sigma(\bb_j^T(t)\bx)$ and $f_B(\bx;t)=\sum_{j\in B}a_j(t) \sigma(\bb_j^T(t)\bx)$ be the functions represented by the active neurons and the background neurons, respectively. Then, the previous argument suggests that $(a_j(t))_{j\in B}$ is approximately the solution of the following problem:
\begin{equation}\label{eqn: eff-objective}
\min_{\{a_j\}_{j\in B} } \EE_{\bx} \big(\sum_{j\in B} a_j \sigma(\bb_j(t)^T\bx) + f_A(\bx;t)- f^*(\bx) \big)^2 
\end{equation}
We arrive at  the following effective dynamics for the original system:
For active neurons, i.e.  $j\in A$, 
\begin{align*}
\dot{a}_j &= \int (f^*(\bx)-f_A(\bx;t)-f^*_B(\bx;t))\sigma(\bb_j^T\bx)\mu(d\bx), \\
\dot{\bb}_j &= \int (f^*(\bx)-f_A(\bx;t)-f^*_B(\bx;t))\tilde{a}_i\sigma'(\tilde{\bb}_i^T\bx)\bx\mu(d\bx),
\end{align*}
For background neurons, i.e. $j\in B$, 
\begin{align*}
a_j(t) &= a_j^*(t),\\
    \dot{\bb}_j(t) &= \int (f^*(\bx)-f_A(\bx;t) - f_B^*(\bx;t))a_j^*(t) \sigma'(\bb_j^T\bx)\bx \pi_0(d\bx).
\end{align*}
Here $\{\ba_j^*(t)\}_{j\in B}$ is the solution of the problem \eqref{eqn: eff-objective}, and $f_B^*(\bx;t) = \sum_{j\in B} a_j^*(t)\sigma(\bb_j(t)^T\bx)$.

\begin{figure}
    \centering
    \begin{subfigure}{0.4\textwidth}
    \includegraphics[width=\textwidth]{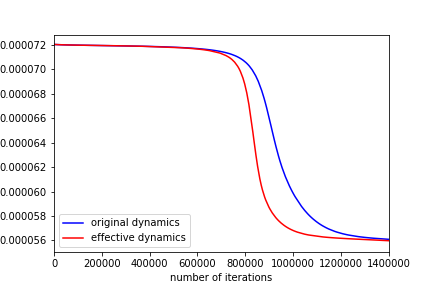}
    \caption{} 
    \label{}
    \end{subfigure}
    \begin{subfigure}{0.4\textwidth}
    \includegraphics[width=\textwidth]{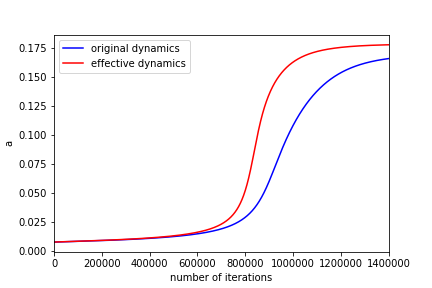}
    \caption{}
    \label{}
    \end{subfigure}
    \caption{\small Comparison of the effective dynamics and the original dynamics. 
    (a) Loss function; (b) The value of the $a$ coefficient of the neuron that jumps from the background group to the activated group during the process.}
    \label{fig:effective}
\end{figure}

Figure \ref{fig:effective} presents the results for the comparison between the effective dynamics and the actual dynamics.
One can see that the effective dynamics agrees well with the actual dynamics except when new active neurons pop out. 

\subsection{ Finite neuron target functions}

\begin{figure}
\centering 
\begin{subfigure}{0.8\textwidth}
\centering
\includegraphics[width=0.45\textwidth]{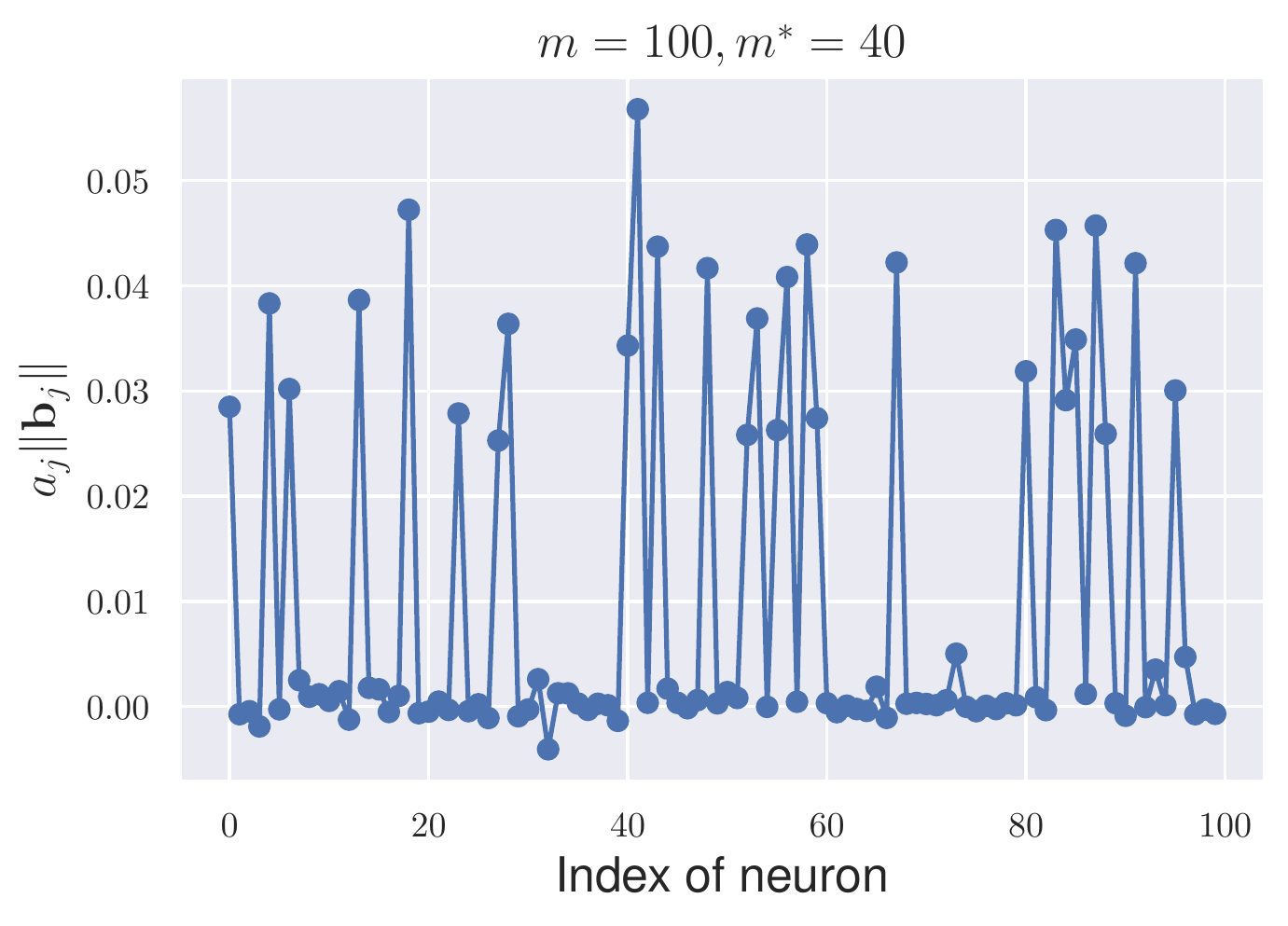}
\includegraphics[width=0.45\textwidth]{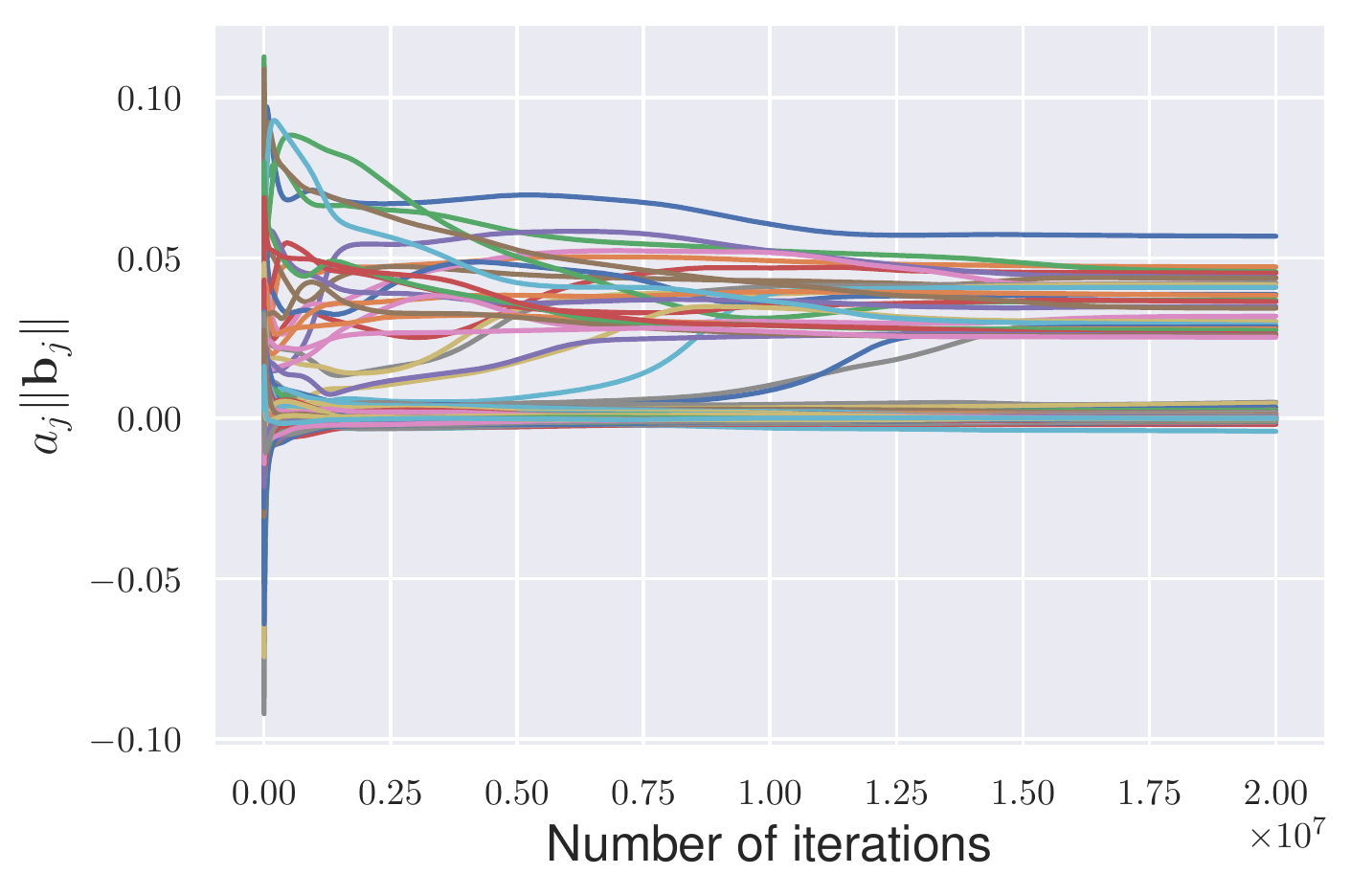}
\caption{$m=100, m^*=40$}\label{}
\end{subfigure}
\begin{subfigure}{0.8\textwidth}
\centering
\includegraphics[width=0.45\textwidth]{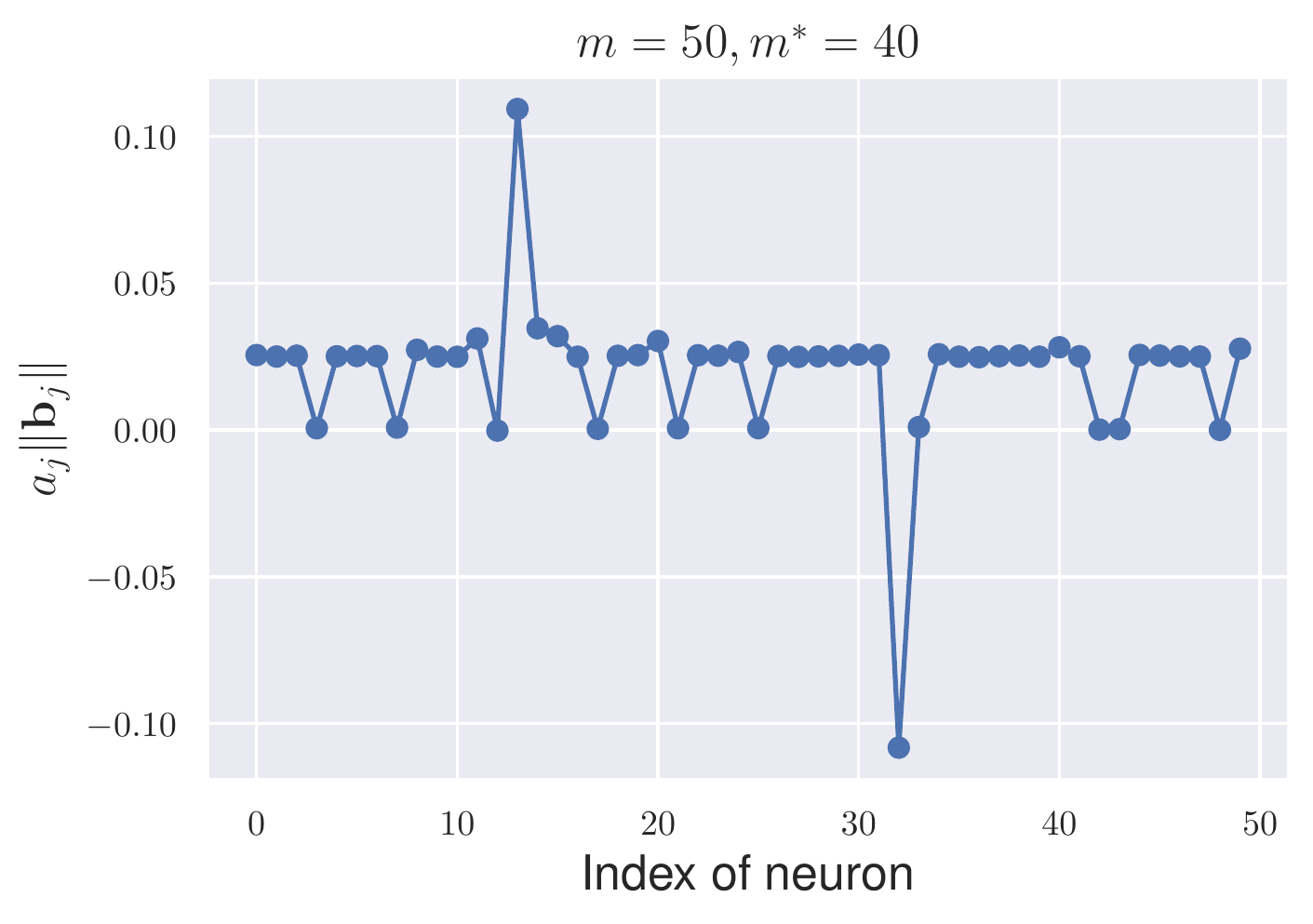}
\includegraphics[width=0.45\textwidth]{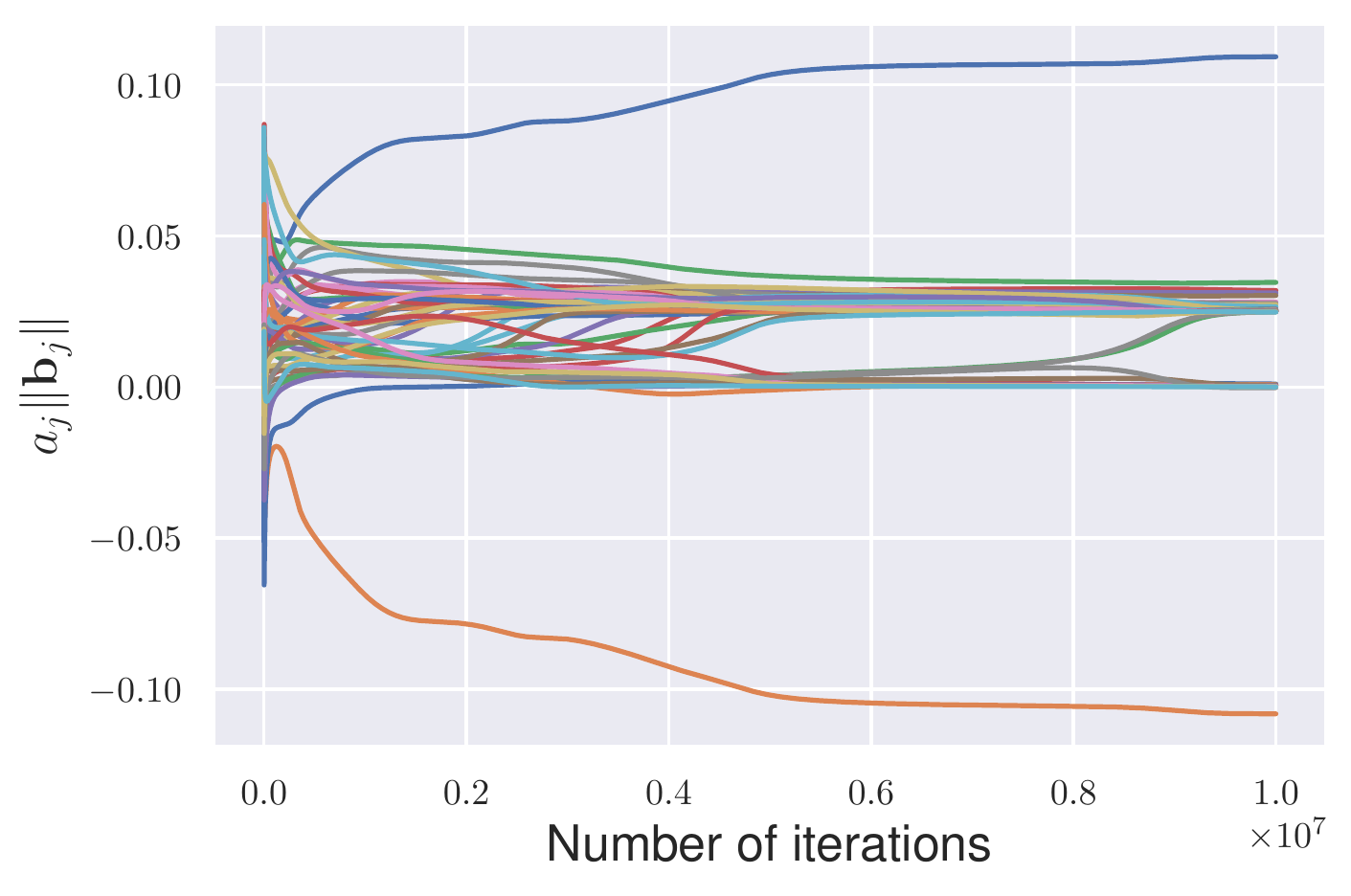}
\caption{$m=50, m^*=40$}\label{}
\end{subfigure}
\caption{ \small Learning finite neurons with $a_j^*=1/m, \bb_j^*\sim \pi_0$. (Left) The magnitude of each neuron of the final solution. (Right) The dynamics of the magnitude for each neuron.}
\label{fig: finite-neurons}
\end{figure}

The observation that GD dynamics picks out a sparse solution and the associated quenching-activation behavior happens in a more general setting: learning finite neurons. 
Consider the target function 
\[
    f^*(\bx) = \sum_{j=1}^{m^*} a_j^* \sigma(\bb_j^*\cdot \bx).
\]
We are interested in the case $m>m^*$.
The single-neuron target function is a special case with $m^*=1$.  Figure \ref{fig: finite-neurons} shows the dynamic behavior for $m=50,100$ and $m^*=40$.  We see the GD dynamics tends to find solutions with the number of active neurons close to $m^*$. The learning process is qualitatively similar to the case of  circle neuron target function
except that the activation process proceeds in a more continuous fashion and therefore the step-like behavior is less
pronounced.

\subsection{Surface neuron target function}
\begin{figure}[!h]
    \centering
    \begin{subfigure}[b]{0.4\textwidth}
        \includegraphics[width=\textwidth]{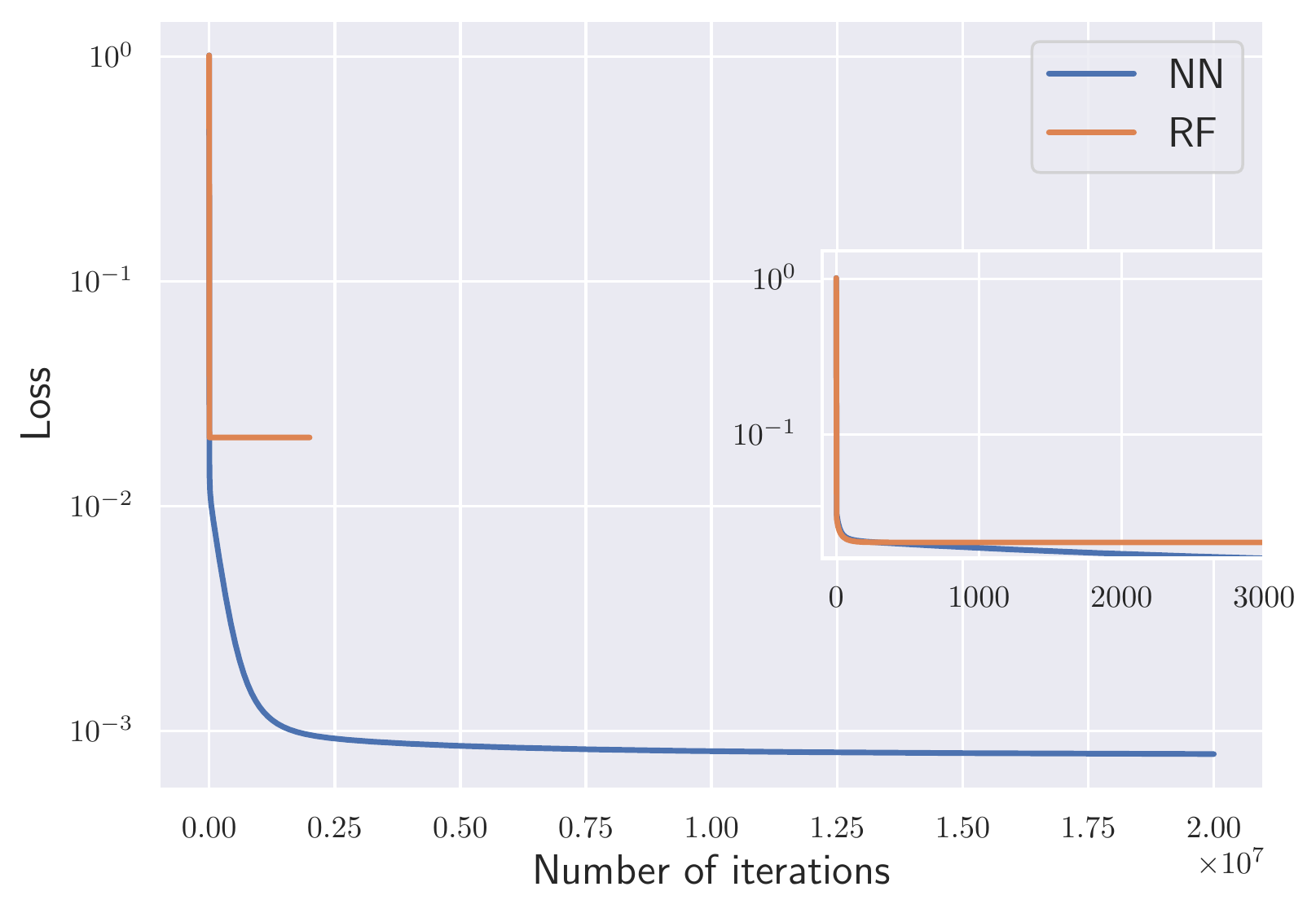}
        \caption{}
        \label{fig: surface-1a}
    \end{subfigure}
    \hspace{0.02\textwidth}
    \centering
    \begin{subfigure}[b]{0.4\textwidth}
        \includegraphics[width=\textwidth]{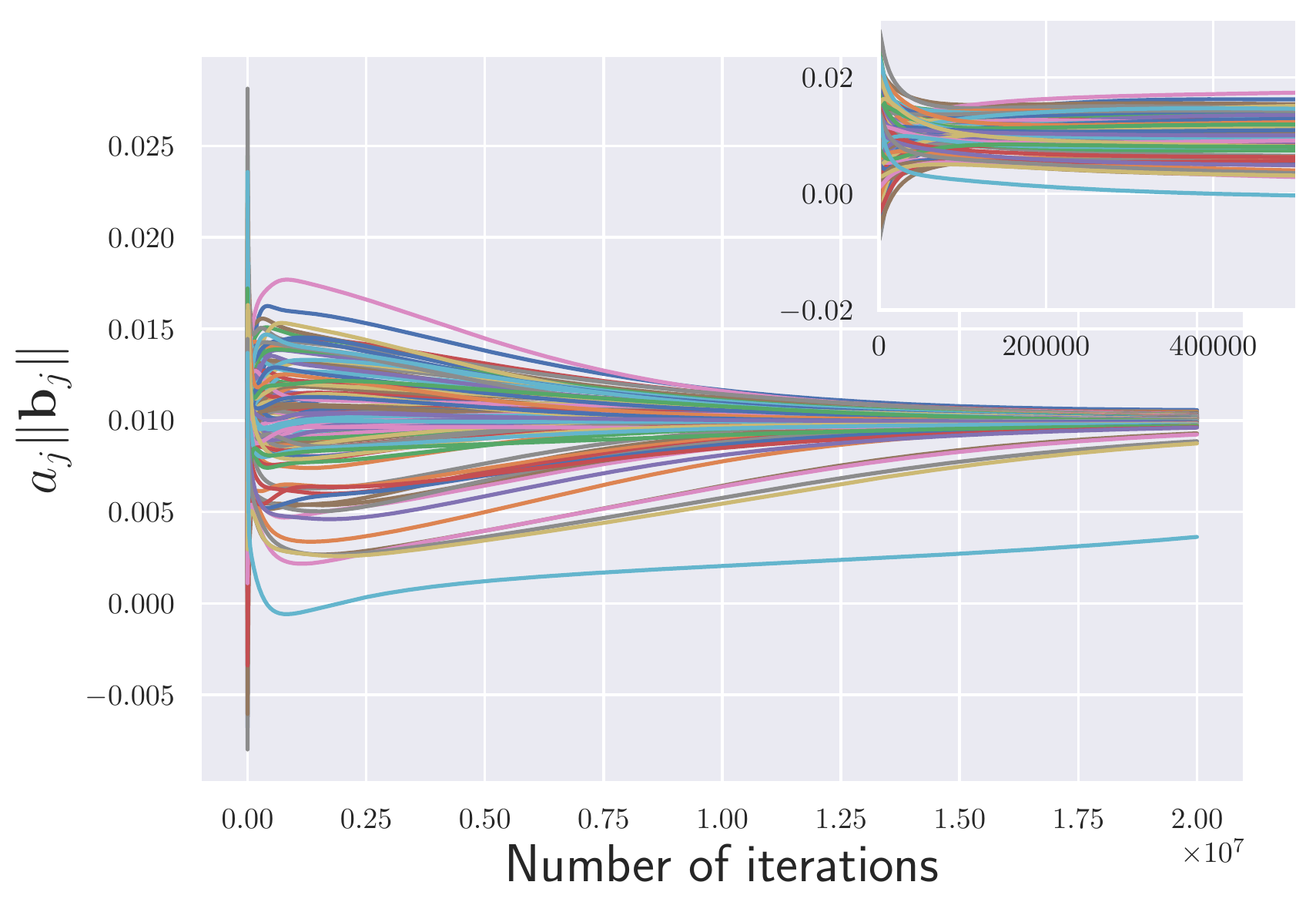}
        \caption{}
        \label{fig: surface-1b}
    \end{subfigure}
    \caption{\small The dynamic behavior of the GD solutions for the surface neuron target function. Here $m=100, d=100$ and learning rate $\eta=0.005$. (a) The dynamic behavior of the population risk; (b) The dynamics  of 
    the outer layer coefficients for each neuron. The inset is the zoom-in of the first $500,000$ iterations.}
    \label{fig: surface}
\end{figure}

The target functions studied above are all functions that can be accurately approximated by a small
set of neurons.  Next we consider an example in the opposite direction,
the ``surface neuron target function'':
{
\[
f_4^*(\bx) = \EE_{\bb\sim\pi_3}[\sigma(\bb^T\bx)]
\]
where $\pi_3$ is the uniform distribution over $\Omega=\{\,\bb: \sum_{i=1}^{d/2} b_i^2=1 \text{ and } b_i=0,\, \forall i>d/2\}$. }
This function is represented by a very large set of neurons, each of which contributes an equal small amount.
Shown in Figure \ref{fig: surface} are  the numerical results. 
There are clear similarities and differences with the previous examples.
On similarities,  that there  are still two phases and there is still some kind quenching process going on. 
On differences, one does not observe the decomposition into active and background neurons, and the activation process
is replaced by smooth changes for all the neurons.


\section{GD dynamics for the case of finite training samples}

We now turn to the more realistic situation when the size of the training set is finite.

\subsection{The highly over-parameterized regime}
\begin{figure}
    \centering
    \begin{subfigure}{0.4\textwidth}
    \includegraphics[width=\textwidth]{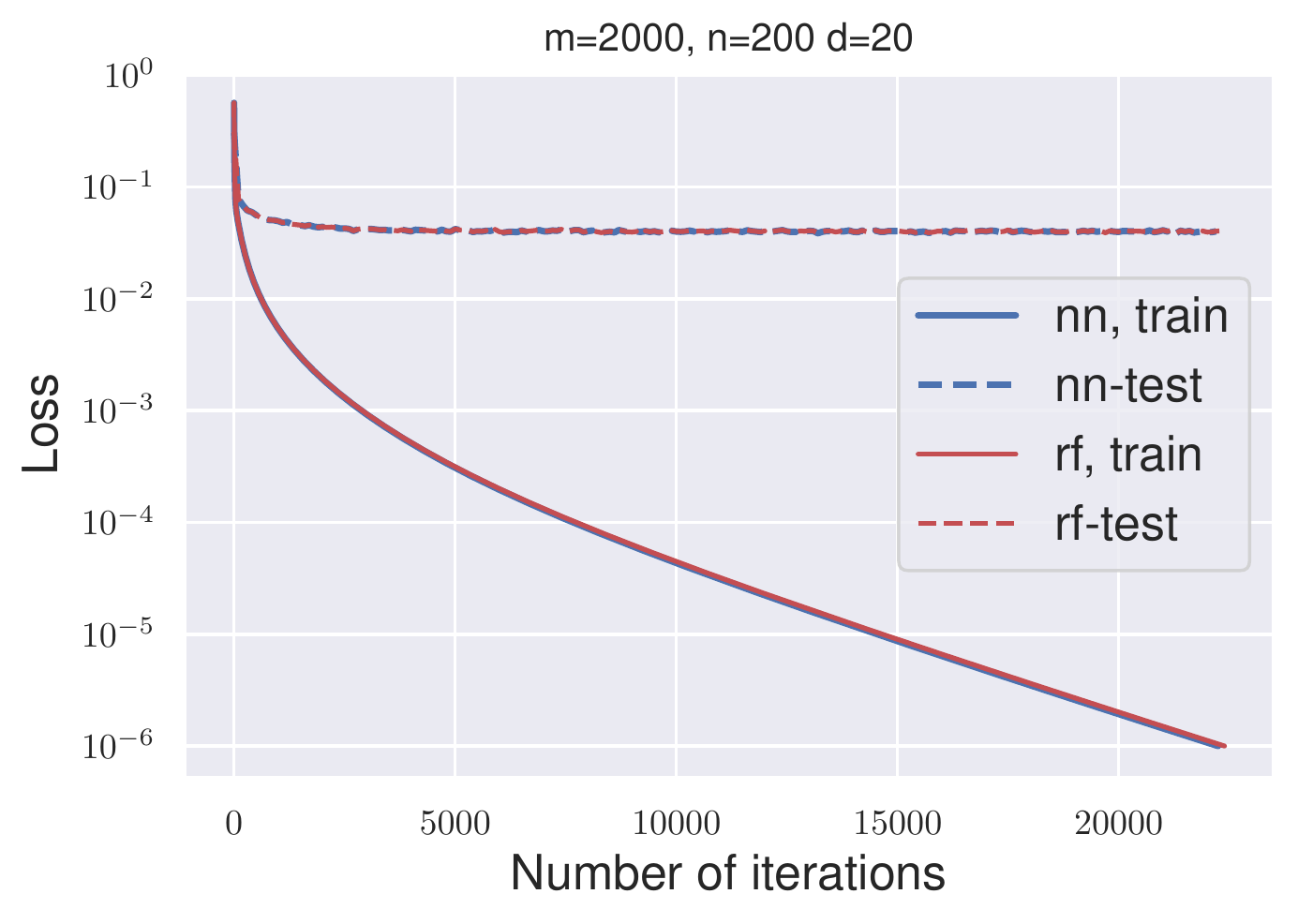}
    \caption{}\label{}
    \end{subfigure}
    \begin{subfigure}{0.4\textwidth}
    \includegraphics[width=\textwidth]{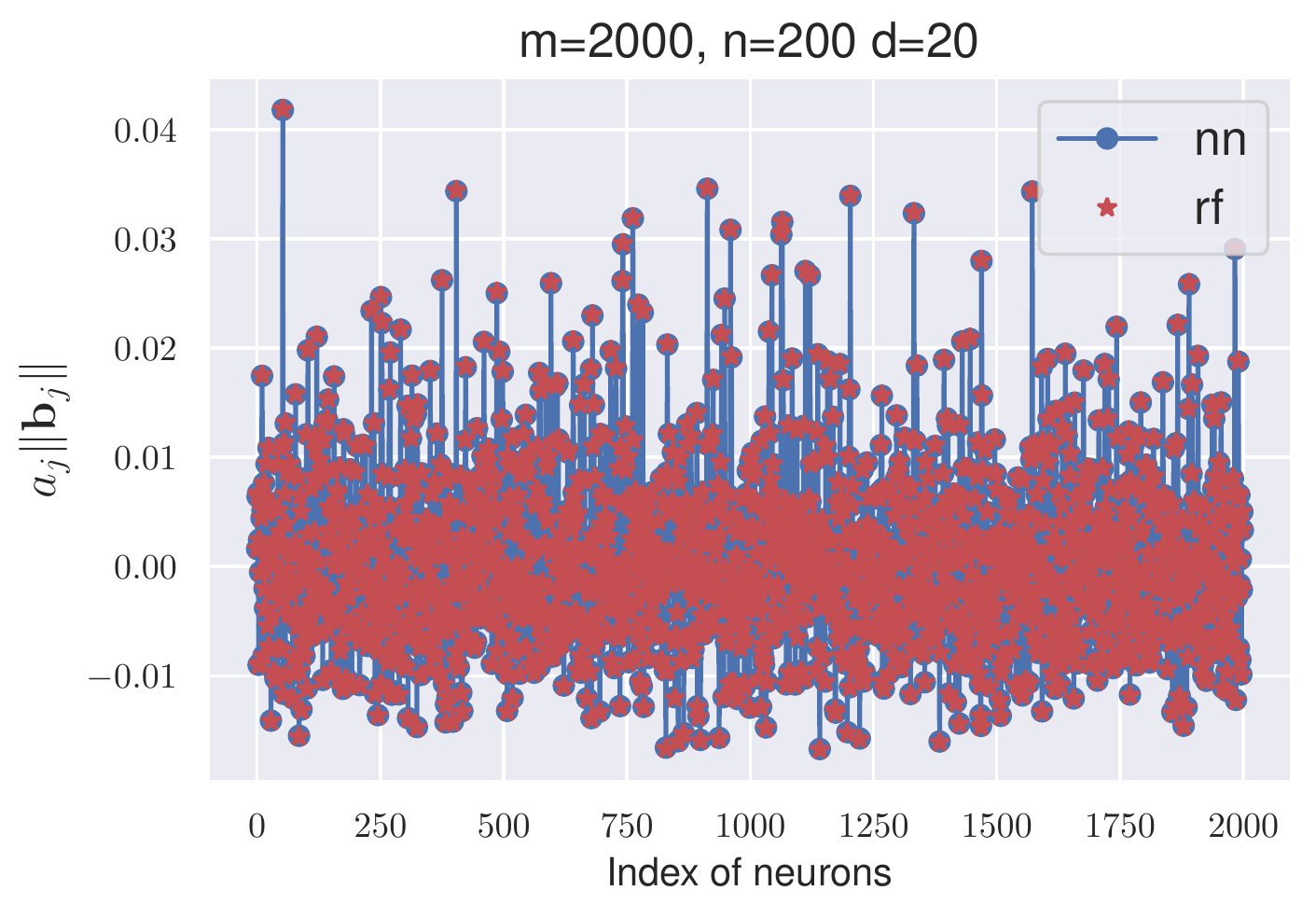}
    \caption{}\label{}
    \end{subfigure}
    \caption{\small Comparison between the GD solutions for 2LNN and RFM for the single neuron target function $f_1^*$. The learning rate $0.001$ and $m=2000, n=200, d=20$. (a) The time history  of the  training and test error. (b) The outer layer coefficient of the converged solutions. }
    \label{fig:HOP-1}
\end{figure}

We first look at the simple situation when the network is highly over-parametrized. Figure \ref{fig:HOP-1} shows some typical results in this regime. Clearly in this regime, the GD dynamics for the NN model stays
uniformly close to that of the RFM for all time, as was proved in \cite{ma2019comparative}. We see that the training error goes to 0 exponentially fast, but the testing error quickly saturates.  Note that Theorem \ref{thm: hop} suggests that the network width should satisfy $m\gtrsim n^2\lambda_n^{-4} \ln(n^2\delta^{-1})$, but in this experiment $m=0.5n^2$ is enough for the single neuron target function.

\subsection{The under-parametrized regime}

Next we look at the regime when the network is under-parametrized, i.e. $m < n/(d+1)$.

\begin{figure}[!h]
    \centering
    \begin{subfigure}{0.32\textwidth}
    \includegraphics[width=\textwidth]{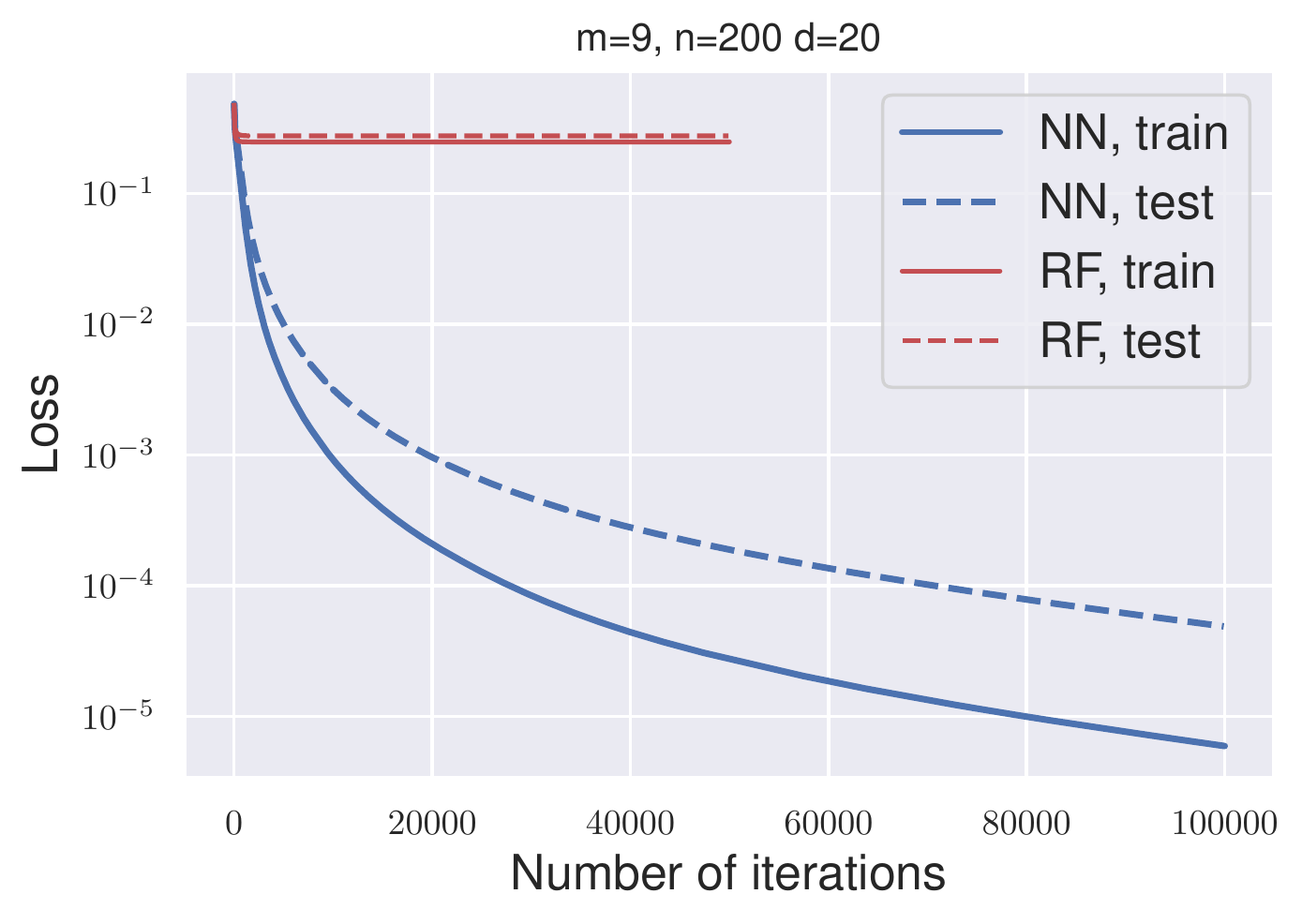}
    \caption{}\label{fig: up-1a}
    \end{subfigure}
    \begin{subfigure}{0.32\textwidth}
    \includegraphics[width=\textwidth]{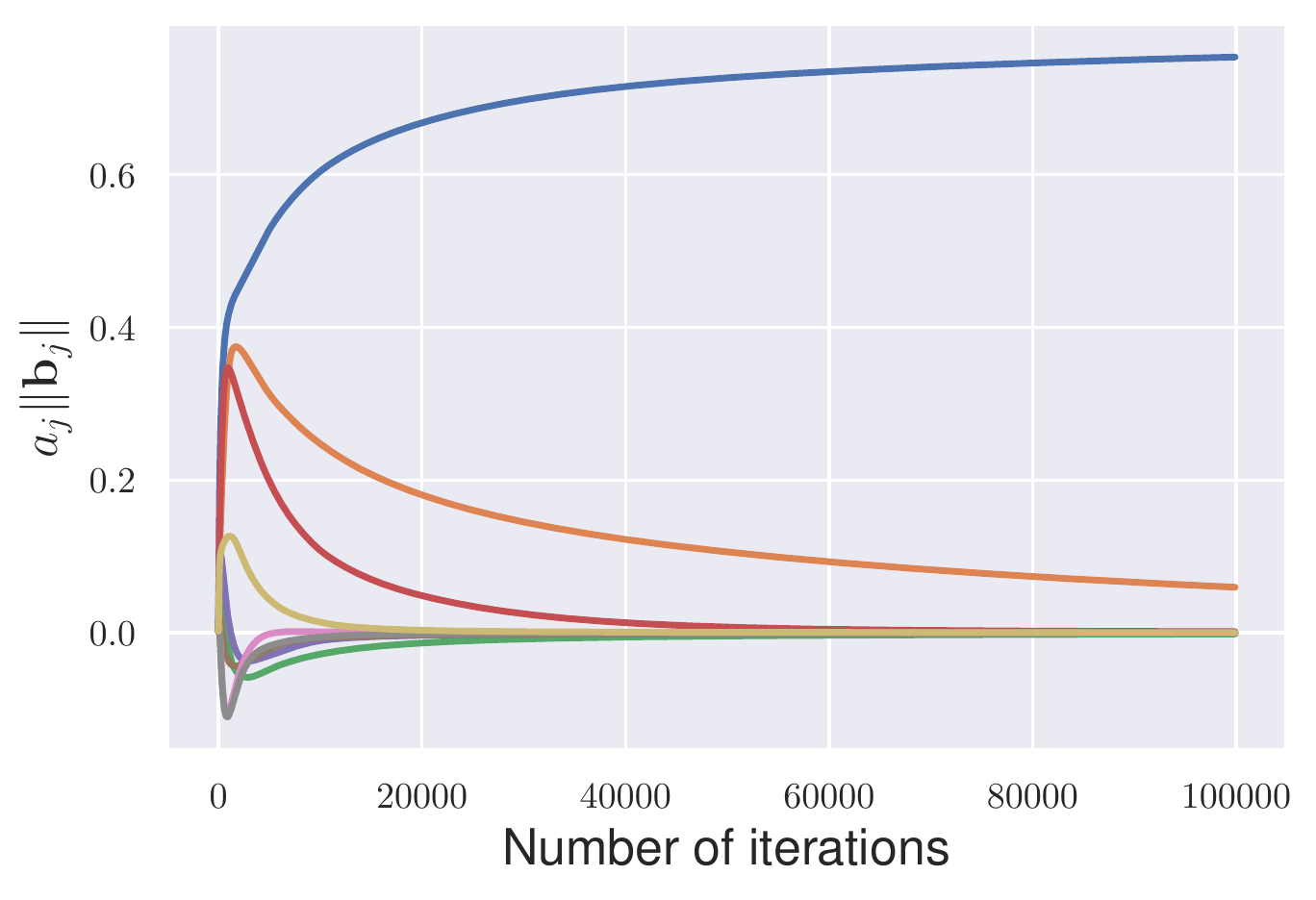}
    \caption{}\label{fig: up-1b}
    \end{subfigure}
    \begin{subfigure}{0.32\textwidth}
    \includegraphics[width=\textwidth]{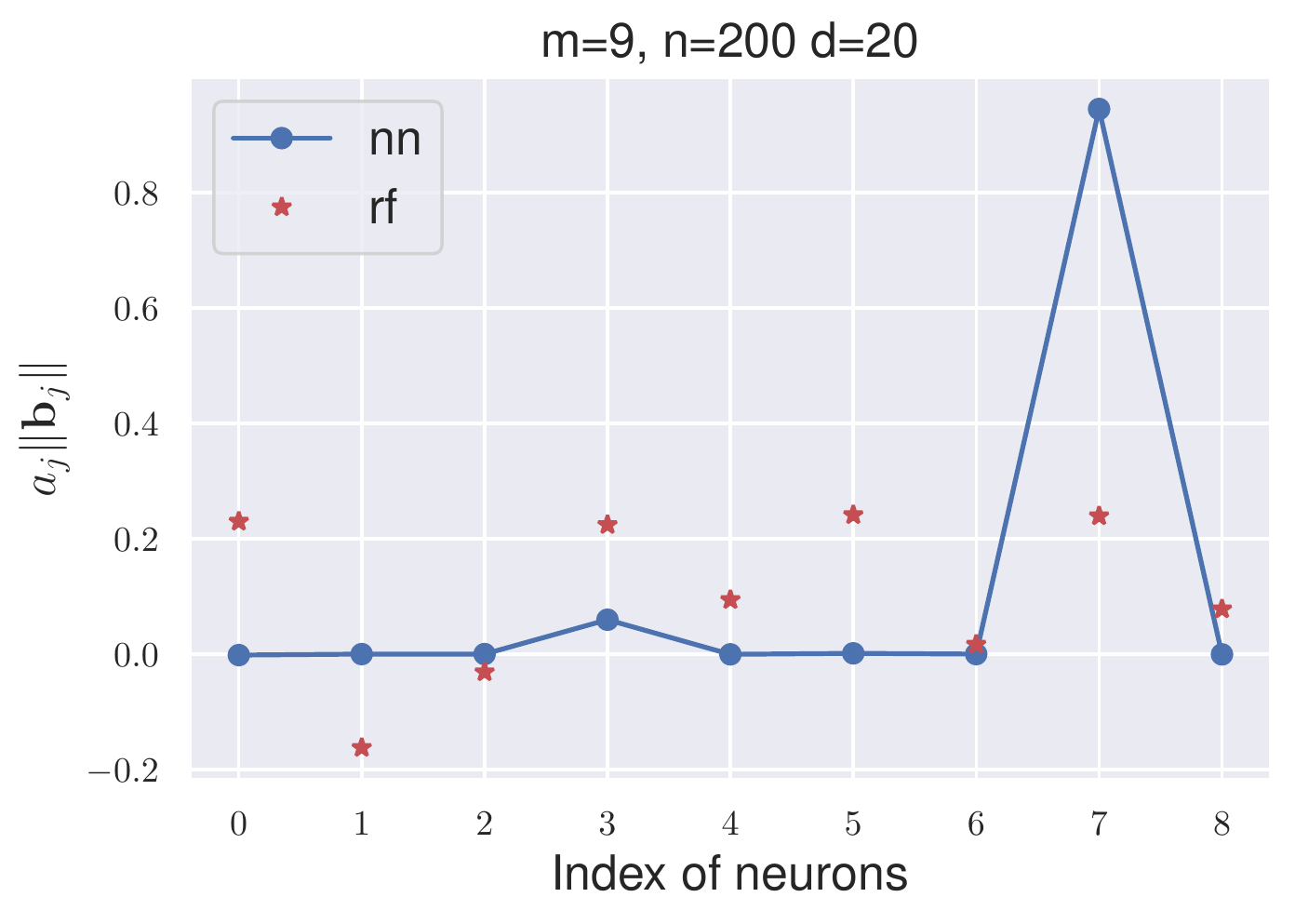}
    \caption{}\label{fig: up-1c}
    \end{subfigure}
    \caption{\small  GD dynamics for the single neuron target function for the under-parametrized case. Here $m=9, d=20, n=200$ and the learning rate $\eta=0.005$.
    (a) Time history of training and test errors. (b) Time history of  the outer layer coefficient for each neuron. 
    (c) The convergent solutions.
    }
    \label{fig: up-1}
\end{figure}

A typical result for the single neuron target function is shown in Figure \ref{fig: up-1}.
We see that overall, the qualitative behavior of GD is similar to that of the case when $n= \infty$ studied in the last section. There are still  two phases, and the convergent solution is sparse. Figure \ref{fig: up-1b} shows that  the quenching process becomes slower compared to the case with infinite data. 

\begin{figure}[!h]
    \centering
    \begin{subfigure}{0.4\textwidth}
    \includegraphics[width=\textwidth]{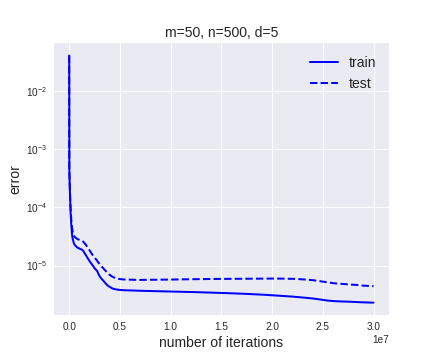}
    \caption{}\label{fig: up-2a}
    \end{subfigure}
    \begin{subfigure}{0.4\textwidth}
    \includegraphics[width=\textwidth]{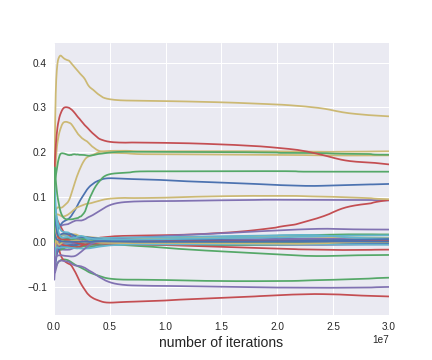}
    \caption{}\label{fig: up-2b}
    \end{subfigure}
    \caption{\small GD dynamics for the circle neuron target function in the under-parametrized case. Here $m=50, n=500, d=5$. (a) The dynamics of training and test loss. (b) The dynamics of the outer layer coefficients.}
    \label{fig: up-2}
\end{figure}

Figure \ref{fig: up-2} shows the result for the circle neuron target function. We see 
again that there are still two phases. But the multi-step phenomena becomes less pronounced compared to
case with  the infinite data. From Figure \ref{fig: up-2b}, the neurons can still roughly be divided into two groups:
active and background neurons. 
 Also we still observe some activation during the second phase, although this process is much more smooth compared to the case with infinite data.

\subsection{The mildly over-parametrized regime}

We call the regime between the highly over-parametrized and under-parametrized regimes the mildly over-parametrized regime.
The key question of interest is how the mildly over-parametrized regime bridges the highly over-parametrized regime and
the under-parametrized regime. 
We have already seen that these two regimes differ in several aspects.
One is that in the highly over-parametrized regime,  the inner layer coefficients $\bb$ barely change.
In the under-parametrized regime,  a small number of neurons experience large changes in their inner-layer coefficients.

\begin{figure}[!h]
    \centering
    \includegraphics[width=0.32\textwidth]{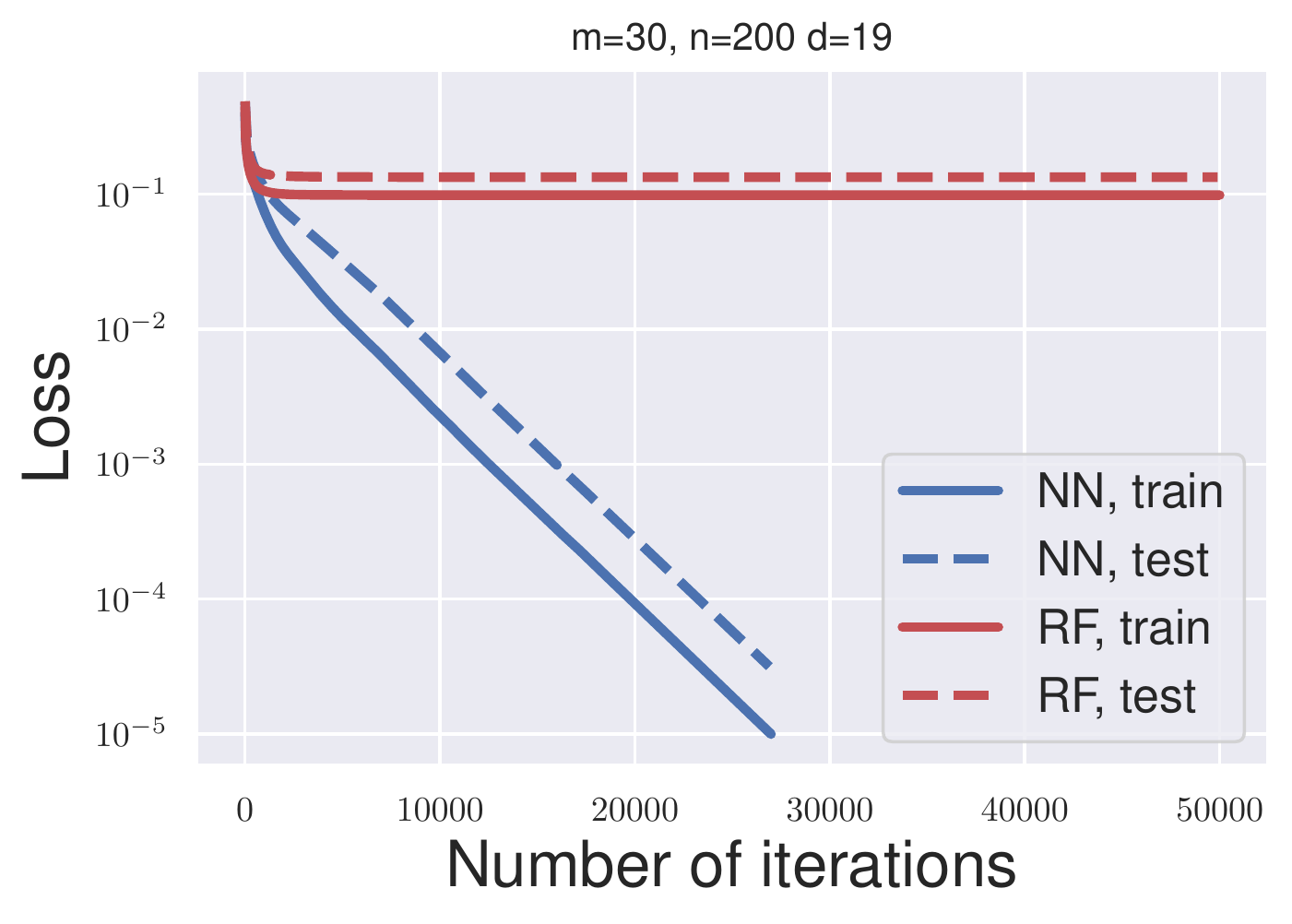}
    \includegraphics[width=0.32\textwidth]{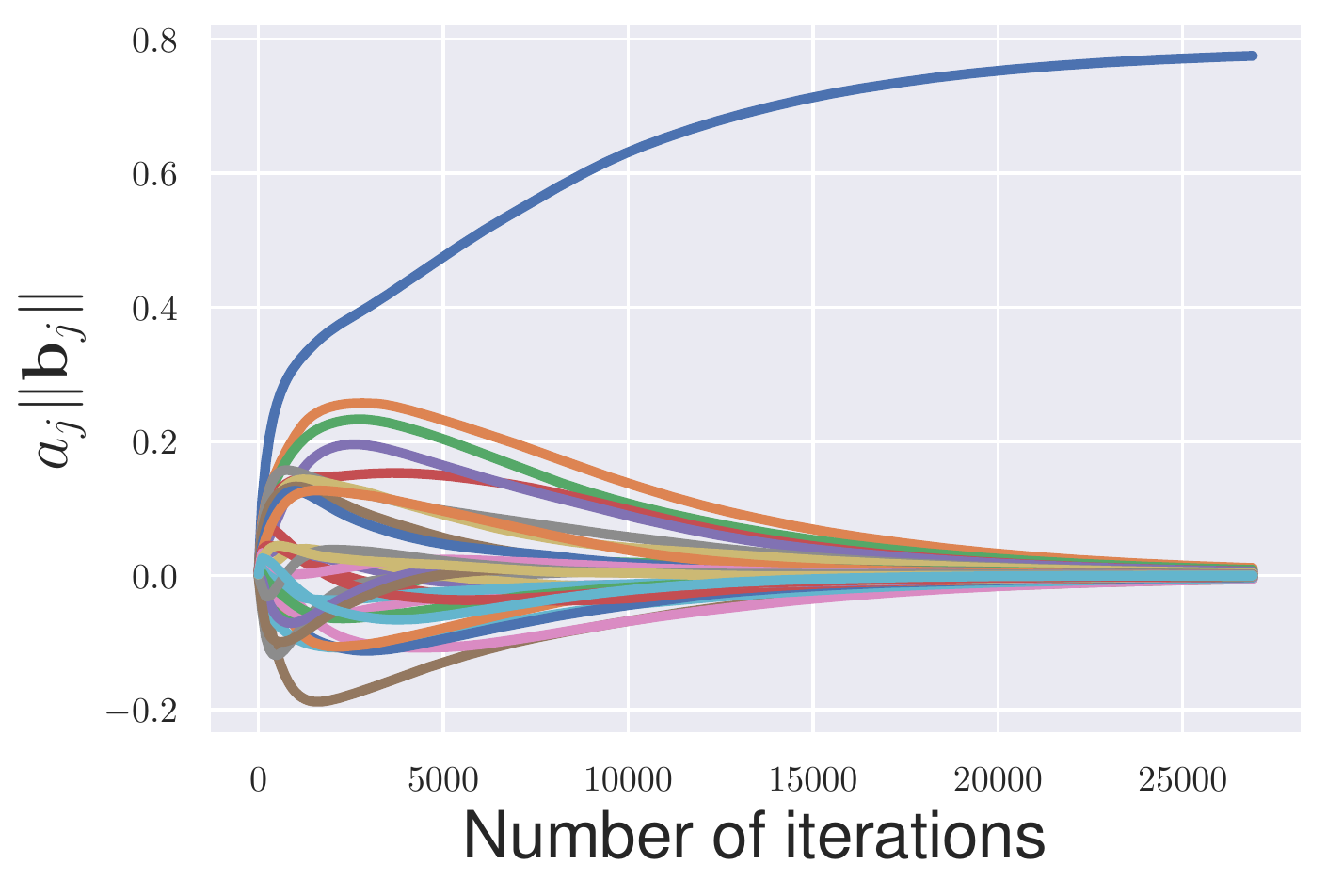}
    \includegraphics[width=0.32\textwidth]{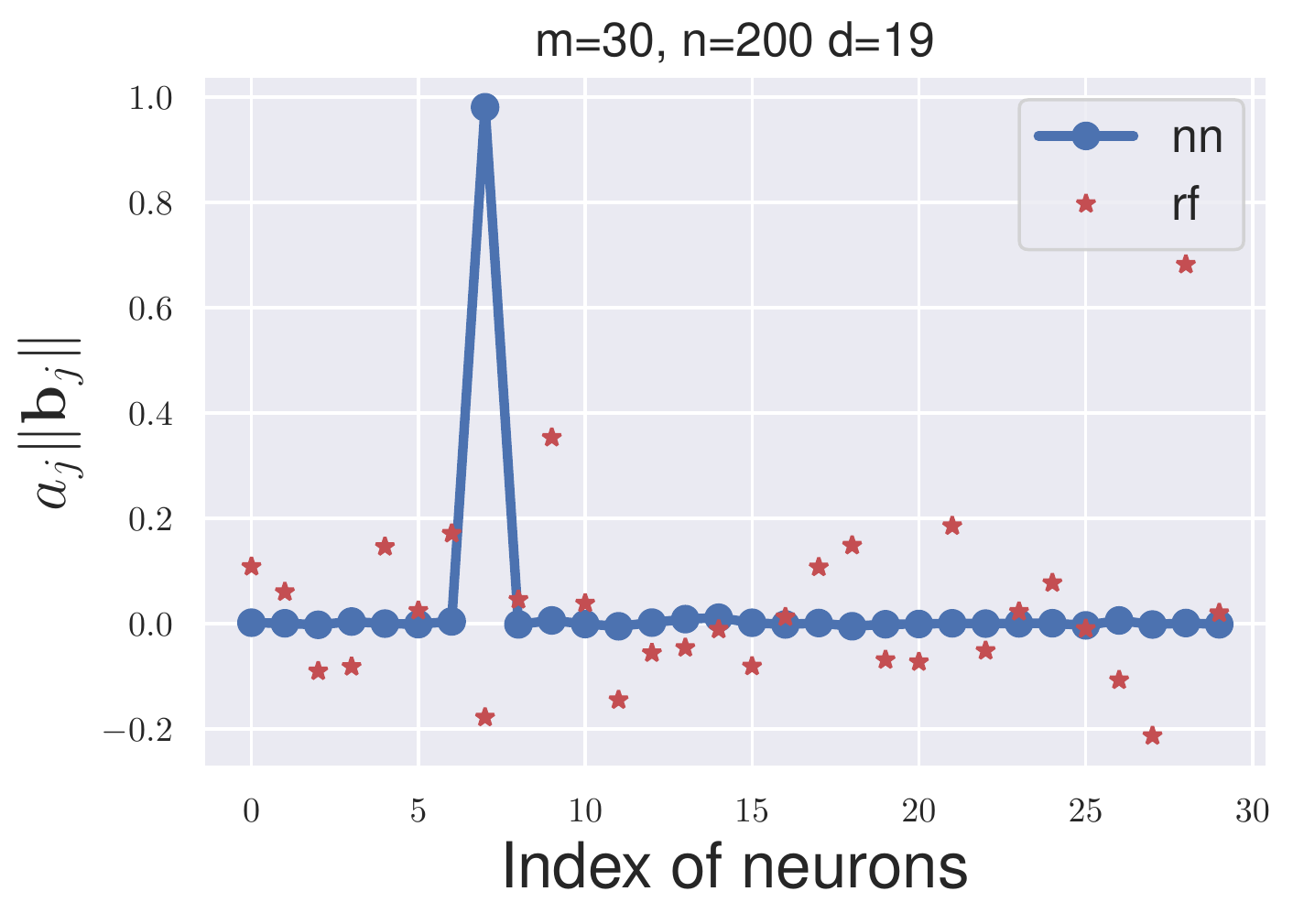}
    \caption{\small Results for the case when $m=3n/(d+1)$. Here $m=30, n=200, d=19$ and the learning rate $\eta=0.001$. (Left) The dynamics of the training and test losses (results from the corresponding random feature model is also plotted for comparison). (Middle) The dynamics of the ``magnitude'' of each neuron. (Right) The ``magnitude'' of each neuron of the convergent solution.}
    \label{fig:MOP-3}
\end{figure}

\begin{figure}[!h]
    \centering
    \includegraphics[width=0.32\textwidth]{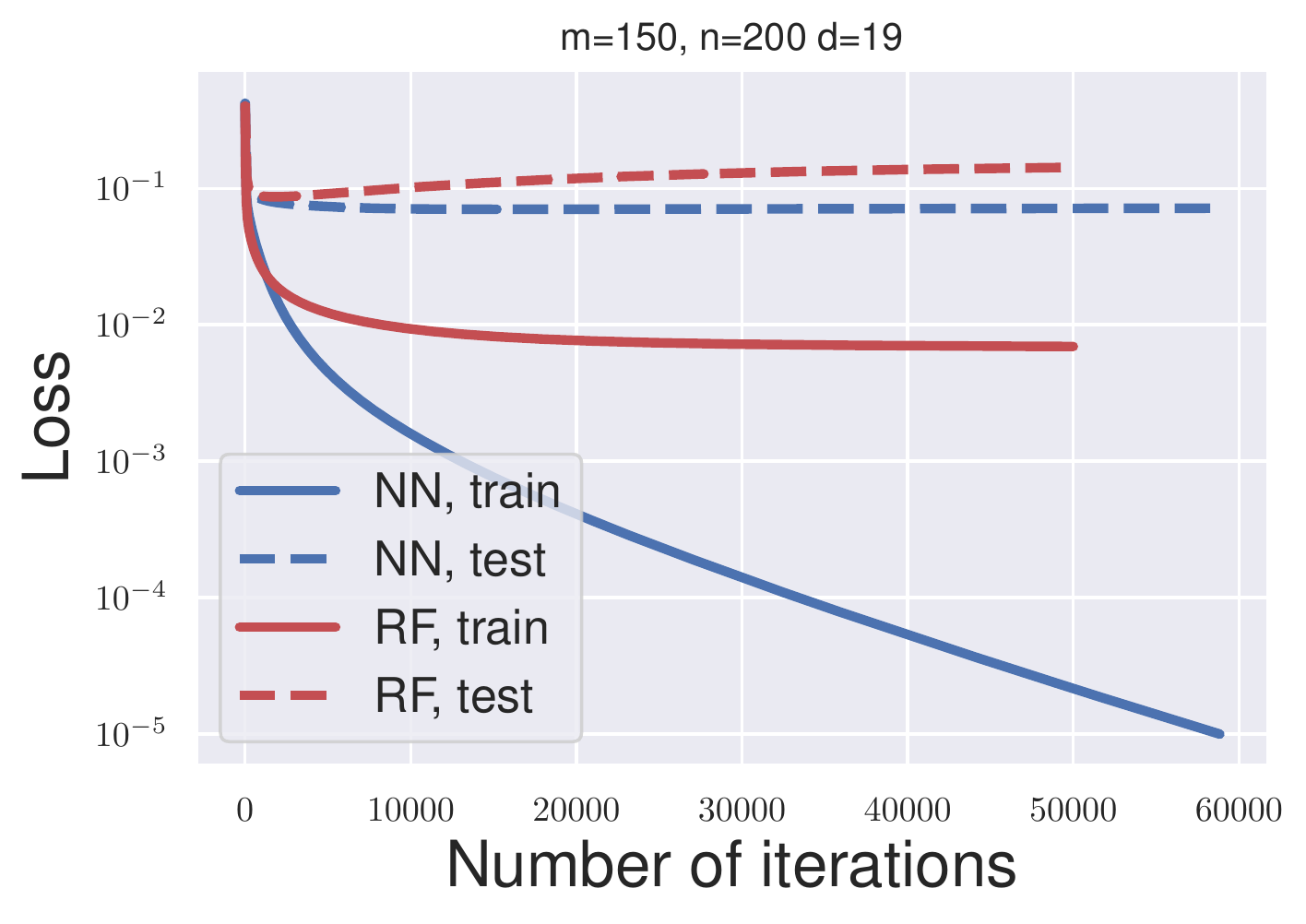}
    \includegraphics[width=0.32\textwidth]{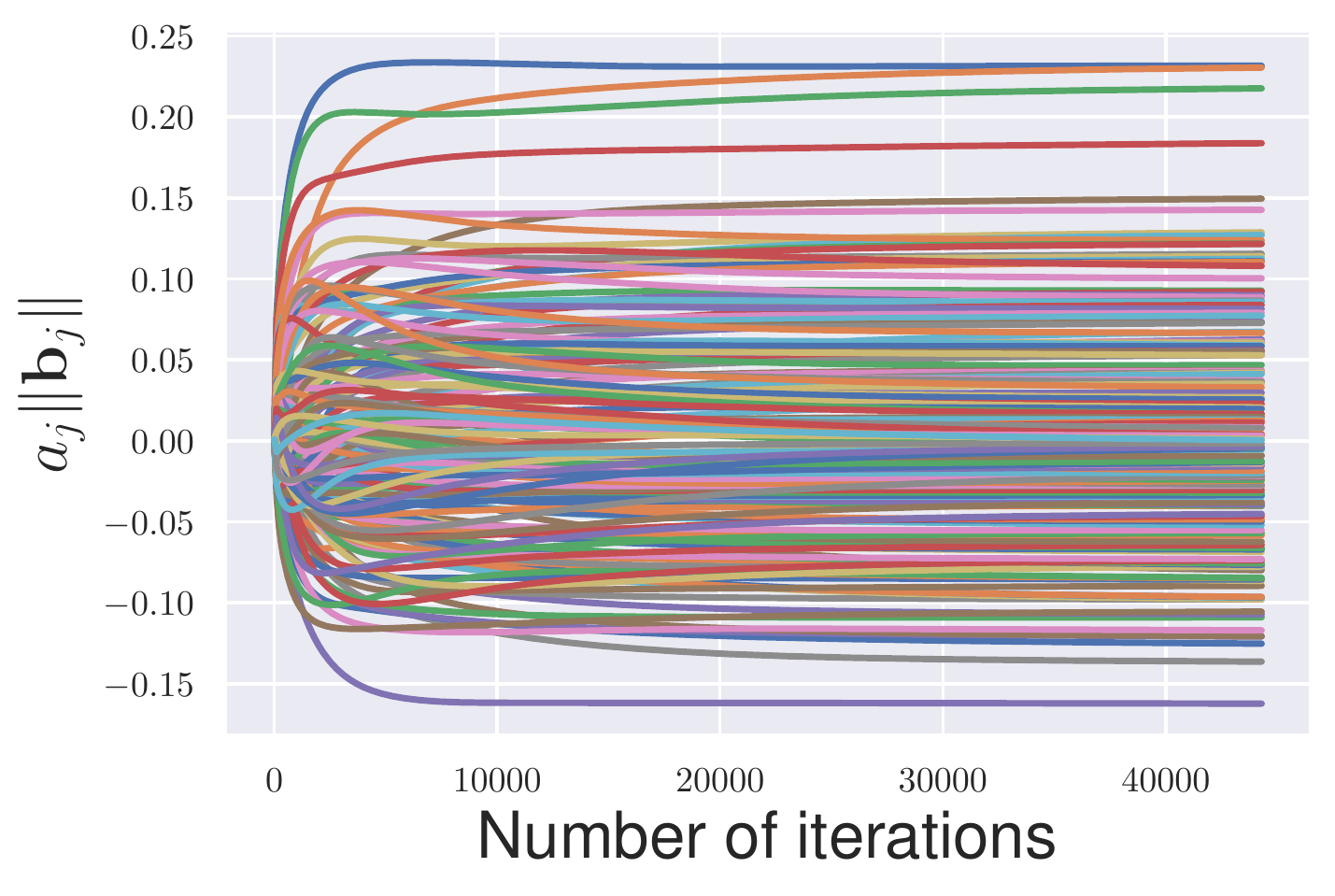}
    \includegraphics[width=0.32\textwidth]{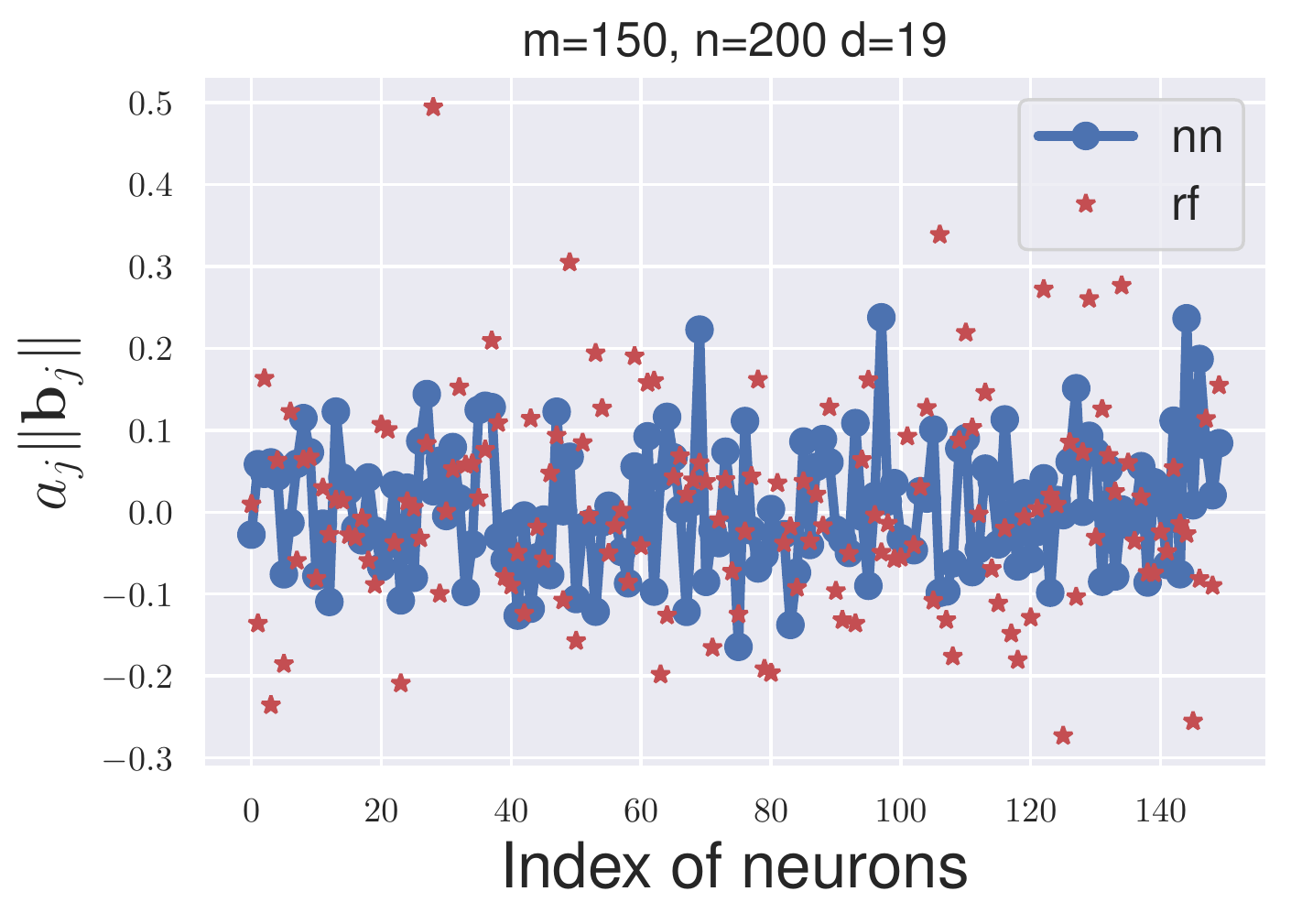}
    \caption{ \small Results  for the case when $m=0.75n$. Here $m=150, n=200, d=19$ and the learning rate $\eta=0.001$. (Left) The dynamics of the training and test losses (results from the corresponding random feature model is also plotted for comparison). (Middle) The dynamics of the ``magnitude'' of each neuron. (Right) The ``magnitude'' of each neuron of the convergent solution. }
    \label{fig:MOP-2}
\end{figure}

First,  we investigate the GD dynamics for the two interesting scalings: $m=c n/(d+1)$ and $m= C n$ where $c$ and $C$ are
constants.
Shown in Figures \ref{fig:MOP-3} and \ref{fig:MOP-2} are two examples $m=3 n/(d+1)$ and  $m=0.75n$ with $n=200, d=19$ respectively.  More examples with the same scalings but different values of $n$ can be found in Appendix \ref{sec: mild-appendix}.

One can see that the behavior shown in Figures \ref{fig:MOP-3}  resembles  the ones shown for the under-parametrized regime,
whereas the behavior shown in  Figures \ref{fig:MOP-2} resembles  the ones shown for the highly over-parametrized regime.
In the first case, the test accuracy  improves substantially after the GD dynamics departs from that of the RFM,
and there is a notable presence of the activation phenomena. In the second case,  the test error saturates soon after the GD dynamics departs from that of the RFM,
and there are no clear presence of the activation phenomena.
We will call the first case  ``NN-like''  and the second case ``RF-like''.

\begin{figure}[!h]
    \centering
    \includegraphics[width=0.45\textwidth]{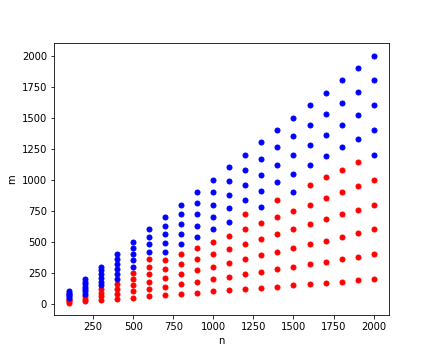}
    \caption{\small Transition from NN-like behavior to RF-like behavior in the mildly over-parameterized regime. The red and blue dots represent the NN-like and RF-like behavior, respectively.}
    \label{fig:MOP-5}
\end{figure}

To examine further the transition between the RF-like and NN-like behavior, we plot in Figure \ref{fig:MOP-5}
where the transition takes place in the $m-n$ plane.  Note that the assignment to the NN-like or RF-like behavior is based on 
the following subjective criterion:
\begin{itemize}
\item  NN-like: test error keeps decreasing in the relatively early time after
the initial fast decreasing phase
\item Kernel-like: test error stays flat or even increases after the initial fast decreasing phase.
\end{itemize}
The distinction becomes fuzzy in the transition region.
In fact, at this point we can not rule out the existence of a third kind of behavior.  We will leave this question to future investigation.

\section{ Generalization error and the path norm}

Next we examine the generalization error of the solutions selected by the GD dynamics.
Since all these target functions we studied are Barron function, the approximation error satisfies a Monte-Carlo like
estimate \cite{barron1993universal, breiman1993hinging, bach2017breaking, e2018priori}. The main interest lies in the estimation error or the
generalization gap which is controlled by the Barron norm of the solution selected by the GD dynamics (see \cite{bach2017breaking, e2018priori}).
Obviously an upper bound for the Barron norm is the path norm of the parameters,  defined by
\begin{equation}
 \|\theta\|_P =   \sum_{j=1}^m  |a_j |  \|\bb_j \|_1 
\end{equation}
Therefore we will study the path norm of the parameters selected by GD.

\begin{figure}[!ht]
\centering
    \begin{subfigure}{0.5\textwidth}
    \includegraphics[width=0.51\textwidth]{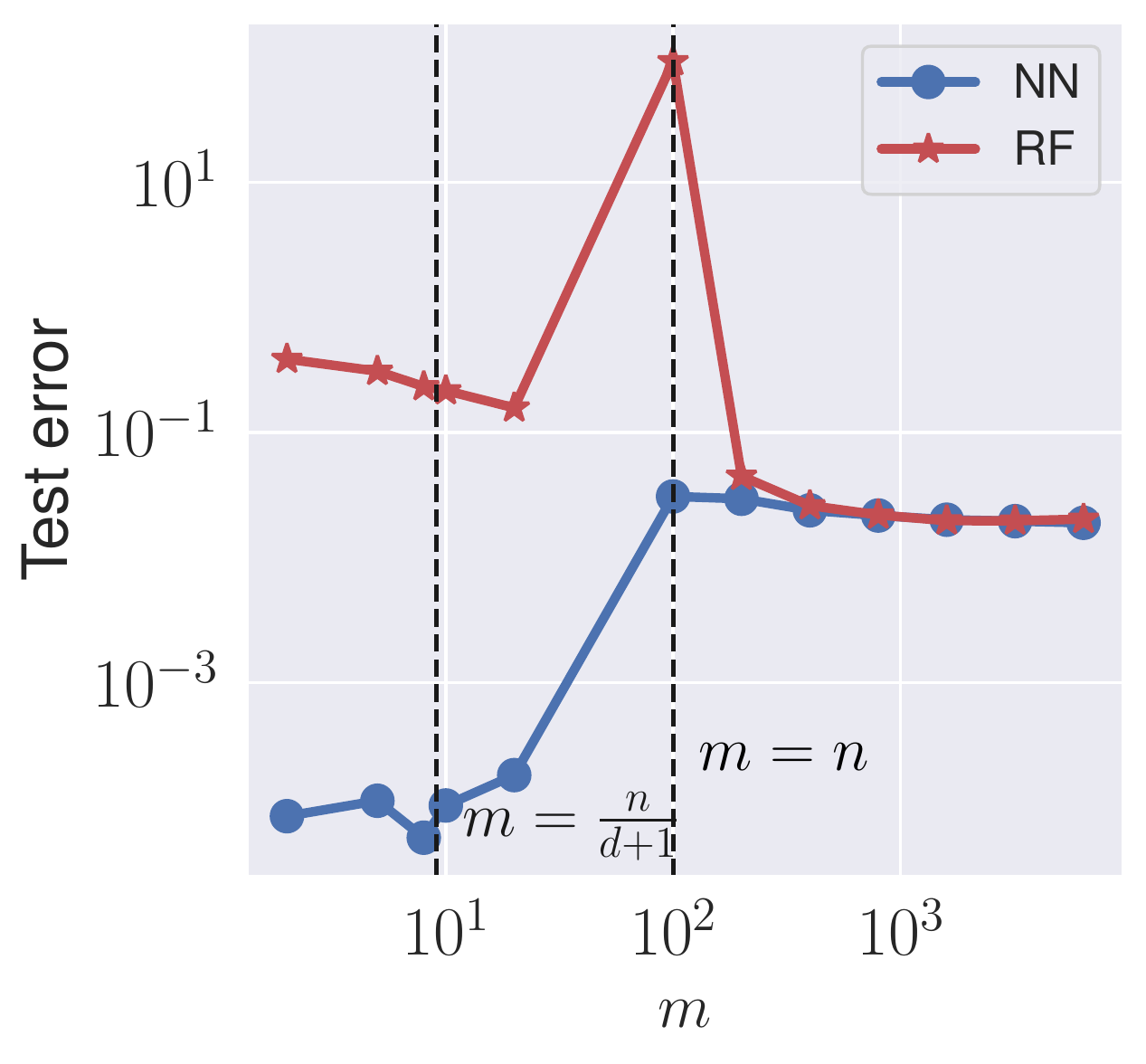}
    \includegraphics[width=0.47\textwidth]{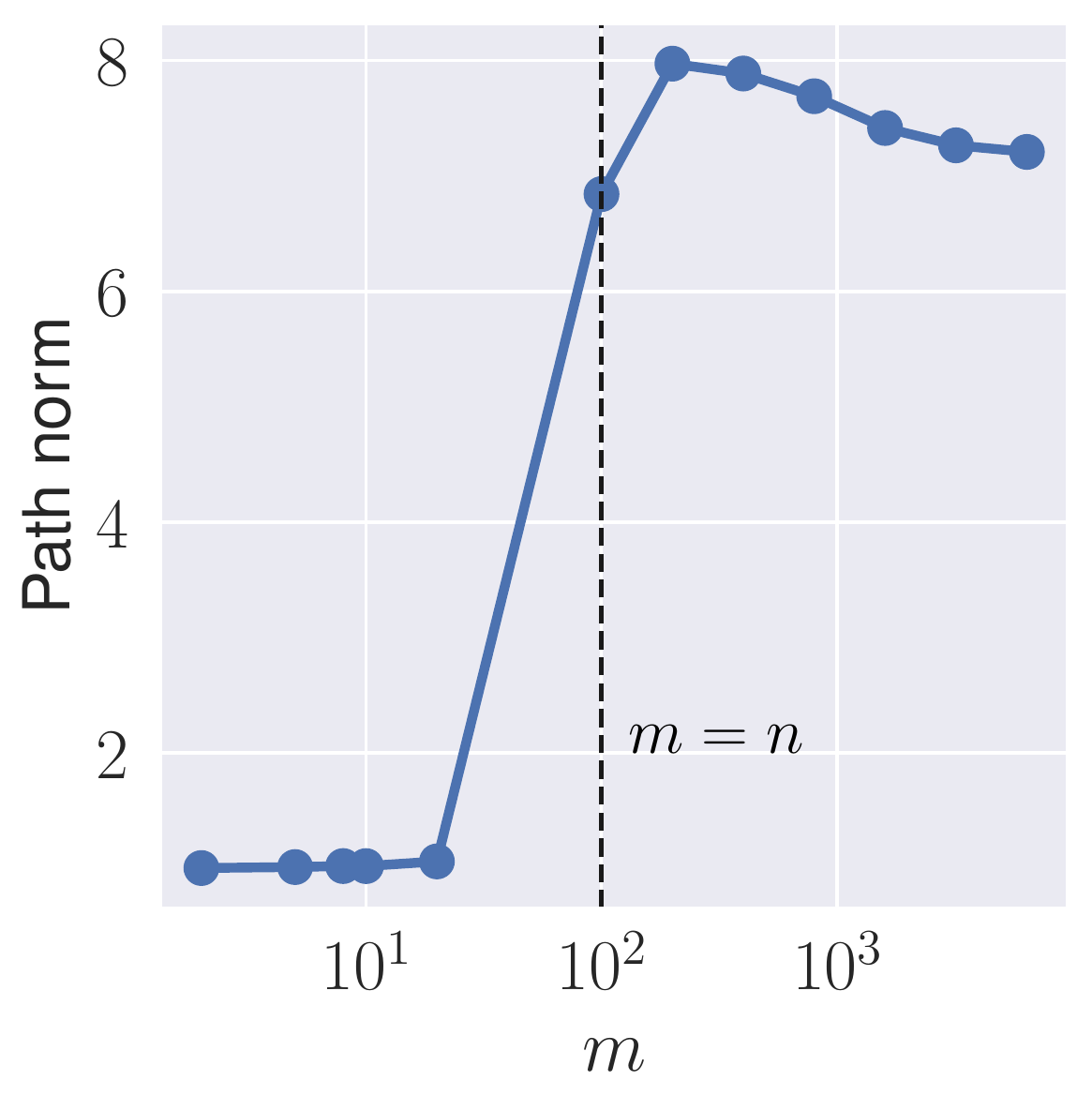}
    \caption{Single neuron $f^*_1$.}\label{}
    \end{subfigure}
    \hspace*{-5mm}
    \begin{subfigure}{0.5\textwidth}
    \includegraphics[width=0.5\textwidth]{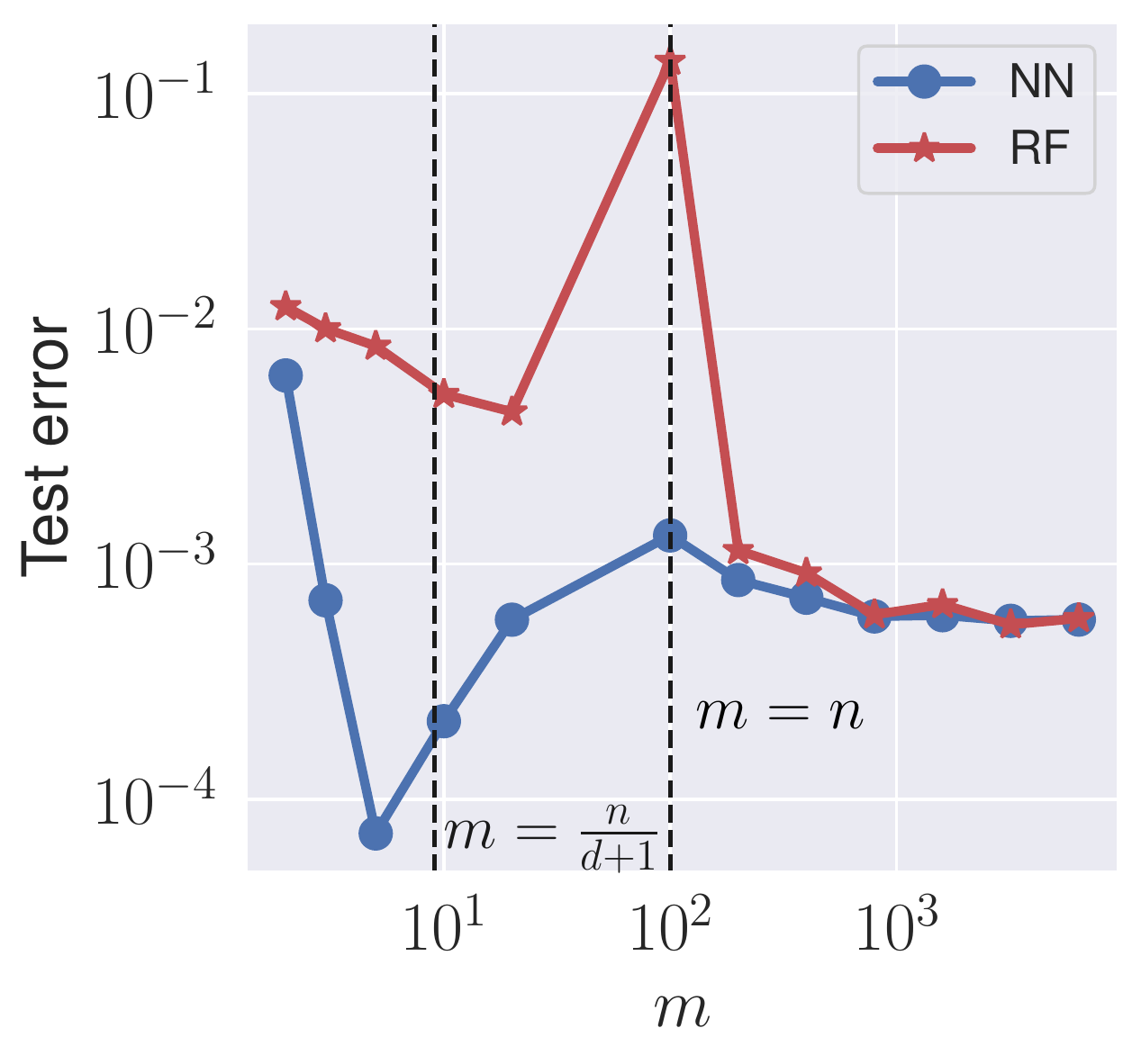}
    \includegraphics[width=0.47\textwidth]{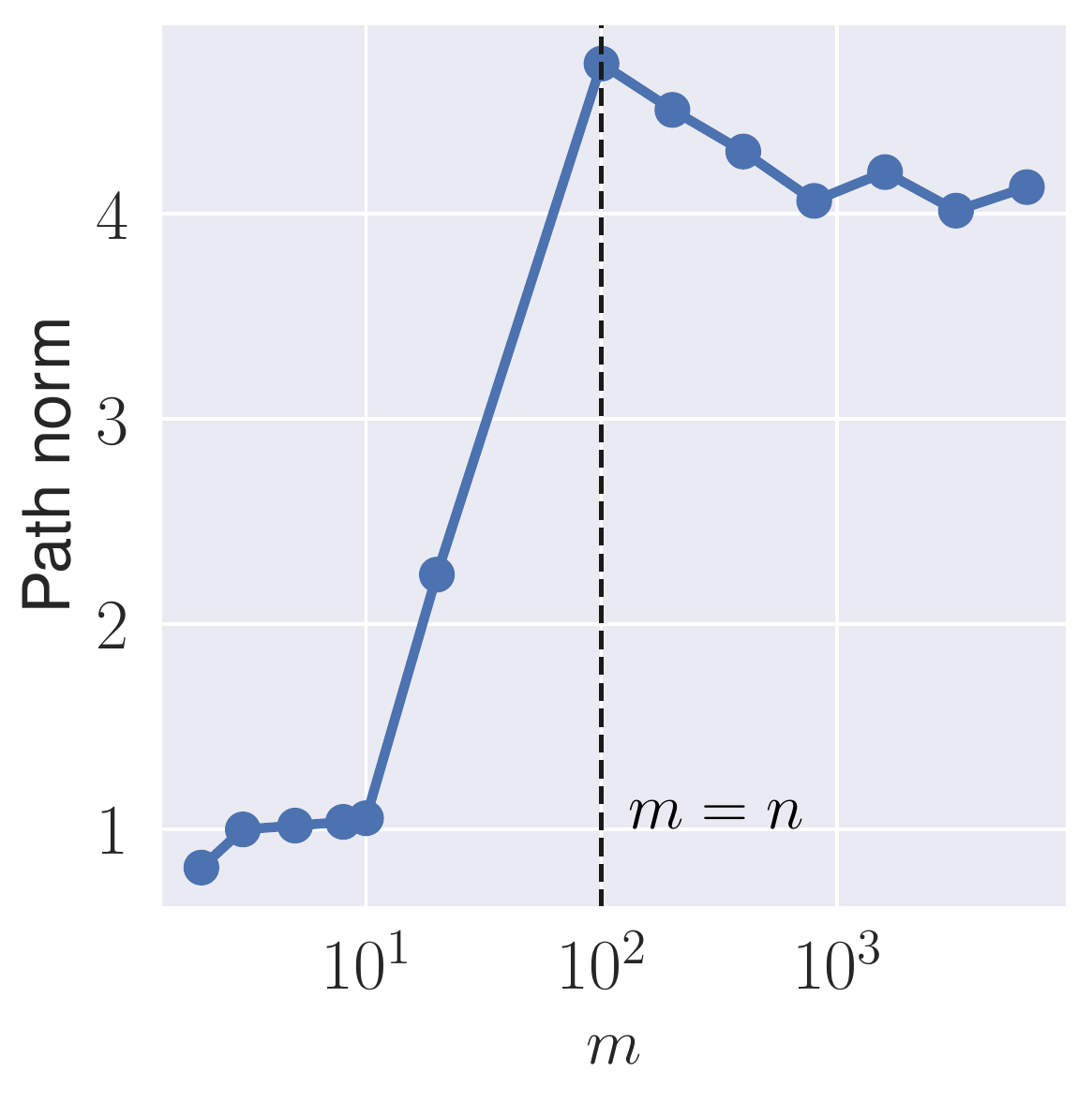}
    \caption{Circle neuron $f^*_2$.}\label{}
    \end{subfigure}
    \caption{ \small Generation error and path norm of  the GD solutions with varying number of neurons. Here $n=100, d=10$, and the learning rate  is $0.001$. GD is stopped after the training error is smaller than $10^{-5}$. For each target function: (Left) test error as a function of the number of neurons; (Right) path norm as a function of the number of neurons.
    }
    \label{fig: mop-1}
\end{figure}

Figure \ref{fig: mop-1} examines 
 the test error and the path norm as $m$ changes for two target functions. 
One can see that as $m$ becomes larger, the test error of the NN model eventually becomes close to that of  the RFM.
One thing to notice is that these changes seem to behave smoothly across the points where $m=n/(d+1)$ and $m=n$, where  the NN and the RFM change from an under-parameterized situation to an over-parameterized
situation, respectively.  We do not observe the peak of test errors around $m=n/(d+1)$ as suggested in \cite{belkin2019reconciling}. We suspect that the peak observed in   \cite{belkin2019reconciling} 
is the result of the special training method used there.  Interestingly, Figure \ref{fig: mop-1} shows that there does exist a peak around $m=n$, the same place  for the RFM. 
This should be the result of the close proximity of the GD dynamics for  2LNN and RFM during the first phase, 
the latter performs extremely badly when $m=n$ due to resonance as shown in \cite{ma2019gd}.
Thus it is not surprising to see that NN also performs the worst around $m=n$.
We also observe that the test error sees a dramatic increase from $m=n/(d+1)$ to $m=n$.
 
Overall, the path norm seems to serve as a good indicator of the generalization performance. 
For example, one also sees a dramatic increase of the path norm from $m\sim n/(d+1)$ to $m\sim n$,
and the path norm peaks around $m=n$.

\begin{figure}[!h]
    \centering
    \begin{subfigure}{0.5\textwidth}
    \includegraphics[width=0.49\textwidth]{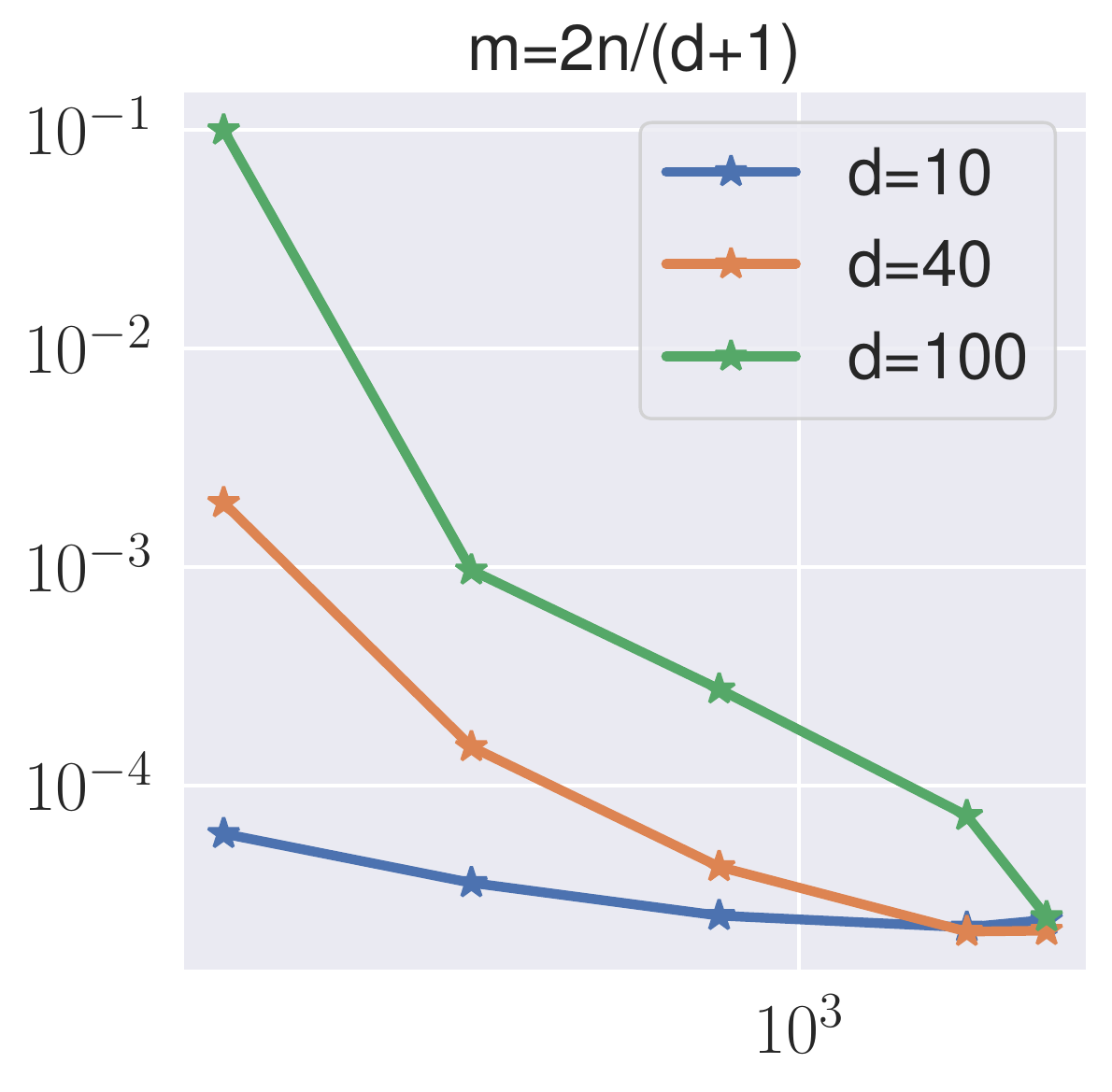}
    \includegraphics[width=0.49\textwidth]{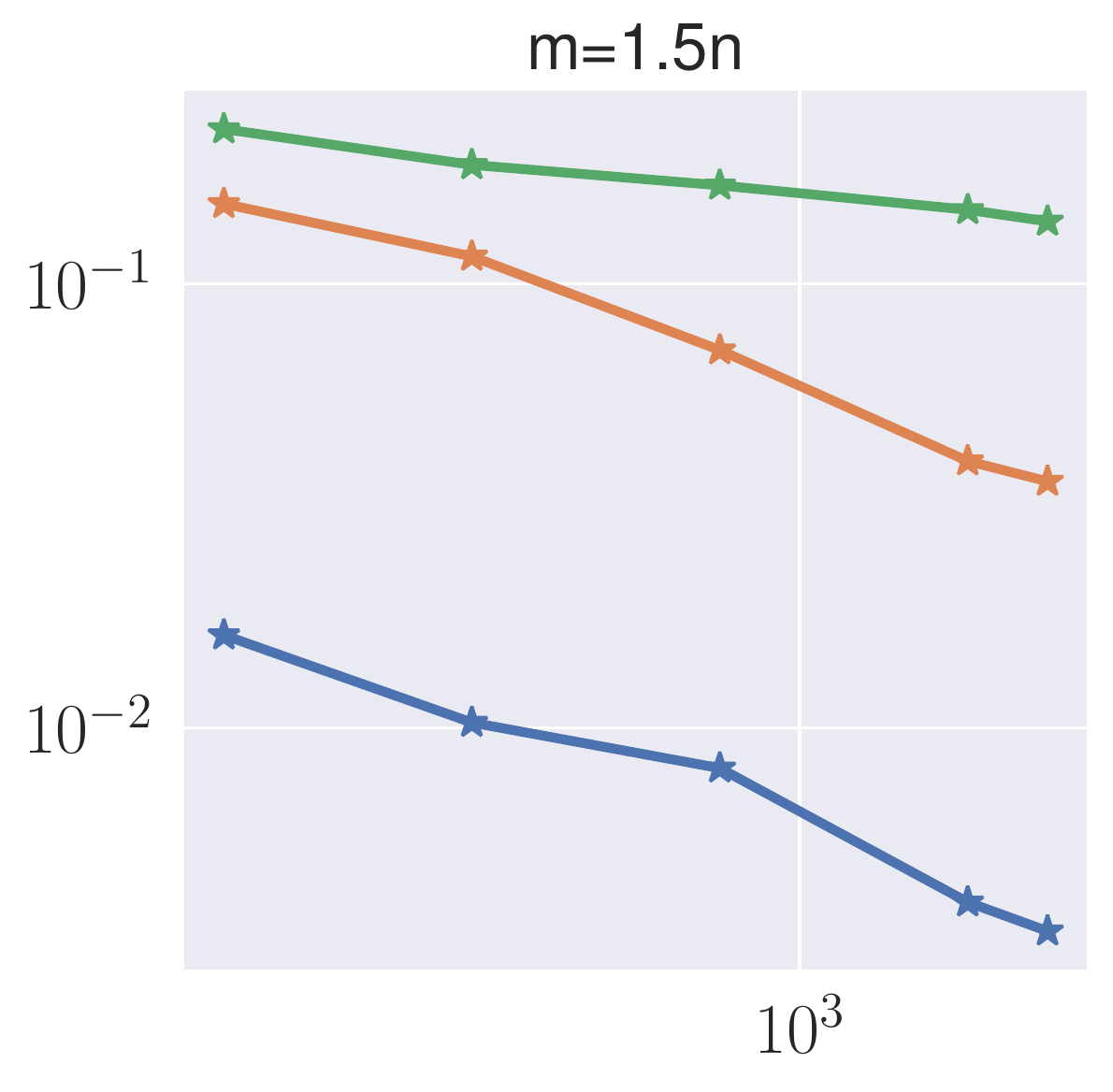}
    \caption{Single neuron $f^*_1$}\label{}
    \end{subfigure}
    \hspace*{-5mm}
    \begin{subfigure}{0.5\textwidth}
    \includegraphics[width=0.49\textwidth]{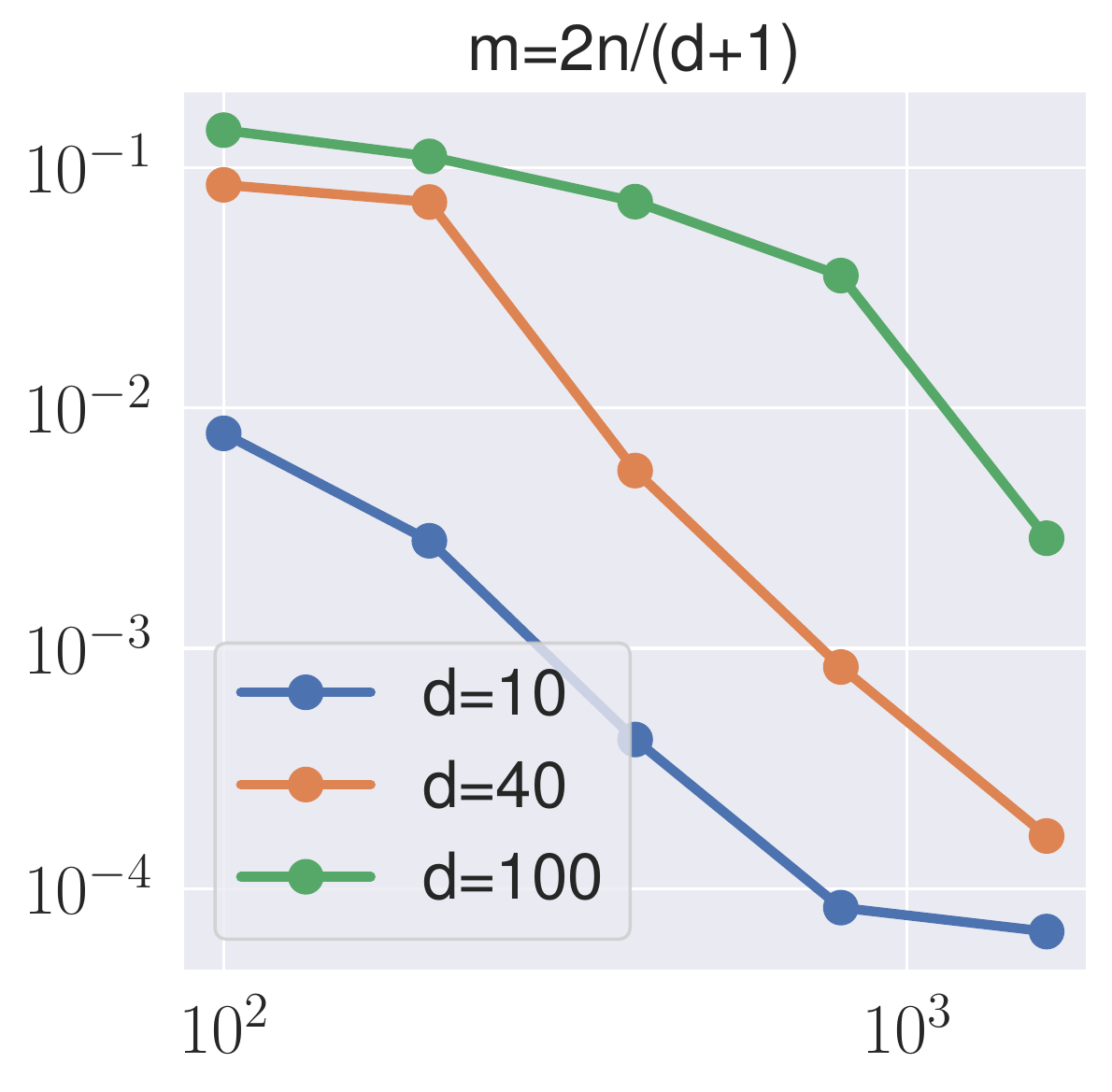}
    \includegraphics[width=0.49\textwidth]{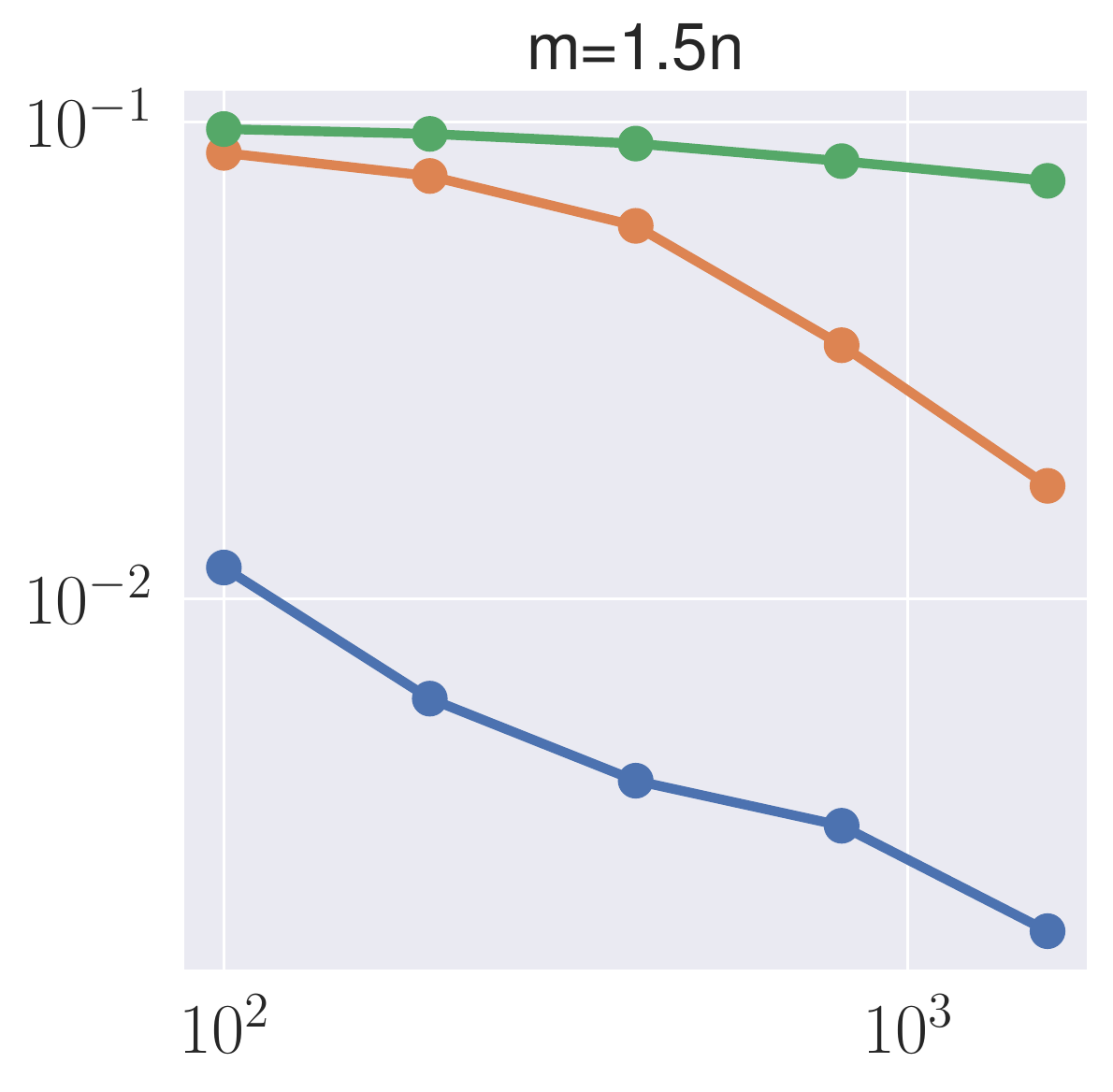}
    \caption{Circle neuron $f^*_2$}\label{}
    \end{subfigure}
    \caption{ \small Test error of the GD solutions with two different scalings. The horizontal axis denotes the number of samples; the vertical axis denotes the test error.  The learning rate is $0.001$ and    GD is stopped when the training error
    is smaller than $10^{-5}$. }
    \label{fig: mop-gen-err}
\end{figure}

\begin{figure}
\centering
\begin{subfigure}{0.4\textwidth}
\includegraphics[width=\textwidth]{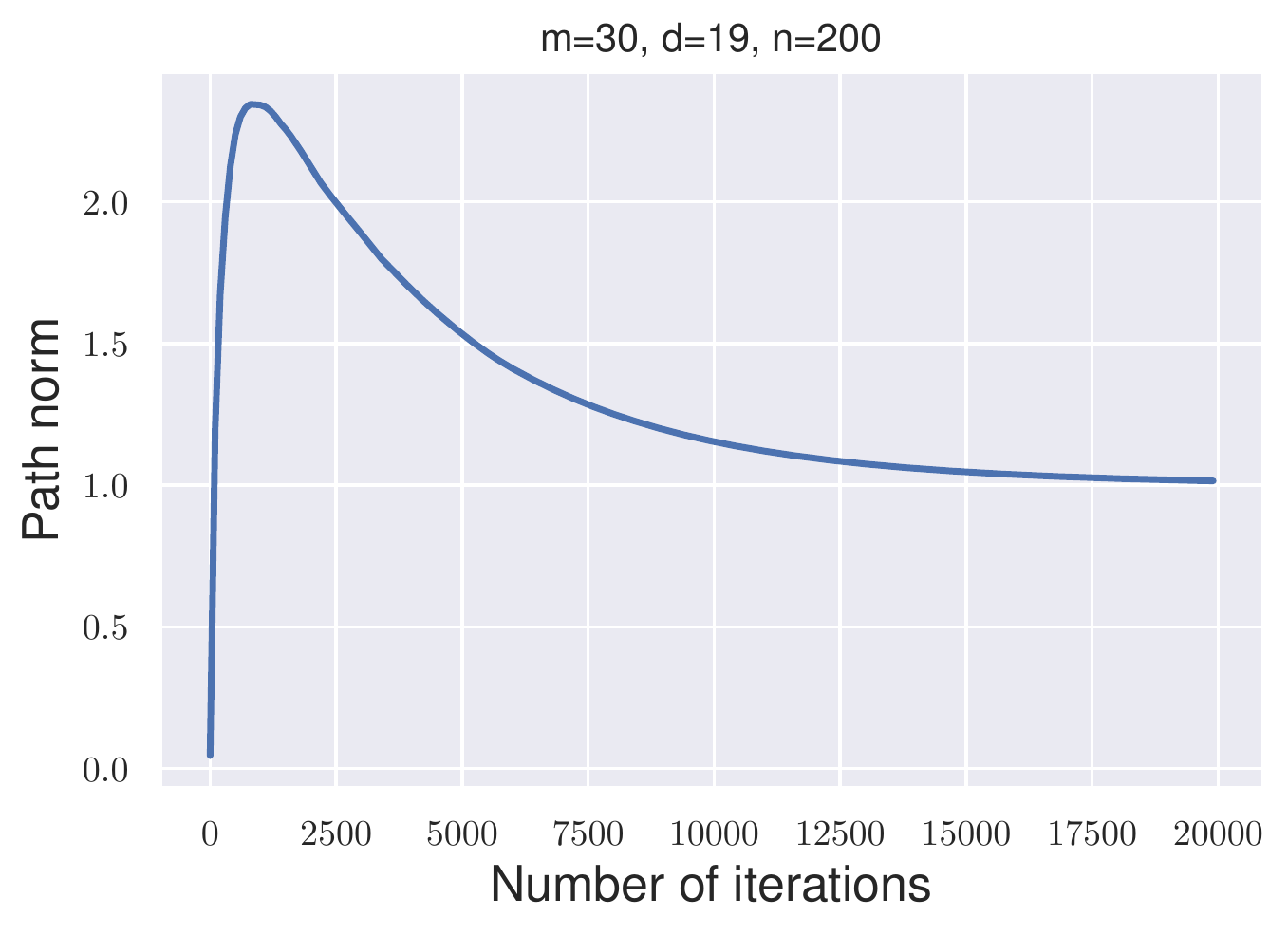}
\caption{$m\sim n/(d+1)$}\label{}
\end{subfigure}
\begin{subfigure}{0.4\textwidth}
\includegraphics[width=\textwidth]{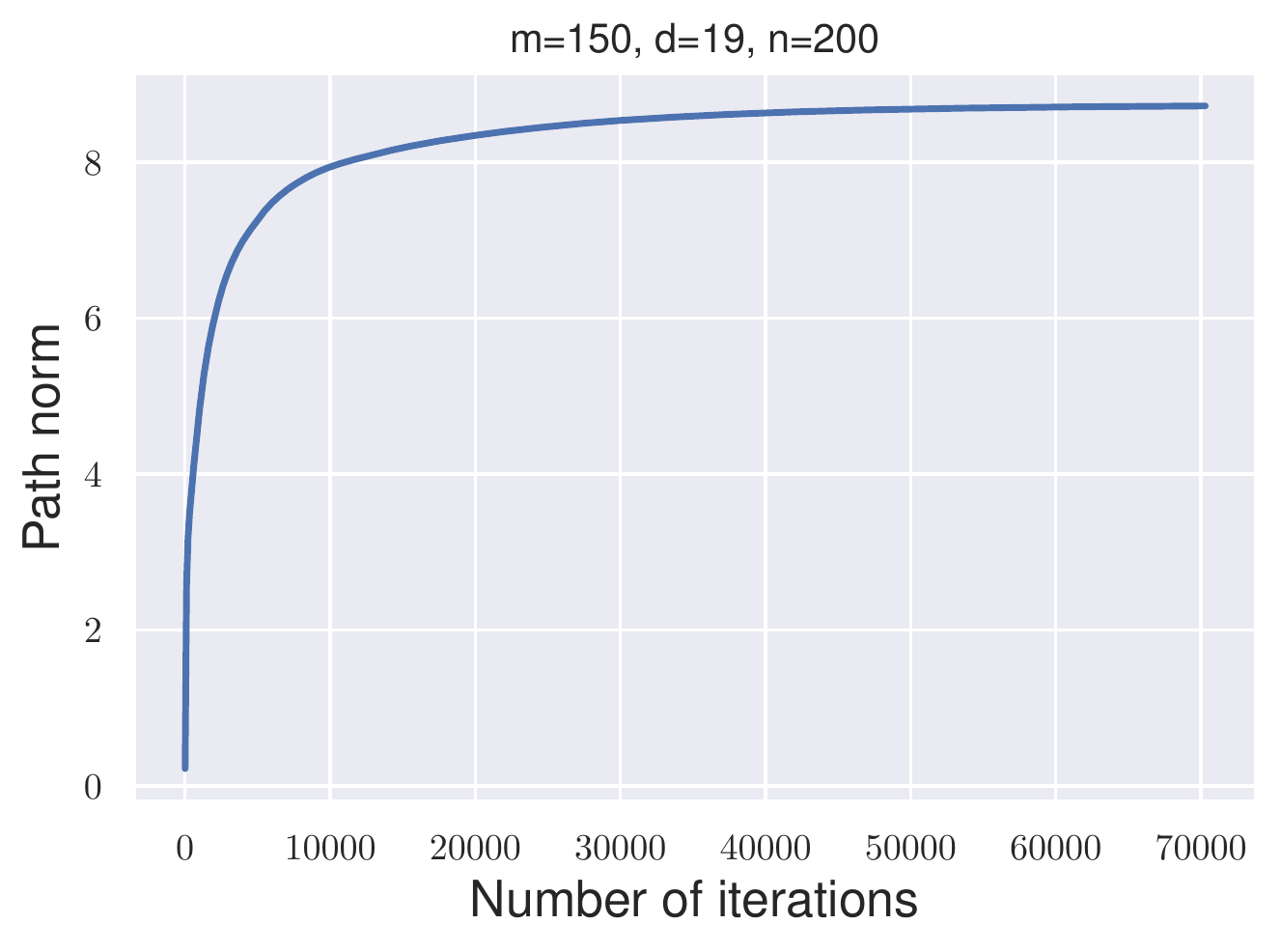}
\caption{$m\sim n$}\label{}
\end{subfigure}
\caption{\small The dynamics of the path norm. Here target function is the single neuron.}
\label{fig: dy-pnorm}
\end{figure}

\paragraph*{The difference between the two regimes}
To further explore the difference between the two regime $m\sim n/(d+1)$ and $m\sim n$, 
we show in Figure \ref{fig: mop-gen-err}  the test error   as   a function
  of  the training sample size for the two regimes. 
  The result suggests that under the scaling $m\sim n$, the test error may suffer from CoD, while for the scaling $m\sim  n/(d+1)$, 
   it  does not seem to be the case.  In Figure \ref{fig: dy-pnorm}, we show the dynamics of the path norm for the two regimes and we see
   that they also behave differently in these two regimes.

\section{GD dynamics with mean-field scaling}
The two-layer neural network under mean-field scaling is given by 
\begin{equation}
    f_m(\bx;\ba,\bB)=\frac{1}{m} \sum_{j=1}^m a_j \sigma(\bb_j^T\bx),
\end{equation}
and the corresponding GD dynamics is given by 
\begin{equation}\label{eqn: gd-mf}
\begin{aligned}
    \dot{a}_j(t) &= -\frac{1}{mn}\sum_{i=1}^n (f_m(\bx_i;\ba(t),\bB(t))-f^*(\bx_i))\sigma(\bb_j(t)^T \bx_i)\\
    \dot{\bb}_j(t) &= - \frac{1}{mn}\sum_{i=1}^n (f_m(\bx_i;\ba(t),\bB(t))-f^*(\bx_i))a_j(t)\sigma'(\bb_j(t)^T\bx_i)\bx_i.
\end{aligned}
\end{equation}
For simplicity, we call the above dynamics GD-MF and \eqref{eqn: GD-xv} GD-conventional. 
 Under the mean-field scaling, the speed for the $a_j$'s and $\bb_j$'s  are of the same order, i.e. there is no time-scale separation.

\begin{figure}
    \centering
    \begin{subfigure}{0.33\textwidth}
    \includegraphics[width=\textwidth]{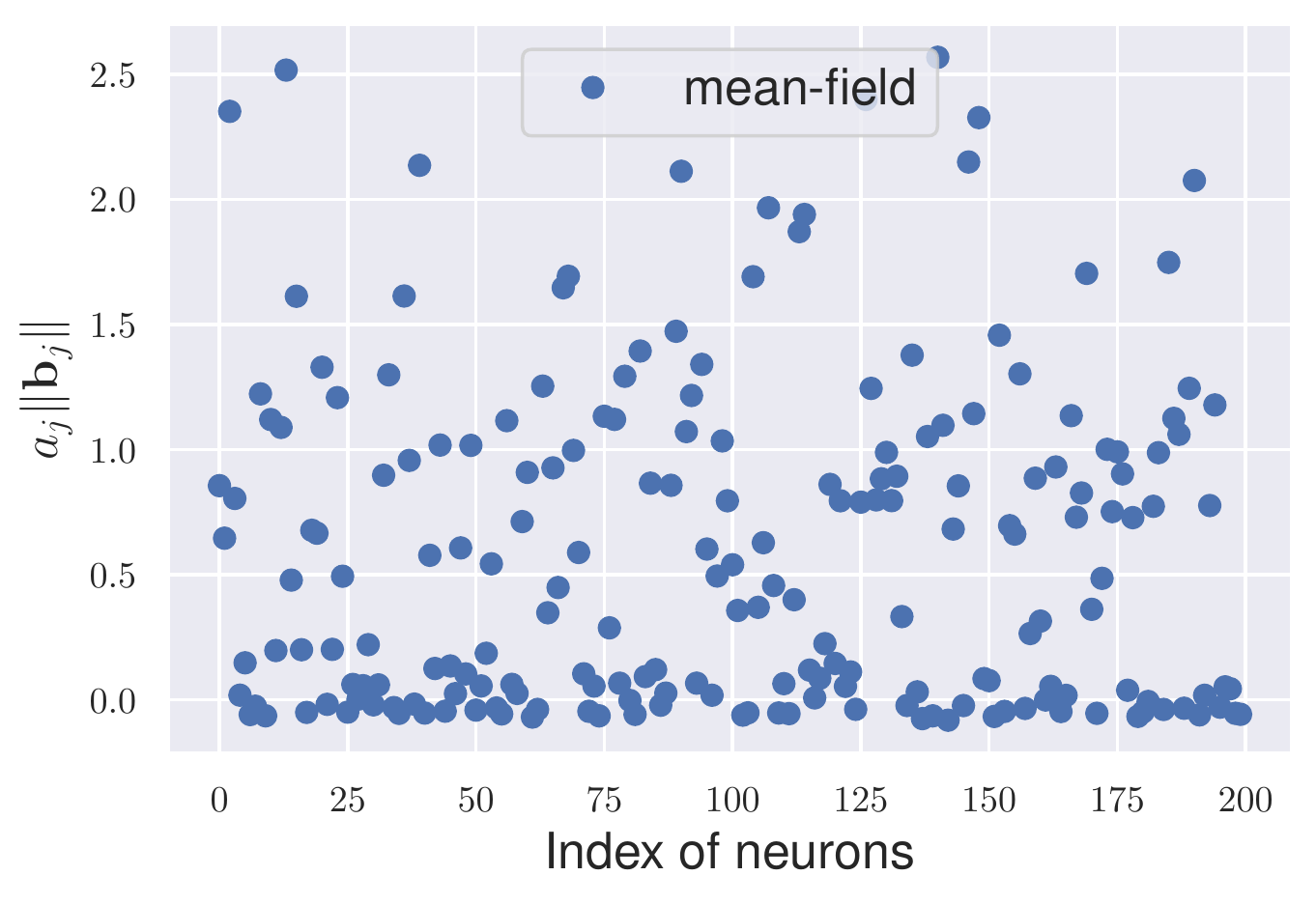}
    \caption{}
    \label{fig: mf-1a}
    \end{subfigure}
    \hspace{-2mm}
    \begin{subfigure}{0.33\textwidth}
    \includegraphics[width=\textwidth]{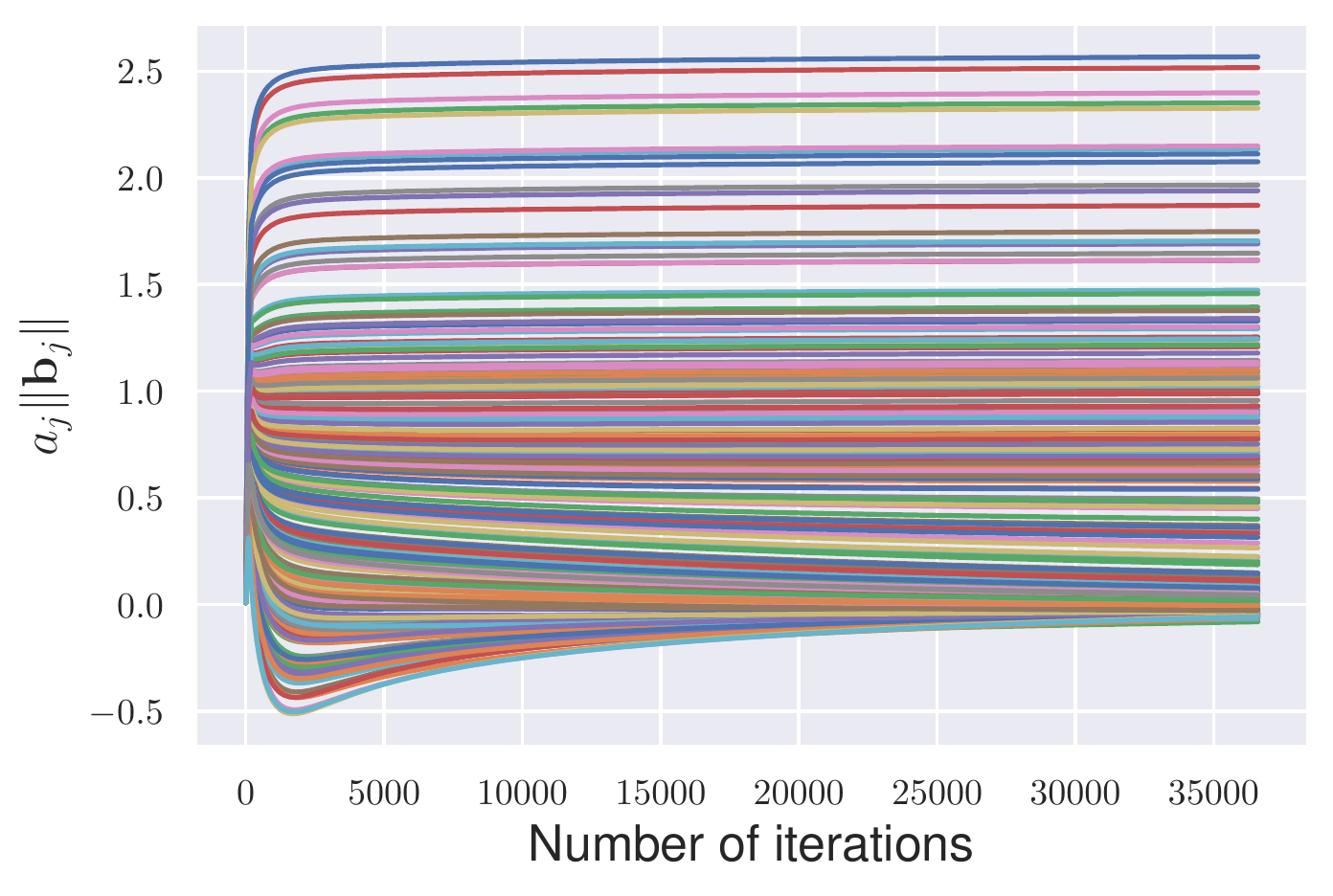}
    \caption{}
    \label{fig: mf-1b}
    \end{subfigure}
    \hspace{-2mm}
    \begin{subfigure}{0.33\textwidth}
    \includegraphics[width=\textwidth]{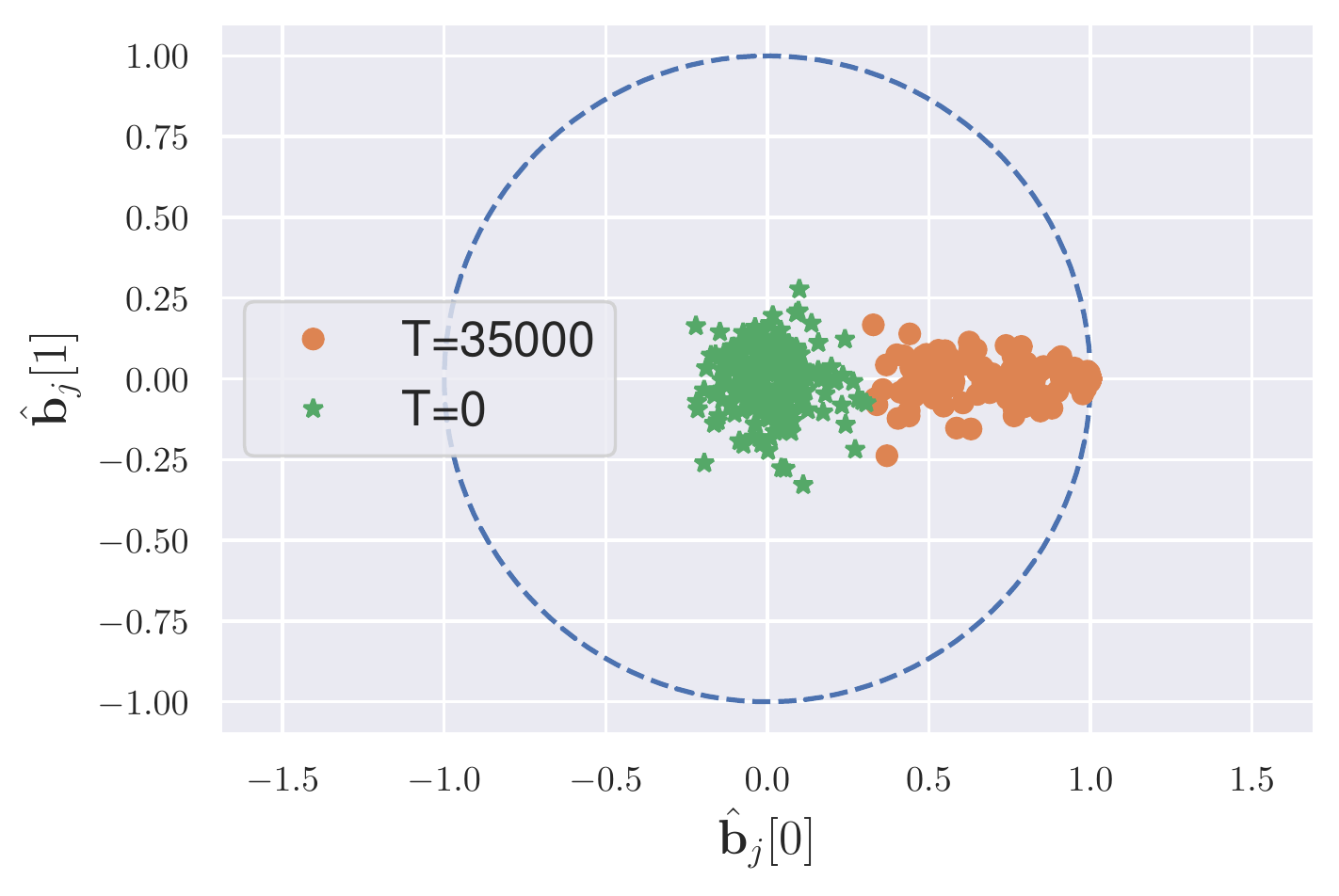}
    \caption{}
    \label{fig: mf-1c}
    \end{subfigure}
    \caption{\small  GD-MF dynamics for the single neuron target function $f_1^*$. Here $m=200, d=100, n=\infty$ and learning rate $\eta=0.001$.
    (a) The outer layer coefficient of each neuron for the converged solution. (b) The time history of
    the outer layer coefficient for each neuron. (c) The projection to the first two coordinates of  $\hat{\bb}$
    for each neuron. The green ones correspond to the random initialization; the orange ones correspond
     to the solutions found by GD-MF. }
    \label{fig: mf-1}
\end{figure}

First let us look at the case when $n=\infty$. 
Figure \ref{fig: mf-1} shows the dynamic behavior of GD-MF for single neuron target function $f_1^*$. Different from GD-conventional, we see that  almost all the neurons move significantly and contribute  to the model.  

\begin{figure}
    \centering
    \begin{subfigure}{0.5\textwidth}
    \includegraphics[width=0.49\textwidth]{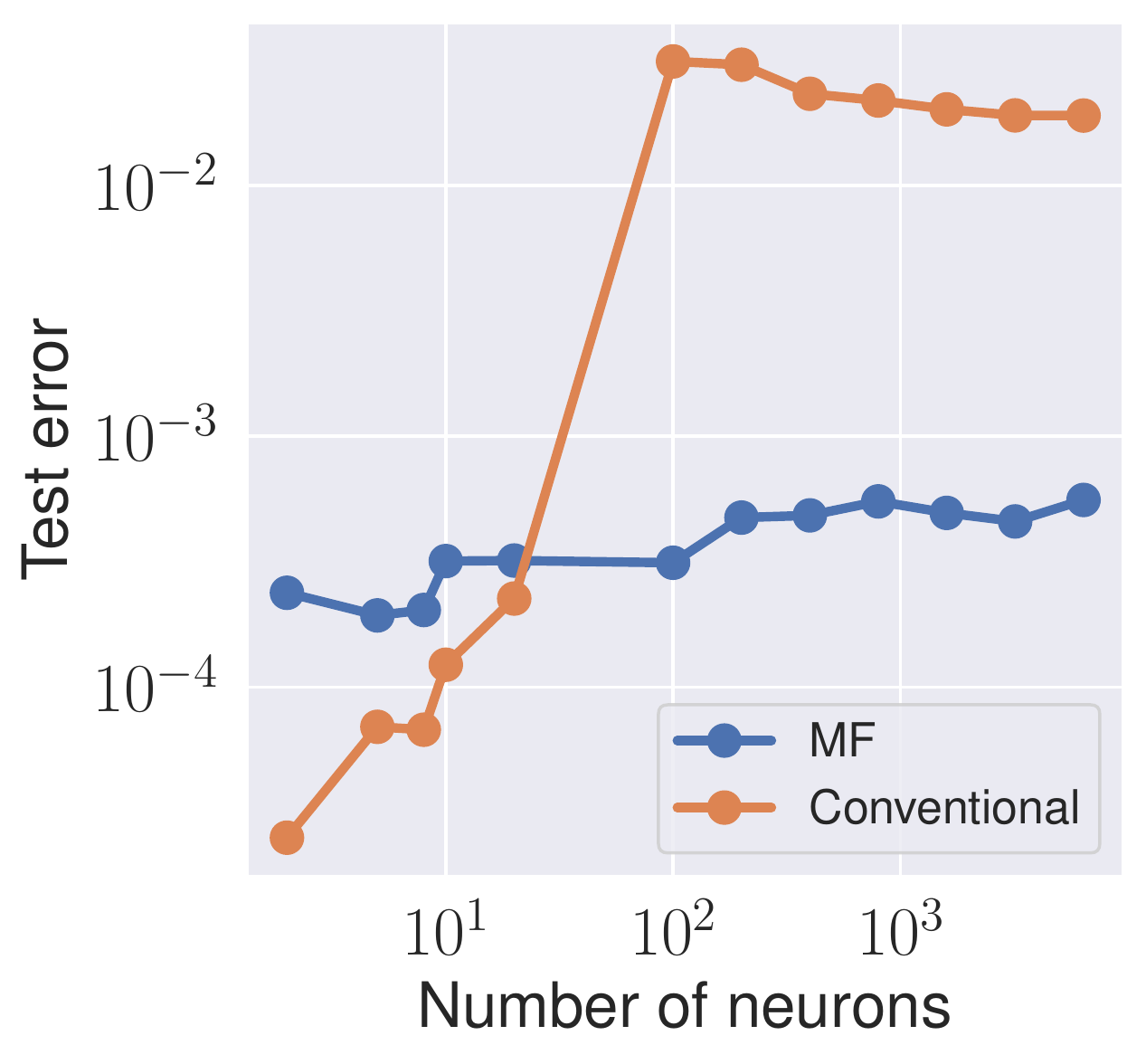}
    \includegraphics[width=0.49\textwidth]{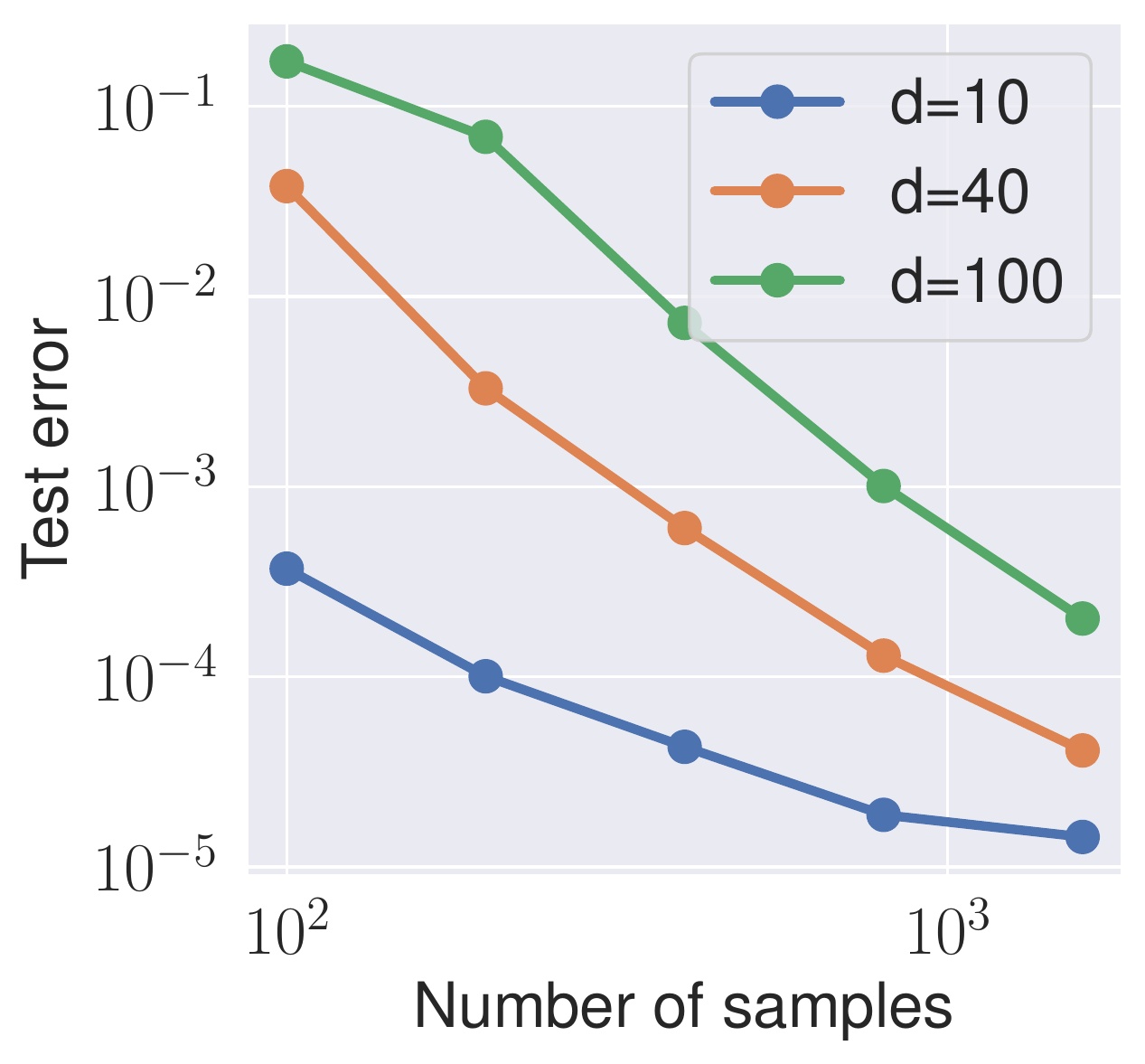}
    \caption{Single neuron $f^*_1$.}
    \label{fig: mf-2a}
    \end{subfigure}
    \hspace*{-3mm}
    \begin{subfigure}{0.5\textwidth}
    \includegraphics[width=0.49\textwidth]{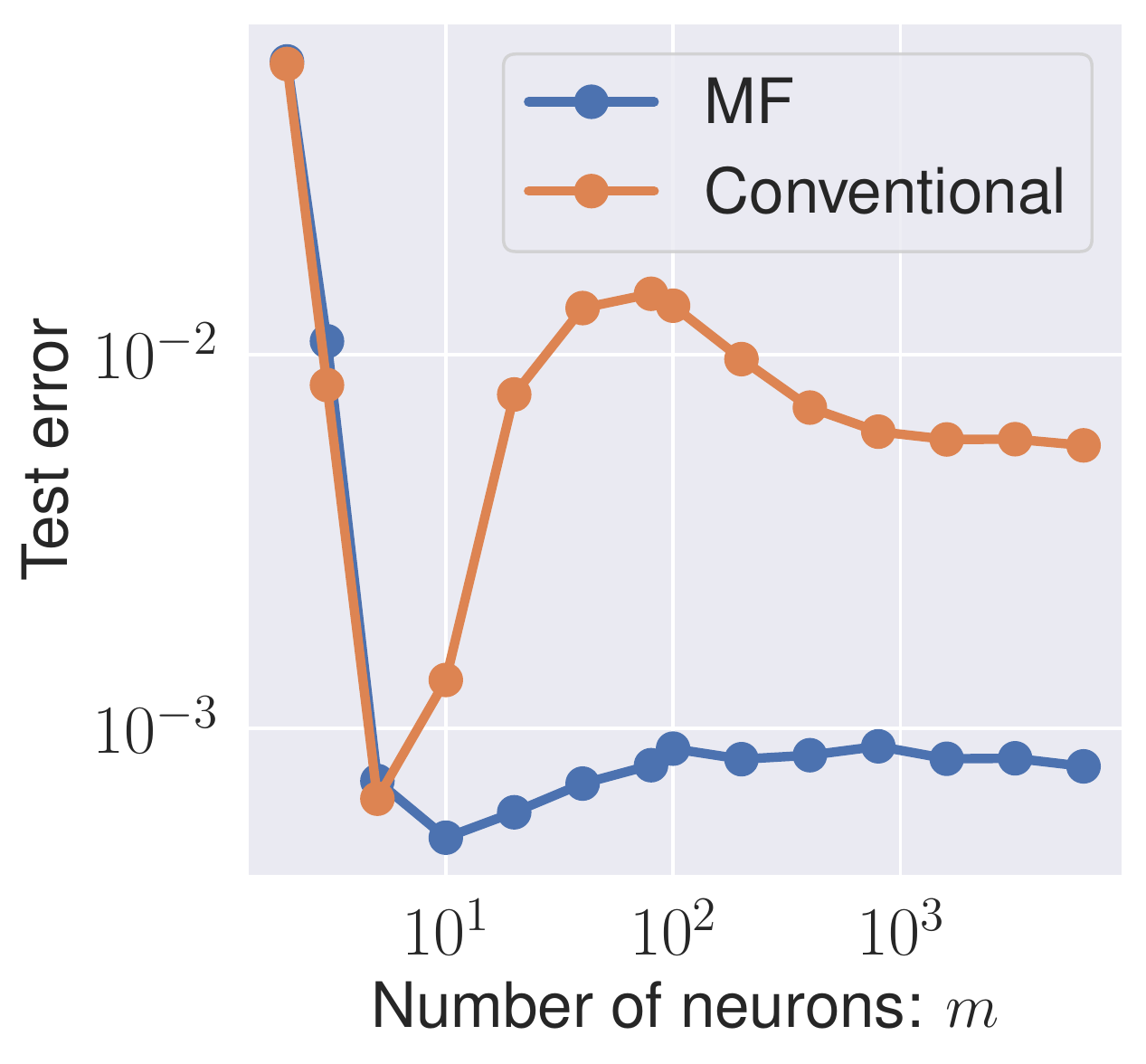}
    \includegraphics[width=0.49\textwidth]{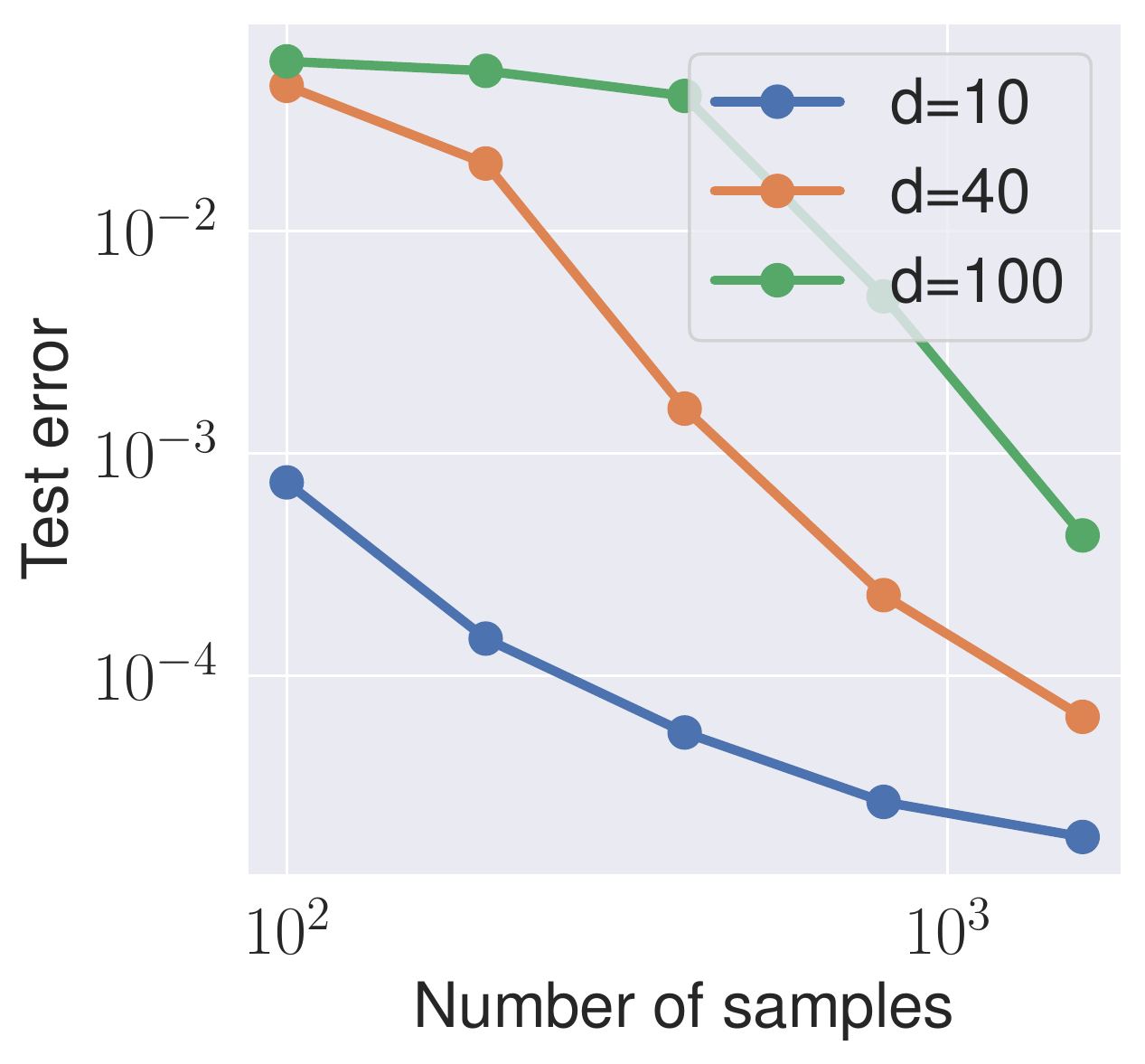}
    \caption{Circle neuron $f^*_2$.}
    \label{fig: mf-2b}
    \end{subfigure}
    \caption{ \small For each target function: (Left) test error of the GD-conventional and GD-MF solutions for different numbers of neurons. (Right) test error of the GD-MF solutions for dimensions: $d=10, 40, 100$ under the scaling $m=3n$.  
    }
    \label{fig: mf-2}
\end{figure}

Figure \ref{fig: mf-2} shows the test error for the case with finite sample. We see that the test 
error of GD-MF  varies more smoothly with the increase of network width. 
There is almost no clear deterioration of the performance even when we increase the network width to the highly over-parametrized regime. { This is clearly different from the case for GD-conventional.
 Moreover, we observe that the GD-MF solutions  do seem to suffer  from CoD at the regime $m\sim n$, which is not the case for GD-conventional.}

\section{Discussions and open questions}

The experimental results shown in this paper suggest the following:

1.  Both the training process and the generalization performance are quite sensitive under the conventional scaling, say with the
Xavier type initialization. With  the mean-field scaling, both are more stable, though not always better.

2.  In the NN-like regime under conventional scaling, the quenching-activation process provides the mechanism of implicit regularization, since it does allow the Barron norm (or the path norm)
to grow out of control.  Consequently the generalization gap is controlled.

An important question is the relation between the GD dynamics for the conventional and mean-field scalings.
Note that there is a one-to-one correspondence between the trajectories of GD-MF and GD-conventional. To see this, 
assume that $(\ba, \bB)$ obeys GD-MF.  Let $\tilde{\ba}={\ba}/{m}$ and $\tilde{\bB}= m \bB$, the dynamics for $\tilde{\ba}$ and $\tilde{\bB}$ becomes
\begin{equation}
\begin{aligned}
    \dot{\tilde{a}}_j(t) &= -\frac{1}{m^2n}\sum_{i=1}^n (f_m(\bx_i;\tilde{\ba}(t),\tilde{\bB}(t))-f^*(\bx_i))\sigma(\tilde{\bb}_j(t)^T \bx_i)\\
    \dot{\tilde{\bb}}_j(t) &= - \frac{1}{n}\sum_{i=1}^n (f_m(\bx_i;\tilde{\ba}(t),\tilde{\bB}(t))-f^*(\bx_i))\tilde{a}_j(t)\sigma'(\tilde{\bb}_j(t)^T\bx_i)\bx_i.
\end{aligned}
\end{equation}
One can easily verify that 
\begin{equation}\label{eqn: traj_mf_tilde}
\begin{aligned}
    \tilde{\ba}(t) &= \frac{1}{\sqrt{m}}\hat{\ba}(\frac{t}{m}) \\
    \tilde{\bB}(t) &= \sqrt{m}\hat{\bB}(\frac{t}{m})
\end{aligned}
\end{equation}
where $\hat{\ba}(\cdot)$ and $\hat{\bB}(\cdot) $ are solutions of GD-conventional with initial condition:
\begin{equation}
\begin{aligned}
\hat{\ba}(0) &= \frac 1{\sqrt{m}} \ba(0),
\hat{\bB}(0) &=\frac 1 {\sqrt{m}}  \bB(0).
\end{aligned}
\end{equation}
For the ReLU activation function considered here, this initial condition represents the same function as the initial condition $(\ba(0), \bB(0)$.
But they are different in the parameter space.  This difference gives rise to different dynamics.
In particular, the initial condition for the mean field model that corresponds to the Xavier initialization satisfies
$\ba(0) = O(1), \bB(0) =O( \sqrt{m})$.  It is not clear whether this is a simple limit in this case.


Many questions remain open.  We list some of these questions here.

1. Under the conventional scaling, does there exist  a third kind of behavior besides the NN-like and RF-like behavior? 
If not, where is exactly the transition between the two?

2. Is it possible to have CoD in time in the GD dynamics for Barron target functions?
We do know it is possible for non-Barron functions \cite{wojtowytsch2020can}.

3. How do the results discussed in this paper manifest  for practical datasets?

Obviously there are more questions that remain.  But we feel that these are the most pressing ones.
In any case, we believe that careful and systematic numerical investigation are needed in order to shed more
light on the mechanics behind the training and generalization performance of neural network models.

{\small 
\bibliographystyle{plain}
\bibliography{ref}
}

\newpage
\appendix

\section{Additional experiment results}
\subsection{The influence of $\beta$}
\label{sec: beta-infuence}
\begin{figure}[!h]
    \centering
    \begin{subfigure}{0.4\textwidth}
    \includegraphics[width=\textwidth]{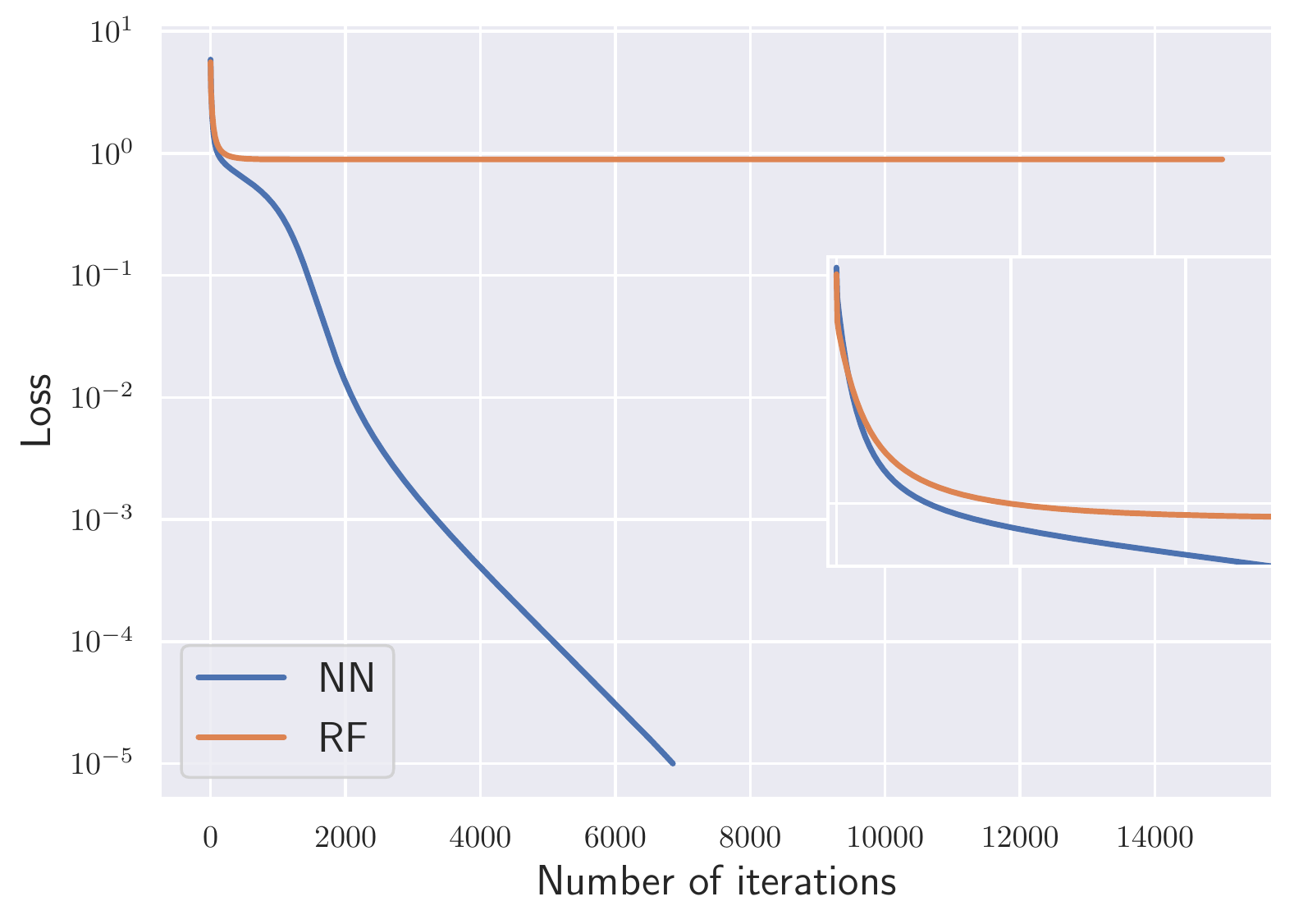}
    \caption{}
    \label{}
    \end{subfigure}
    \begin{subfigure}{0.4\textwidth}
    \includegraphics[width=\textwidth]{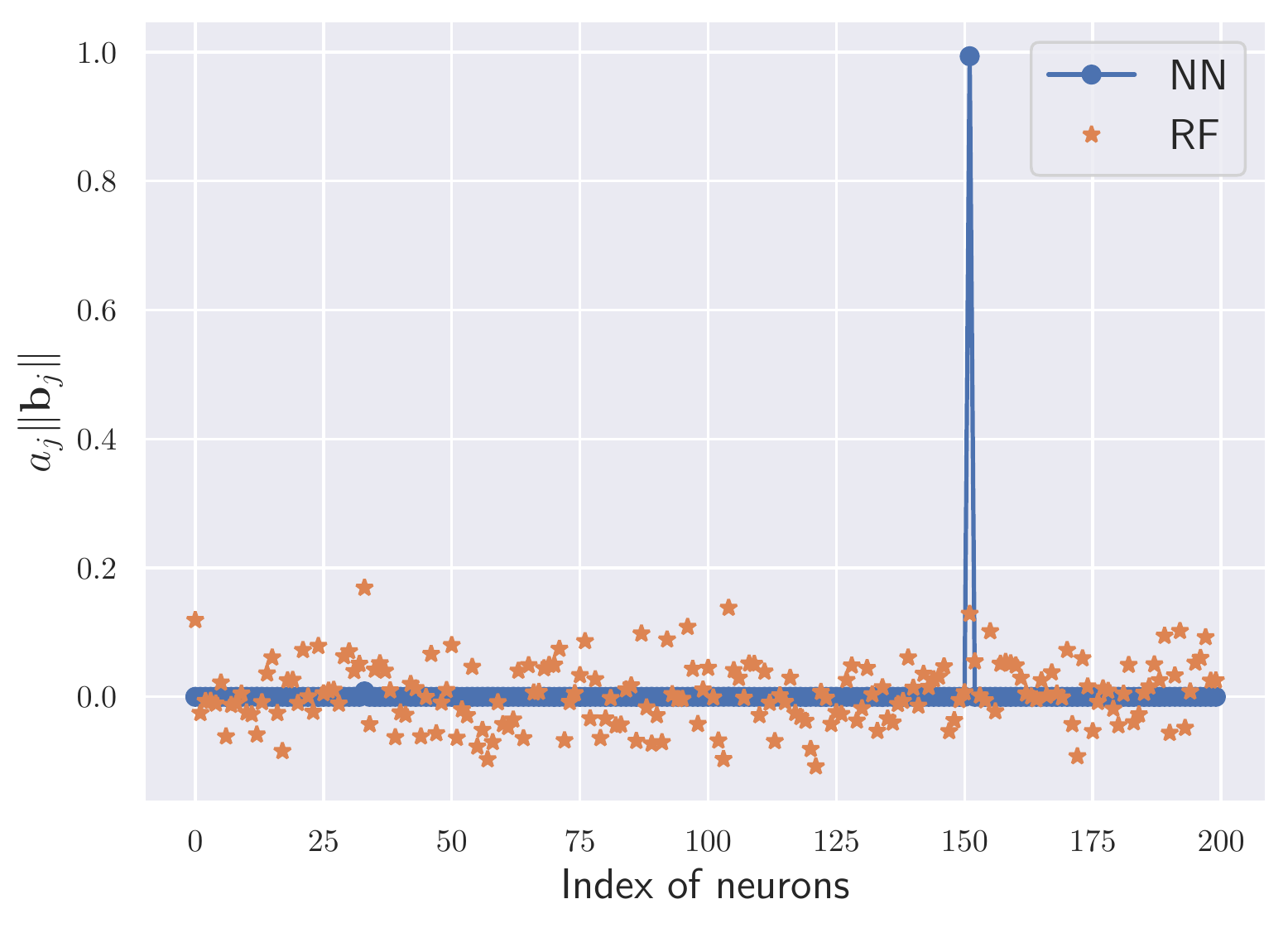}
    \caption{} 
    \label{}
    \end{subfigure}
    \begin{subfigure}{0.4\textwidth}
    \includegraphics[width=\textwidth]{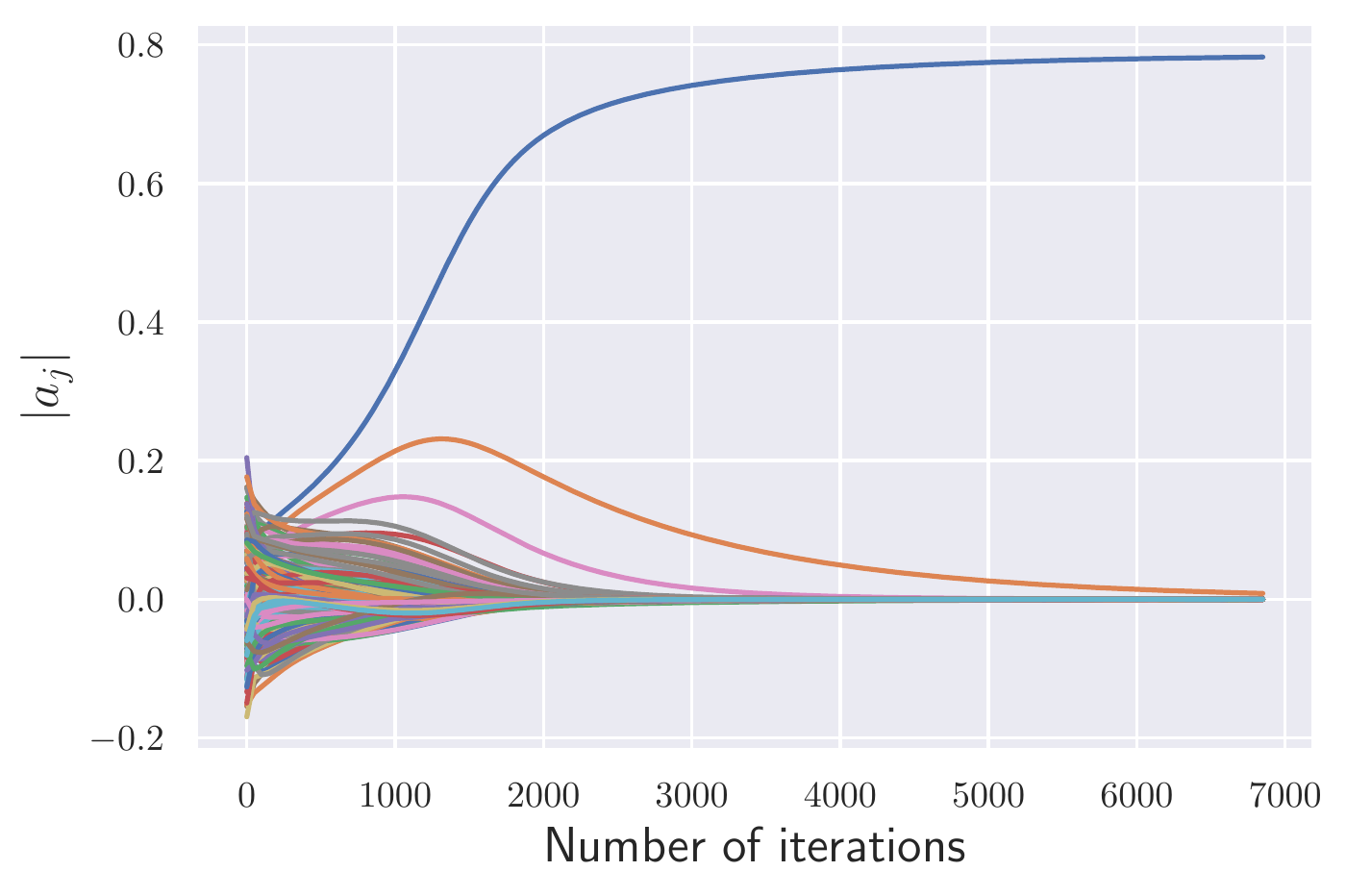}
    \caption{}
    \label{}
    \end{subfigure}
    \begin{subfigure}{0.4\textwidth}
    \includegraphics[width=\textwidth]{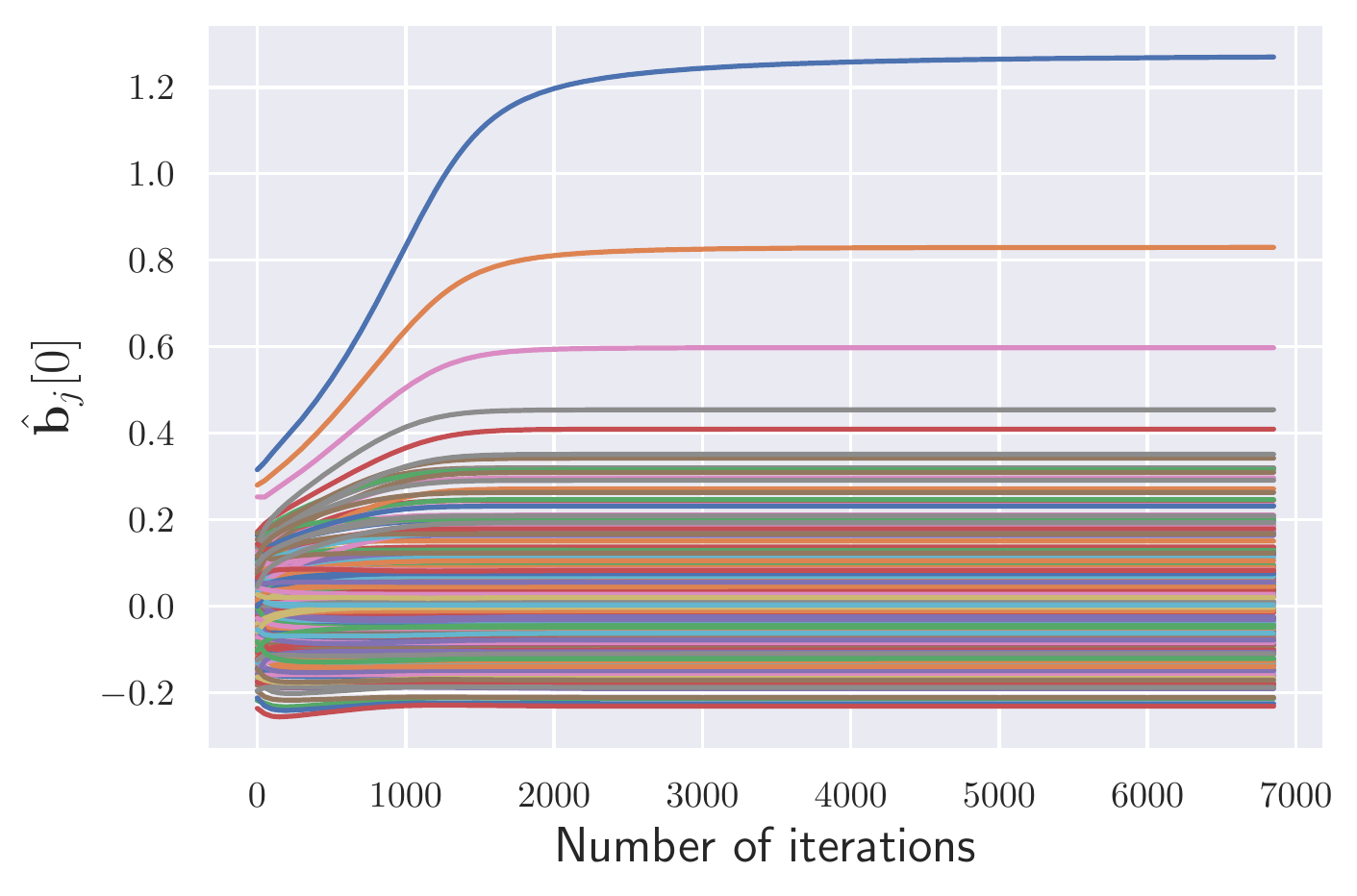}
    \caption{}
    \label{}
    \end{subfigure}
    \caption{\small The dynamic behavior of GD with $\beta=1/\sqrt{m}$. Here the target function is the single neuron $f_1^*$. $m=200, d=100$ and learning rate $\eta=10^{-3}$.  (a) The dynamic behavior of the population risk; (b) The $a$ coefficient of each neuron for the converged solution. We also plot the results of RF as comparison. (c) The dynamics of the $a$ coefficient of each neuron. (d) The dynamics of $\{\hat{\bb}_j[0]\}$, the projection of $\hat{\bb}_j$ to $\bb^*=\be_1$. }
    \label{fig: appendix-one-1}
\end{figure}

First we consider the case when $\beta=1/\sqrt{m}$.  Figure \ref{fig: appendix-one-1} shows the numerical results with the same setting as in Section \ref{sec: single-neuron-inifite-data}. We see that the dynamic behavior is  qualitatively the same as the case $\beta=0$.

For $\beta=1/m^{\gamma}$ with $\gamma\geq 1/2$,  the time-scale separation between the inner and outer layers always hold initially as long as $m$ is large enough. In this case we see basically the same kind of behavior as shown in the main text.
To simplify the experiments and the presentation, we set $\beta=0$, i.e. $\gamma=\infty$. In this case  we do not need to
take  very large values of $m$.

\subsection{Additional experimental results for the  mildly over-parametrized regime}
\label{sec: mild-appendix}
\begin{figure}
    \centering
    \begin{subfigure}{0.32\textwidth}
    \includegraphics[width=\textwidth]{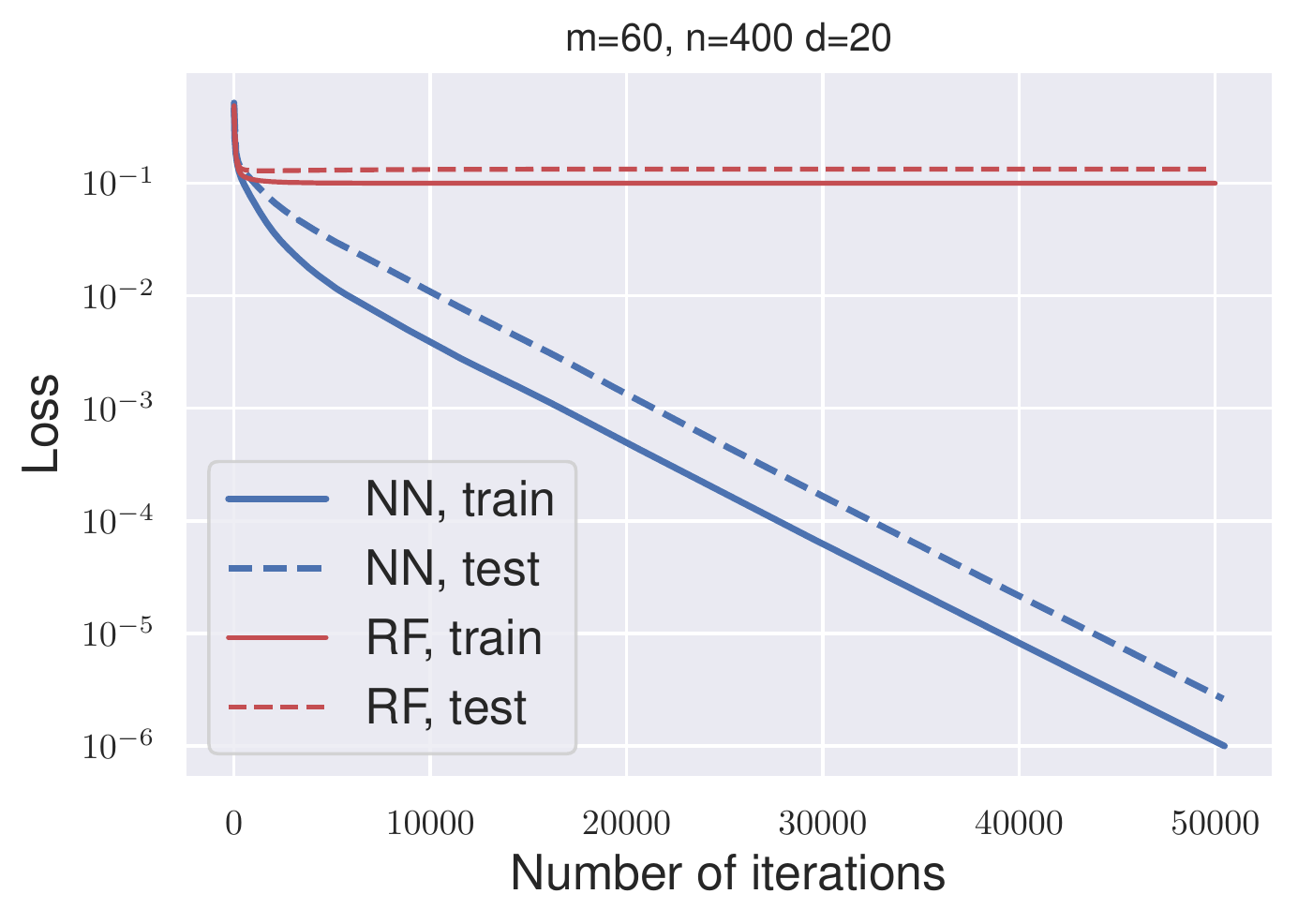}
    \caption{}\label{fig: appendix-mop-1a}
    \end{subfigure}
    \begin{subfigure}{0.32\textwidth}
    \includegraphics[width=\textwidth]{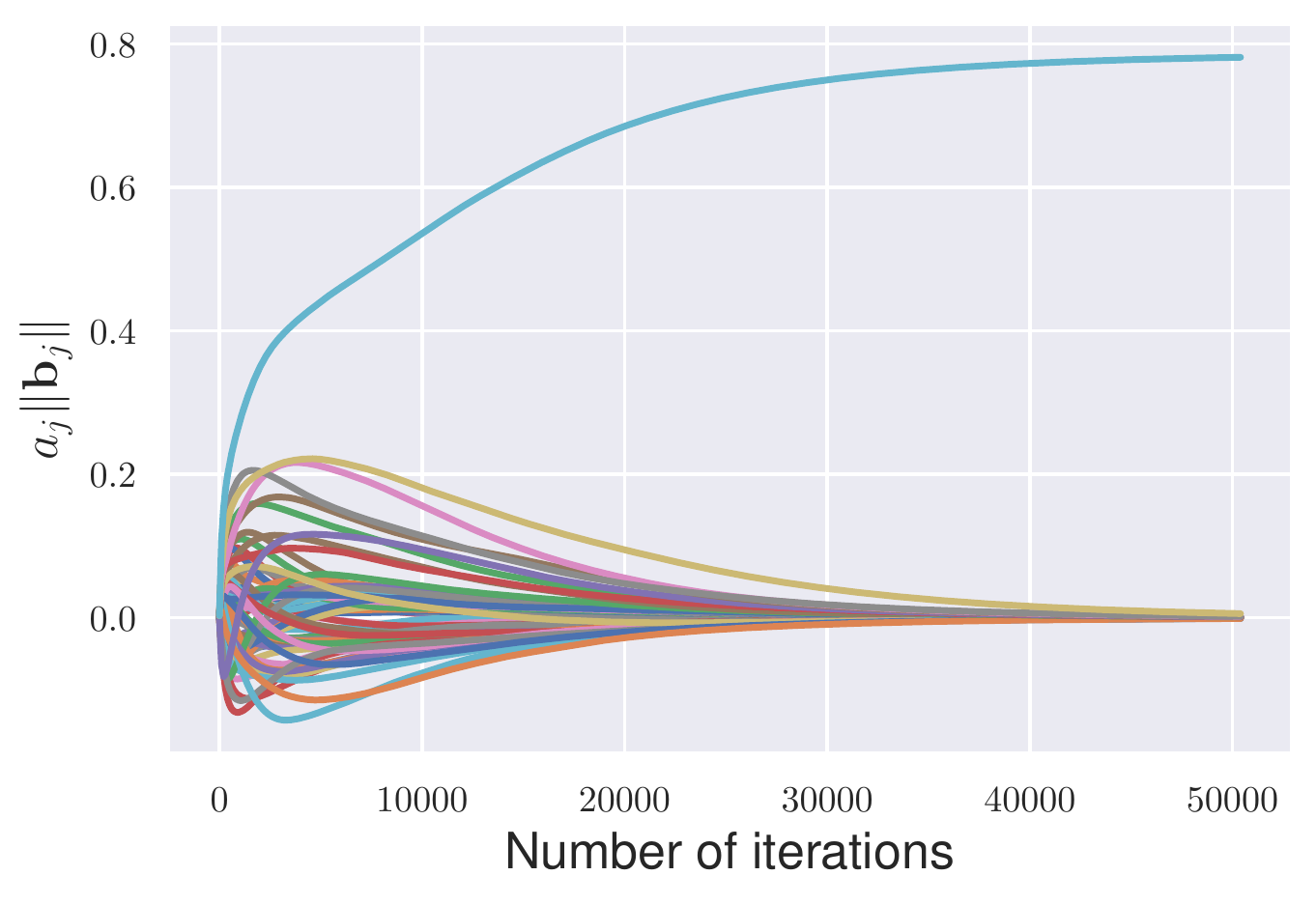}
    \caption{}\label{fig: appendix-mop-1b}
    \end{subfigure}
    \begin{subfigure}{0.32\textwidth}
    \includegraphics[width=\textwidth]{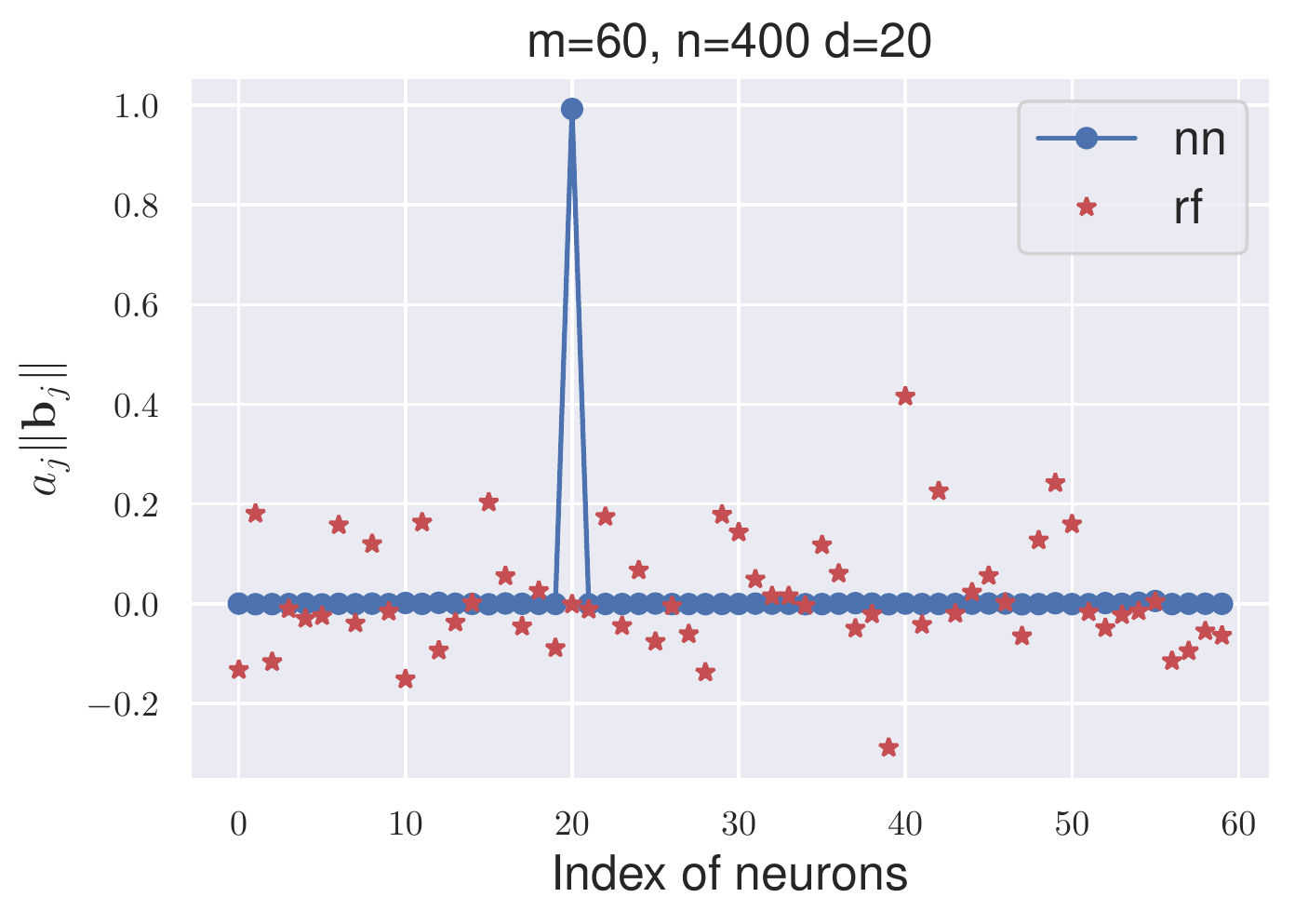}
    \caption{}\label{fig: appendix-mop-1c}
    \end{subfigure}
    \caption{\small   The dynamic behavior of the GD solutions for $m=3n/(d+1)$. Here $m=60, n=400, d=19$ and learning rate $\eta=0.001$. 
    }
    \label{fig: appendix-mop-1}
\end{figure}

\begin{figure}
    \centering
    \begin{subfigure}{0.32\textwidth}
    \includegraphics[width=\textwidth]{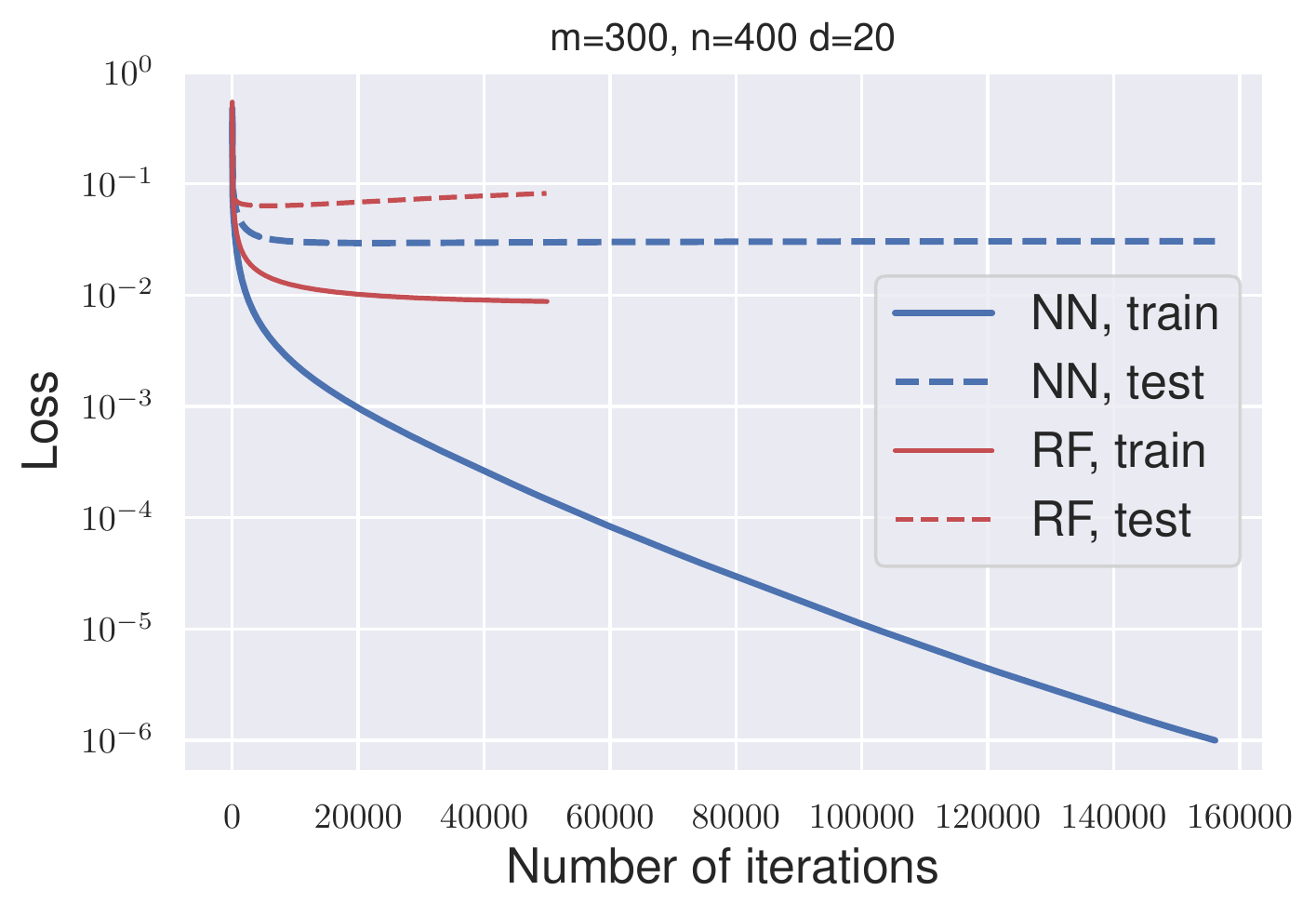}
    \caption{}\label{fig: appendix-mop-2a}
    \end{subfigure}
    \begin{subfigure}{0.32\textwidth}
    \includegraphics[width=\textwidth]{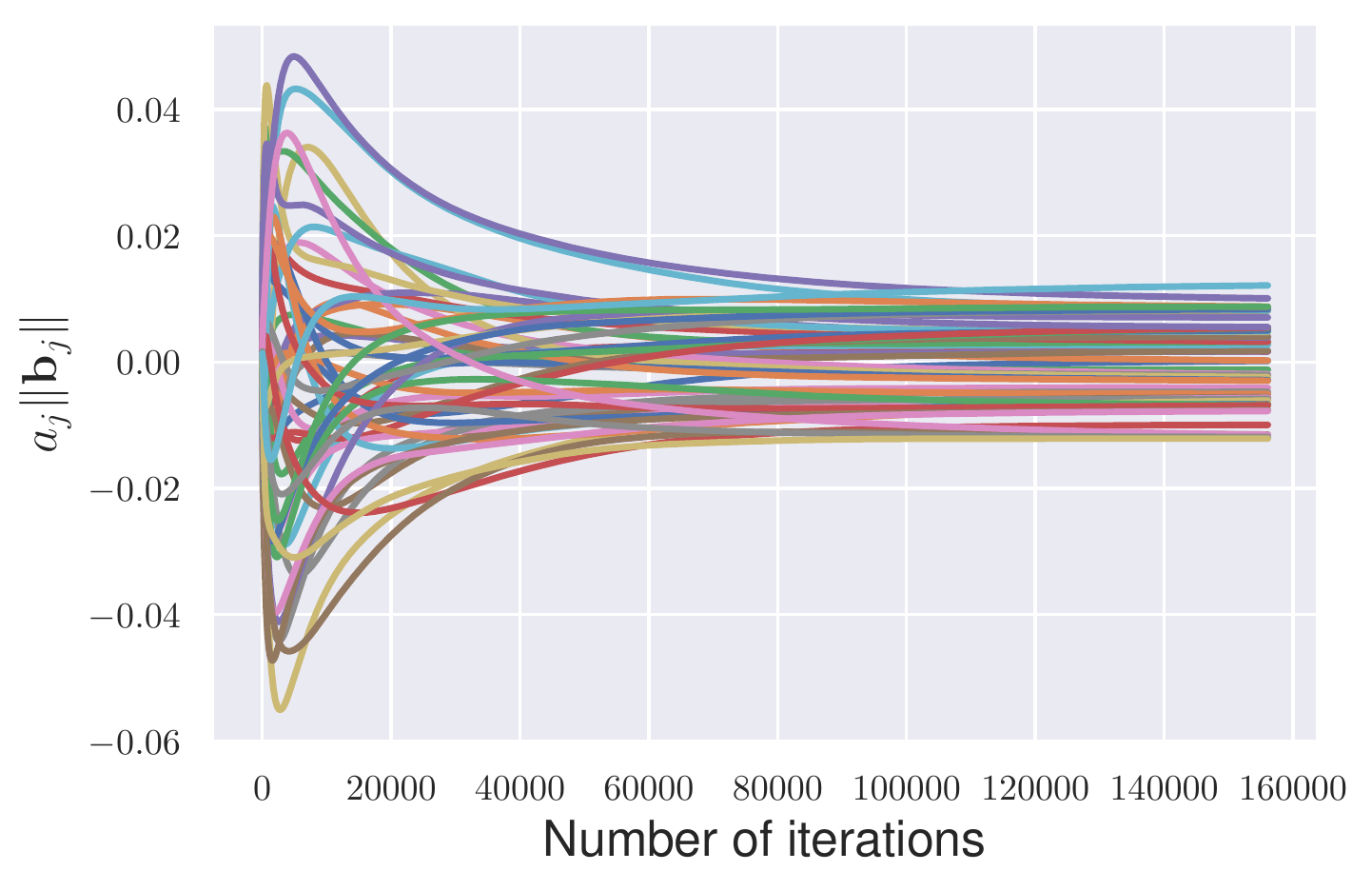}
    \caption{}\label{fig: appendix-mop-2b}
    \end{subfigure}
    \begin{subfigure}{0.32\textwidth}
    \includegraphics[width=\textwidth]{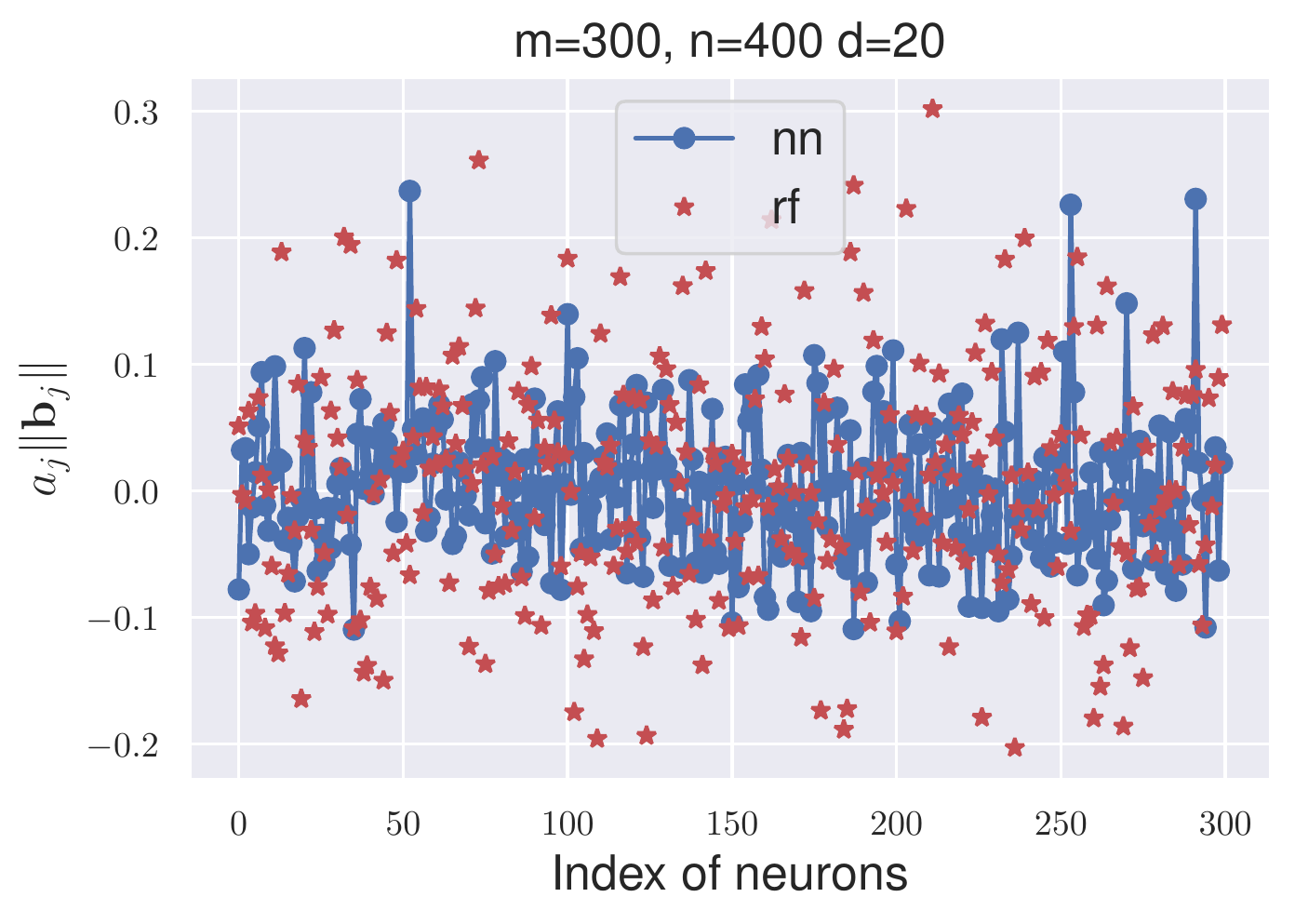}
    \caption{}\label{fig: appendix-mop-2c}
    \end{subfigure}
    \caption{\small   The dynamic behavior of  the GD solutions for $m=0.75n$. Here $m=300, n=400, d=19$ and learning rate $\eta=0.001$. 
    }
    \label{fig: appendix-mop-2}
\end{figure}

\end{document}